\documentclass{article} 
\usepackage[accepted]{icml2021}

\usepackage[colorlinks=True]{hyperref}

\usepackage[T1]{fontenc}
\usepackage[utf8]{inputenc}
\usepackage{multirow}
\usepackage{placeins}
\usepackage{graphicx}
\usepackage{url}
\usepackage{amsthm}
\usepackage{float}
\usepackage{amsmath}
\usepackage{microtype}
\usepackage{wrapfig}
\usepackage{xspace}
\usepackage{graphicx}
\usepackage{float}
\usepackage{subcaption}
\usepackage{booktabs} 
\usepackage{colortbl}
\usepackage{xcolor}
\usepackage{hyperref}
\usepackage{amssymb}
\usepackage{amsmath,bm}
\usepackage{enumitem}

\usepackage{pifont}

\usepackage{scalerel,stackengine}
\stackMath
\newcommand\reallywidehat[1]{%
\savestack{\tmpbox}{\stretchto{%
  \scaleto{%
    \scalerel*[\widthof{\ensuremath{#1}}]{\kern.1pt\mathchar"0362\kern.1pt}%
    {\rule{0ex}{\textheight}}
  }{\textheight}%
}{2.4ex}}%
\stackon[-6.9pt]{#1}{\tmpbox}%
}

\usepackage{hyperref}

\newcommand{\TrH}{$\mathrm{Tr}(\mathbf{H})$\xspace} 
 
\newcommand{\TrHf}{$\mathrm{Tr}(\mathbf{H_f})$\xspace} 
\newcommand{\TrFi}{$\mathrm{Tr}(\mathbf{F_i})$\xspace} 
\newcommand{\TrF}{$\mathrm{Tr}(\mathbf{F})$\xspace} 
\newcommand{\F}{$\mathbf{F}$\xspace}

\newcommand{\GPx}{\textsc{GP}\textsubscript{x}\xspace}
\newcommand{\GPr}{\textsc{GP}\textsubscript{r}\xspace}
\newcommand{\GP}{\textsc{GP}\xspace}
\newcommand{\FP}{\textsc{FP}\xspace}

\icmltitlerunning{Catastrophic Fisher Explosion: Early Phase Fisher Matrix Impacts Generalization}
\begin{document}
\twocolumn[
\icmltitle{Catastrophic Fisher Explosion: Early Phase Fisher Matrix Impacts Generalization}
\icmlcorrespondingauthor{Stanisław Jastrzębski}{staszek.jastrzebski@gmail.com}
\icmlsetsymbol{equal}{*}
\begin{icmlauthorlist}
\icmlauthor{Stanisław Jastrzębski}{langone,cds}
\icmlauthor{Devansh Arpit}{salesforce}
\icmlauthor{Oliver Åstrand}{cds}
\icmlauthor{Giancarlo Kerg}{um}
\icmlauthor{Huan Wang}{salesforce}
\icmlauthor{Caiming Xiong}{salesforce}
\icmlauthor{Richard Socher}{salesforce}
\icmlauthor{Kyunghyun Cho}{equal,cds,cifar}
\icmlauthor{Krzysztof J. Geras}{equal,langone,cds}
\end{icmlauthorlist}
\icmlaffiliation{cds}{Center of Data Science, New York University, USA}
\icmlaffiliation{langone}{NYU Langone Medical Center, New York University, USA}
\icmlaffiliation{salesforce}{Salesforce Research, USA}
\icmlaffiliation{um}{Université de Montréal, Canada}
\icmlaffiliation{cifar}{CIFAR Azrieli Global Scholar}
\vskip 0.3in
]

\printAffiliationsAndNotice{\icmlEqualContribution} 

\begin{abstract}

The early phase of training a deep neural network has a dramatic effect on the local curvature of the loss function. For instance, using a small learning rate does not guarantee stable optimization because the optimization trajectory has a tendency to steer towards regions of the loss surface with increasing local curvature. We ask whether this tendency is connected to the widely observed phenomenon that the choice of the learning rate strongly influences generalization. We first show that stochastic gradient descent (SGD) implicitly penalizes the trace of the Fisher Information Matrix (FIM), a measure of the local curvature, from the start of training. We argue it is an implicit regularizer in SGD by showing that explicitly penalizing the trace of the FIM can significantly improve generalization. We highlight that poor final generalization coincides with the trace of the FIM attaining a large value early in training, to which we refer as catastrophic Fisher explosion. Finally, to gain insight into the regularization effect of penalizing the trace of the FIM, we show that it limits memorization by reducing the learning speed of examples with noisy labels more than that of the examples with clean labels. 

\end{abstract}

\section{Introduction}

The exact mechanism behind implicit regularization effects in training of deep neural networks (DNNs) remains an extensively debated topic despite being considered a critical component in their empirical success~\citep{neyshabur2017,zhang_understanding_2016,jiang_fantastic_2020}. For instance, it is commonly observed that using a moderately large learning rate in the early phase of training results in better generalization~\citep{lecun_efficient_2012,goodfellow_deep_2016,jiang_fantastic_2020,bjorck_understanding_2018}.

\begin{figure*}
    \centering
    \begin{subfigure}[t]{0.45\textwidth}
       \includegraphics[height=0.5\textwidth]{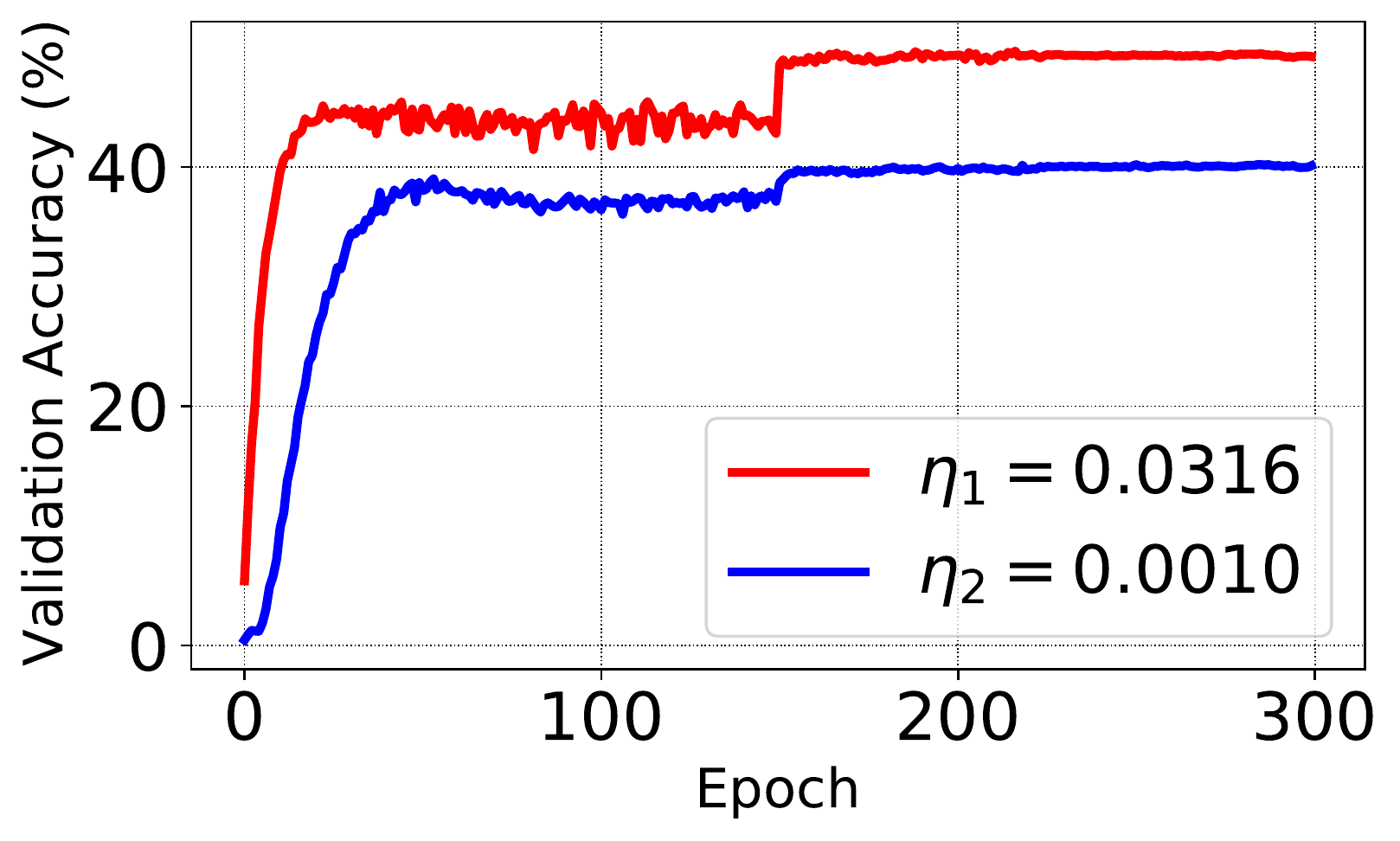}
 \hspace{-10em}\caption{Validation accuracy\,\,\,\,\,\,\,\,\,\,\,\,\,}
    \end{subfigure} \hspace{-4em}
    \begin{subfigure}[t]{0.45\textwidth}
       \includegraphics[height=0.51\textwidth]{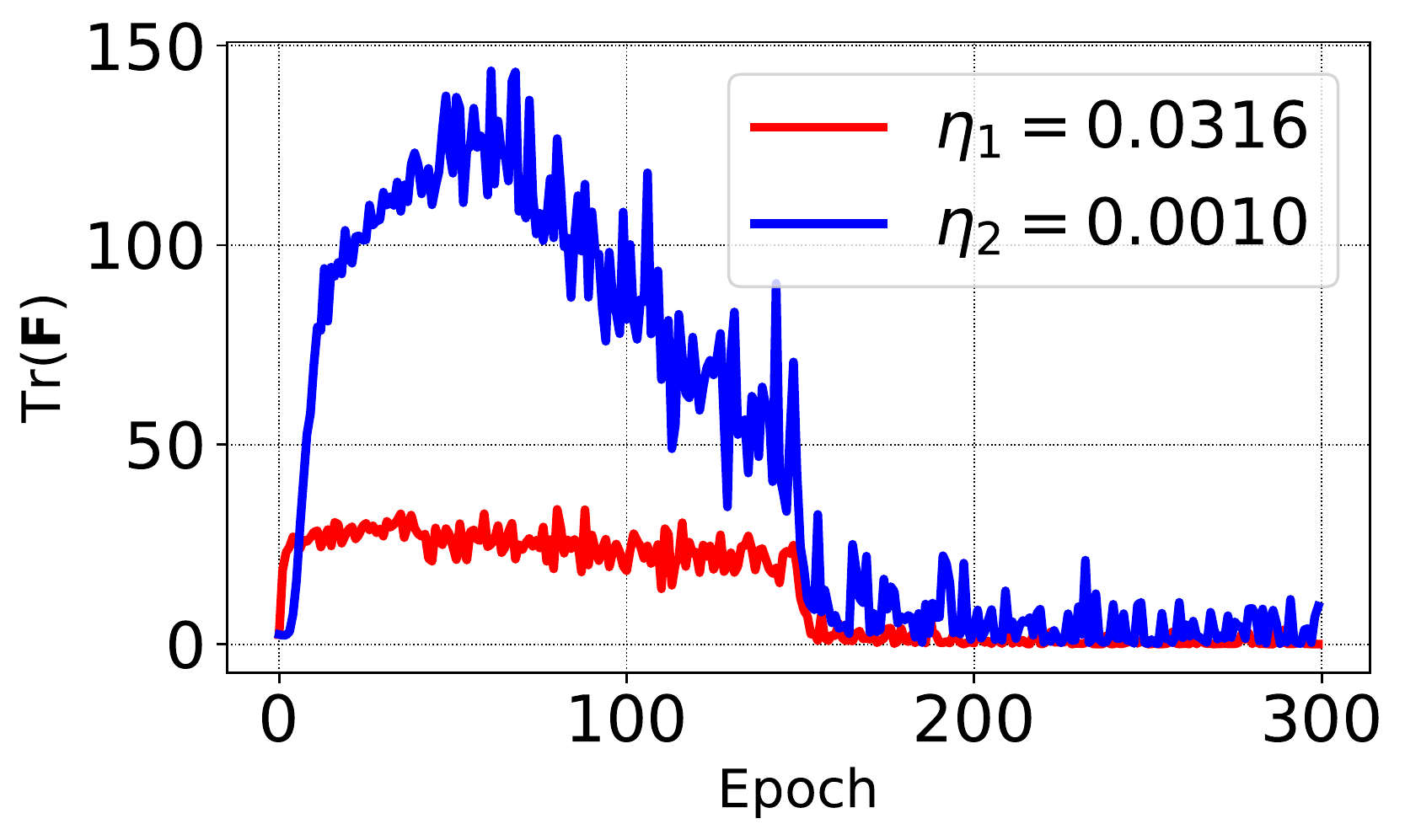}
\caption{The trace of the Fisher Information Matrix\,\,\,\,\,}
    \end{subfigure}
    \caption{Catastrophic Fisher explosion phenomenon demonstrated for Wide ResNet trained using stochastic gradient descent on the TinyImageNet dataset. Training with a small learning rate leads to a sharp increase in the trace of the Fisher Information Matrix (FIM) early in training (right), which coincides with strong overfitting (left). The trace of the FIM is a measure of the local curvature of the loss surface. Training is done with either a learning rate optimized using grid search ($\eta_1=0.0316$, red), or a small learning rate ($\eta_2=0.001$, blue). }
    \label{fig:casestudy}
\end{figure*}

Recent work suggests that the early phase of training of DNNs might hold the key to understanding some of these implicit regularization effects. In particular, the learning rate in the early phase of training has a dramatic effect on the local curvature of the loss function~\citep{Jastrzebski2020The,cohen2021gradient}. It has also been found that when using a small learning rate, the local curvature of the loss surface increases along the optimization trajectory until optimization is \emph{close to instability}.\footnote{That is, a small further increase in the local curvature is not possible without divergence. Interestingly, when training using gradient descent with a learning rate $\eta$, the largest eigenvalue of the Hessian of the training loss was observed to reach the critical value of $\frac{2}{\eta}$~\citep{cohen2021gradient}, at which training oscillates along the eigenvector corresponding to the largest eigenvalue of the Hessian.} 

These observations lead to a natural question: does the instability, and the corresponding dramatic change in the local curvature, in the early phase of training influence generalization? We investigate this question through the lens of the Fisher Information Matrix (FIM), a matrix that can be seen as approximating the local curvature of the loss surface~\citep{martens2014,thomas2020interplay}. \citet{achille_critical_2017,jastrzebski_relation_2018,golatkar2019,lewkowycz2020large} independently suggest that effects of the early phase of training on the local curvature critically influence the final generalization, but did not directly test this proposition.

Our main contribution is to show that \emph{implicit regularization effects due to using a large learning rate can be explained by its impact on the trace of the FIM (\TrF), a quantity that reflects the local curvature, from the beginning of training}. Our results suggest that the instability in the early phase is a critical phenomenon for understanding optimization in DNNs. This is in contrast to many prior theoretical works, which generally do not connect implicit regularization effects in SGD to the large instability in the early phase of training~\citep{chaudhari2018stochastic,smith2021on}.

We demonstrate on image classification tasks that \TrF early in training correlates with the final generalization performance across settings with different learning rates or batch sizes. We then show evidence that explicitly regularizing \TrF, which we call Fisher penalty, recovers generalization degradation due to training with a sub-optimal (small) learning rate, and can significantly improve generalization when training with the optimal learning rate. On the other hand, achieving large \TrF early in training, which may occur in practice when using a relatively small learning rate, or due to bad initialization, coincides with poor generalization. We call this phenomenon catastrophic Fisher explosion. Figure~\ref{fig:casestudy} illustrates this effect on the TinyImageNet dataset~\citep{Le2015TinyIV}.

 Our second contribution is an analysis of why implicitly or explicitly regularizing \TrF impacts generalization. Our key finding is that penalizing \TrF significantly improves generalization in training with noisy labels. We make a theoretical and empirical argument that penalizing \TrF can be seen as penalizing the gradient norm of noisy examples, which slows down their learning. We hypothesize that implicitly or explicitly regularizing \TrF amplifies the implicit bias of SGD to avoid memorization~\citep{arpit_closer_2017,Rahaman2018OnTS}. Finally, we also show that small \TrF in the early phase of training biases optimization towards a flat minimum~\citep{keskar_large-batch_2017}.

\begin{figure*}
    \centering
        \begin{subfigure}[t]{0.32\textwidth}
       \includegraphics[width=0.9\columnwidth ,trim=0.1in 0.in 0.in 0.in,clip]{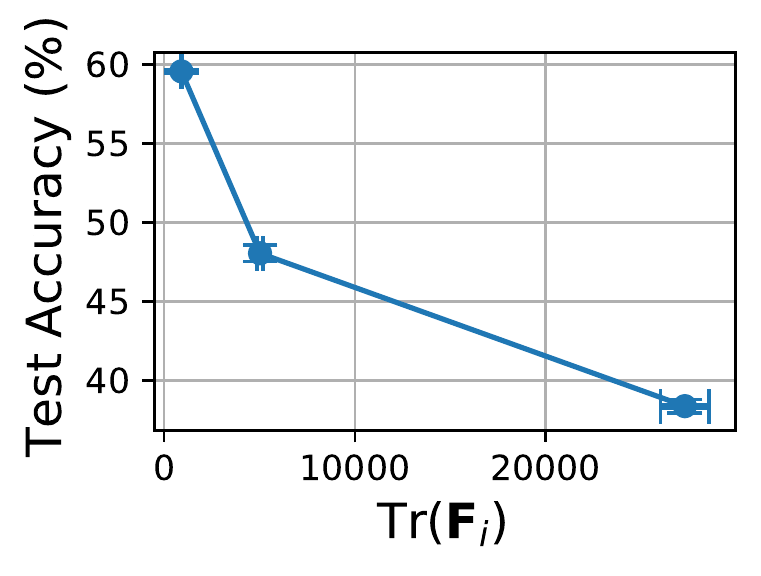}
\caption{ImageNet (w/o augmentation)}
    \end{subfigure}
    \hfill
        \begin{subfigure}[t]{0.32\textwidth}
       \includegraphics[width=0.9\columnwidth ,trim=0.1in 0.in 0.in 0.in,clip]{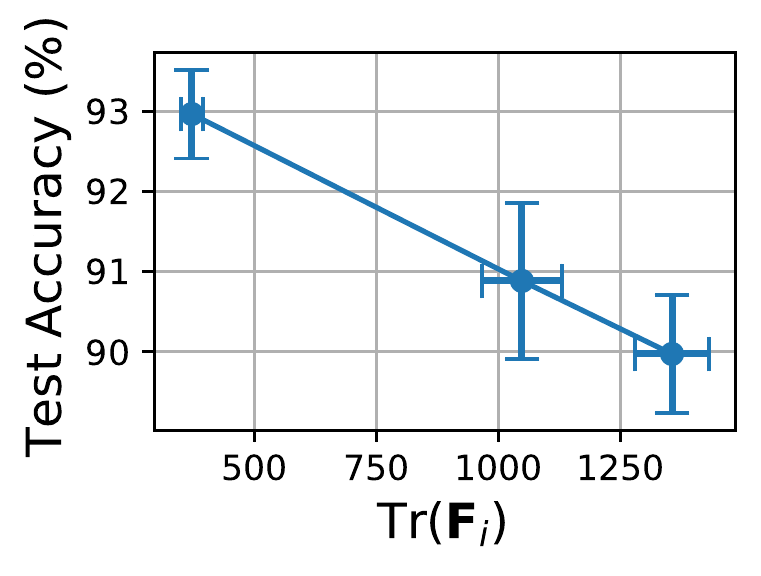}
    \caption{CIFAR-10 (with augmentation)}
    \end{subfigure}
    \hfill
    \begin{subfigure}[t]{0.32\textwidth}
      \includegraphics[width=0.97\columnwidth ,trim=0.1in 0.in 0.in 0.in,clip]{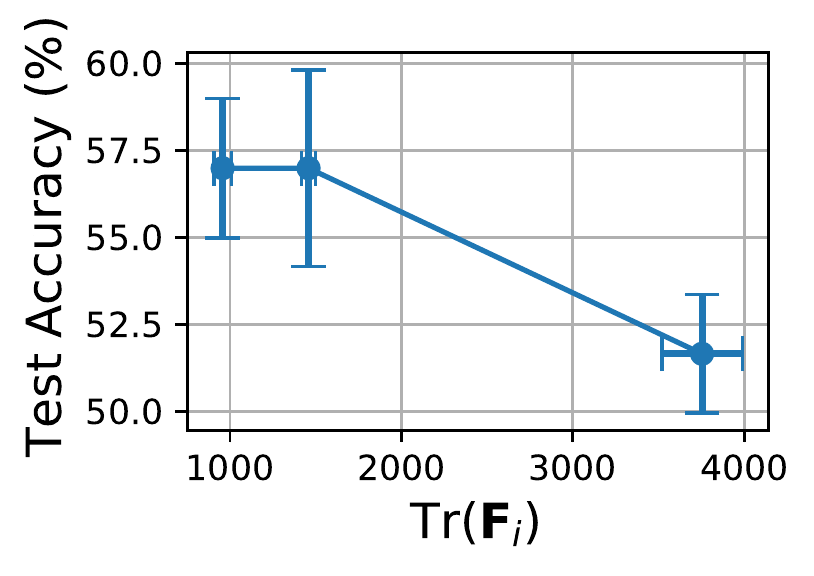}
    \caption{CIFAR-100 (w/o augmentation)}
    \end{subfigure}
    \caption{Association between \TrF in the initial phase of training (\TrFi) and test accuracy on ImageNet, CIFAR-10 and CIFAR-100 datasets. Each point corresponds to multiple seeds and a specific value of learning rate. \TrFi is recorded during the early phase of training (2-7 epochs, see the main text for details). The plots show that early \TrF is predictive of final generalization. Analogous results illustrating the influence of batch size are shown in Appendix \ref{sec_early_final_appendix}}.
    \label{fig:IF_LR_trF}
\end{figure*}

\section{Implicit and explicit regularization of the Fisher Information Matrix}

\paragraph{Fisher Information Matrix} 

Consider a probabilistic classification model $p_{\bm{\theta}}(y|\bm{x})$, where $\bm{\theta}$ denotes its parameters. Let $\ell(\bm{x},y;\bm{\theta})$ be the cross-entropy loss function calculated for input $\bm{x}$ and label $y$. Let $g(\bm{x},y; \bm{\theta})=\frac{\partial}{\partial \mathbf{\theta}}\ell(\bm{x}, y; \bm{\theta})$ denote the gradient of the loss computed for an example $(\bm{x},y)$. The central object that we study is the Fisher Information Matrix \F defined as
\begin{equation}
\mathbf{F}(\bm{\theta}) = \mathbb{E}_{x \sim \mathcal{X},\hat y \sim p_{\theta}(y|\bm{x})} [{g}(\bm{x},\hat y) {g}(\bm{x},\hat y)^T ],
\end{equation}
where the expectation is often approximated using the empirical distribution $\mathcal{X}$ induced by the training set. Later, we also look into the Hessian $\mathbf{H}(\bm{\theta})=\frac{\partial^2}{\partial \bm{\theta}^2} \ell(\bm{x},y;\bm{\theta})$. We denote the trace of $\mathbf{F}$ and $\mathbf{H}$ matrices by \TrF and \TrH.

The FIM can be seen as an approximation to the Hessian~\citep{martens2014}. In particular, as $p(y|\bm{x}; \theta) \rightarrow \hat p(y | \bm{x})$, where $\hat p(y | \bm{x})$ is the empirical label distribution, the FIM converges to the Hessian. \citet{thomas2020interplay} showed on image classifications tasks that $\mathrm{Tr}(\mathbf{H})\approx \mathrm{Tr}(\mathbf{F})$ along the optimization trajectory, which we also demonstrate in Supplement~\ref{app:TrH_and_TrF_correlate}. Crucially, note that while \TrH uses label information, \TrF does not use any label information, in contrast to the ``empirical Fisher'' studied for example in \citet{kunstner2019limitations}.

\paragraph{Fisher Penalty} 

The early phase has a drastic effect on the trajectory in terms of the local curvature of the loss surface~\citep{achille_critical_2017,jastrzebski_relation_2018,gur-ari_gradient_2018,lewkowycz2020large,leclerc2020regimes}. In particular, \citet{lewkowycz2020large,jastrzebski_relation_2018} show that using a large learning rate in stochastic gradient descent biases training towards low curvature regions of the loss surface early in training. For example, using a large learning rate in SGD was shown to result in a rapid decay of \TrH along the optimization trajectory~\cite{jastrzebski_relation_2018}. 

Our main contribution is to propose and investigate a specific mechanism by which using a large learning rate or a small batch size implicitly influences final generalization. Our first insight is to shift the focus from studying the Hessian, to studying the properties of the FIM. Concretely, we hypothesize that using a large learning rate or a small batch size improves generalization by implicitly penalizing \TrF from the very beginning of training. 

In order to study the effect of implicit regularization of \TrF, we introduce a regularizer that explicitly penalizes \TrF. First, we note that \TrF can be written as 
 \begin{equation}
 \label{eq:trf}
\mathrm{Tr}(\mathbf{F}) = \mathbb{E}_{x \sim \mathcal{X},\hat y \sim p_{\theta}(y|\bm{x})} \left[ \Vert \frac{\partial}{\partial \mathbf{\theta}}\ell(\bm{x},\hat y) \Vert_2^2 \right].
 \end{equation}
Thus, to regularize \TrF, we can simply add $\frac{1}{M} \sum_{i=1}^M \left\|  g(\bm{x}_i, \hat y_i) \right\|^2$ term to the loss function, which can be efficiently back-propagated through. This, however, requires a large number of samples to efficiently regularize \TrF, as we show in our experiments. Instead, we add the following term to the loss function:
\begin{equation}
\label{eq_fisher_objective}
\ell'(\bm{x}, y) = \frac{1}{B} \sum_{i=1}^B \ell(\bm{x}_i,y_i) + \alpha \left\| \frac{1}{B} \sum_{i=1}^B  g(\bm{x}_i, \hat y_i) \right\|^2,
\end{equation}
where $(\bm{x}, y)$ is a mini-batch of size $B$, $\hat y_i$ is sampled from $p_{\bm{\theta}}(y|\bm{x}_i)$, $\alpha$ is a hyperparameter. Importantly, the equation does not involve target labels. Finally, we compute the gradient of the second term only every 10 optimization steps, and in a given iteration use the most recently computed gradient. We refer to this regularizer as Fisher penalty (\FP).

In the experiments, we show that this formulation efficiently penalizes \TrF. We attribute this largely to the fact that $ \left\| \frac{1}{B} \sum_{i=1}^B g(\bm{x}_i, \hat y_i) \right\|^2$ and \TrF are strongly correlated during training. We discuss this observation, and ablate the approximations, in more detail in Supplement~\ref{app:approx_in_FP}.

\begin{table*}
\centering
\caption{Fisher penalty (\FP) effectively models implicit regularization that arises in SGD due to using large learning rates. Using a 10-30x smaller learning rate (Baseline) results in up to 9\% degradation in test accuracy on popular image classification benchmarks (c.f. to \textit{optimal} $\eta^*$). Adding \FP, which explicitly regularizes \TrF, substantially improves generalization and closes the gap to $\eta^*$. Green cells correspond to runs that finished with, at most, 1\% lower test accuracy than when training with $\eta*$. }

\label{tab:fisher_penalty_setting1}
\begin{tabular}{cll|ll|ll}
\toprule
              Setting & $\eta^*$ & Baseline & \GPx & \GP &  \FP & \GPr \\
                       \midrule
              Wide ResNet / TinyImageNet (aug.) &  54.67\% &     52.57\% &              52.79\% &            \cellcolor{green!25}56.44\% &          \cellcolor{green!25}\textbf{56.73\%} &              \cellcolor{green!25}55.41\% \\
\midrule
        DenseNet / CIFAR-100 (w/o aug.)   &  66.09\% &     58.51\% &              62.12\% &            64.42\% &         \cellcolor{green!25} \textbf{66.41\%} &              \cellcolor{green!25}66.39\% \\
           VGG11 / CIFAR-100 (w/o aug.) &  45.86\% &     36.86\% &              \cellcolor{green!25}45.26\% &            \cellcolor{green!25}47.35\% &          \cellcolor{green!25}\textbf{49.87\%} &              \cellcolor{green!25}48.26\% \\
         WResNet / CIFAR-100 (w/o aug.) &  53.96\% &     46.38\% &     \cellcolor{green!25}\textbf{58.68\%} &            \cellcolor{green!25}57.68\% &                   \cellcolor{green!25}57.05\% &              \cellcolor{green!25}58.15\% \\
         \midrule
           SimpleCNN / CIFAR-10 (w/o aug.)   &  76.94\% &     71.32\% &              75.68\% &            75.73\% &                   \cellcolor{green!25}79.66\% &     \cellcolor{green!25}\textbf{79.76\%} \\
\bottomrule
\end{tabular}
\end{table*}

\paragraph{Catastrophic Fisher Explosion} 

To illustrate the concepts mentioned in this section, we train a Wide ResNet model (depth 44, width 3)~\citep{zagoruyko2016} on the TinyImageNet dataset with SGD and two different learning rates. We illustrate in Figure~\ref{fig:casestudy} that the small learning rate leads to dramatic overfitting, which coincides with a sharp increase in \TrF in the early phase of training. We also show in Supplement~\ref{app:closer_look} that these effects cannot be explained by the difference in learning speed between runs with smaller and learning rates. We call this phenomenon catastrophic Fisher explosion. 

\paragraph{Why Fisher Information Matrix} The benefit of the FIM is that it can be efficiently regularized during training. In contrast, \citet{wen2018smoothout,foret2021sharpnessaware} had to rely on certain approximations to efficiently regularize curvature. Equally importantly, the FIM is related to the gradient norm, and as such its effect on learning is more interpretable. We will leverage this connection to argue that \FP slows down learning on noisy examples in the dataset.

\paragraph{Concurrent work on a different mechanism}

\citet{barrett2020implicit,smith2021on} concurrently argue that the implicit regularization effects in SGD can be expressed as a form of a gradient norm penalty. The key difference is that we link the mechanism of implicit regularization to the fact optimization is \emph{close to instability} in the early phase. Supporting this view, we observe that \FP empirically performs better than gradient penalty, and that \FP works best when applied from the start of training.

\section{Early-phase \TrF and final generalization}
\label{sec_trf_generalization}

Using a large learning rate ($\eta$) or small batch size ($S$) in SGD steers optimization to a lower curvature region of the loss surface. However, it remains a hotly debated topic whether such choices explain strong regularization effects~\citep{dinh_sharp_2017,yoshida2017spectral,He2019,tsuzuku2019normalized}. We begin by studying the connection between \TrF and generalization in experiments across which we vary $\eta$ or $S$ in SGD.

\paragraph{Experimental setup} 

We run experiments in two settings: (1) ResNet-18 with Fixup~\cite{he_deep_2015,zhang2019fixup} trained on the ImageNet dataset~\citep{deng_imagenet_2009}, (2) ResNet-26 initialized as in \cite{arpit2019initialize} and trained on the CIFAR-10 and CIFAR-100 datasets~\citep{krizhevsky_learning_2009}. We train each architecture using SGD, with various values of $\eta$, $S$, and random seed.

We define \TrFi as \TrF during the initial phase of training. How long we consider the early-phase \TrF to be is determined by measuring when the training loss crosses a task-specific threshold $\epsilon$ that roughly corresponds to the moment when \TrF achieves its maximum value. For ImageNet, we use learning rates 0.001, 0.01, 0.1, and $\epsilon = 3.5$. For CIFAR-10, we use learning rates 0.007, 0.01, 0.05, and $\epsilon = 1.2$. For CIFAR-100, we use learning rates 0.001, 0.005, 0.01, and $\epsilon = 3.5$. In all cases, training loss reaches $\epsilon$ between 2 and 7 epochs across different hyper-parameter settings. We repeat similar experiments for different batch sizes in Supplement~\ref{sec_early_final_appendix}. The remaining training details can be found in Supplement~\ref{sec_early_final_details}.

\paragraph{Results} 

Figure~\ref{fig:IF_LR_trF} shows the association between \TrFi and test accuracy across runs with different learning rates. We show results for CIFAR-10 and CIFAR-100 when varying the batch size in Figure \ref{fig:IF_LR_trF_bs} in the Supplement. We find that \TrFi correlates well with the final generalization in our setting, which provides initial evidence for the importance of \TrF. It also serves as a stepping stone towards developing a more granular understanding of the role of implicit regularization of \TrF in the following sections.

\section{Fisher Penalty}
\label{sec:fisher_penalty}

To understand the significance of the identified correlation between \TrFi and generalization, we now run experiments in which we directly penalize \TrF. We focus our attention on the identified effect that using a high learning rate has on \TrF, especially early in training.

\begin{figure}
    \centering
    \begin{subfigure}[t]{0.236\textwidth}
    \centering
       \includegraphics[width=1.0\columnwidth]{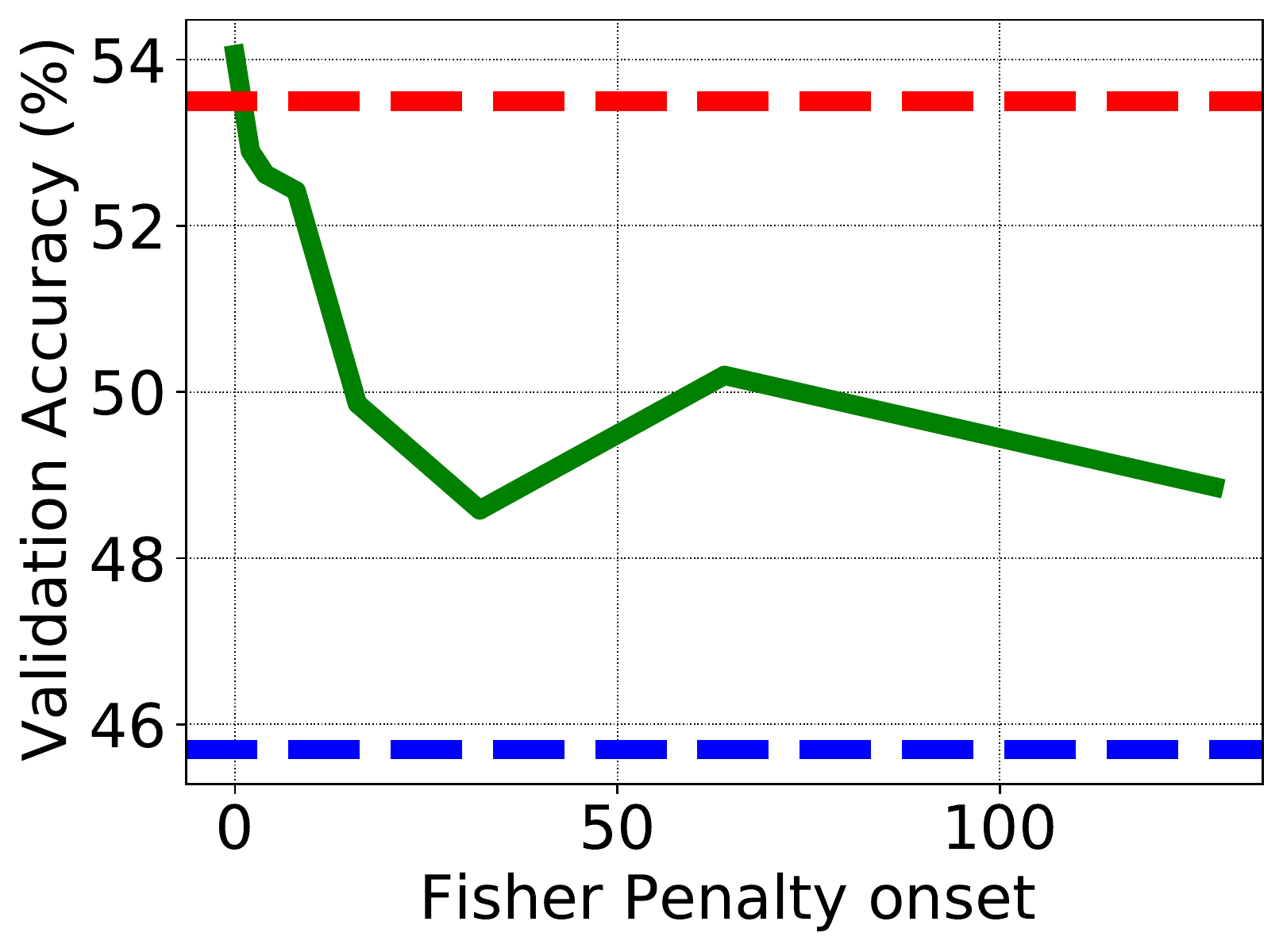}
\caption{Wide ResNet on CIFAR-100}
    \end{subfigure}
    \begin{subfigure}[t]{0.236\textwidth}
       \includegraphics[width=1.0\columnwidth]{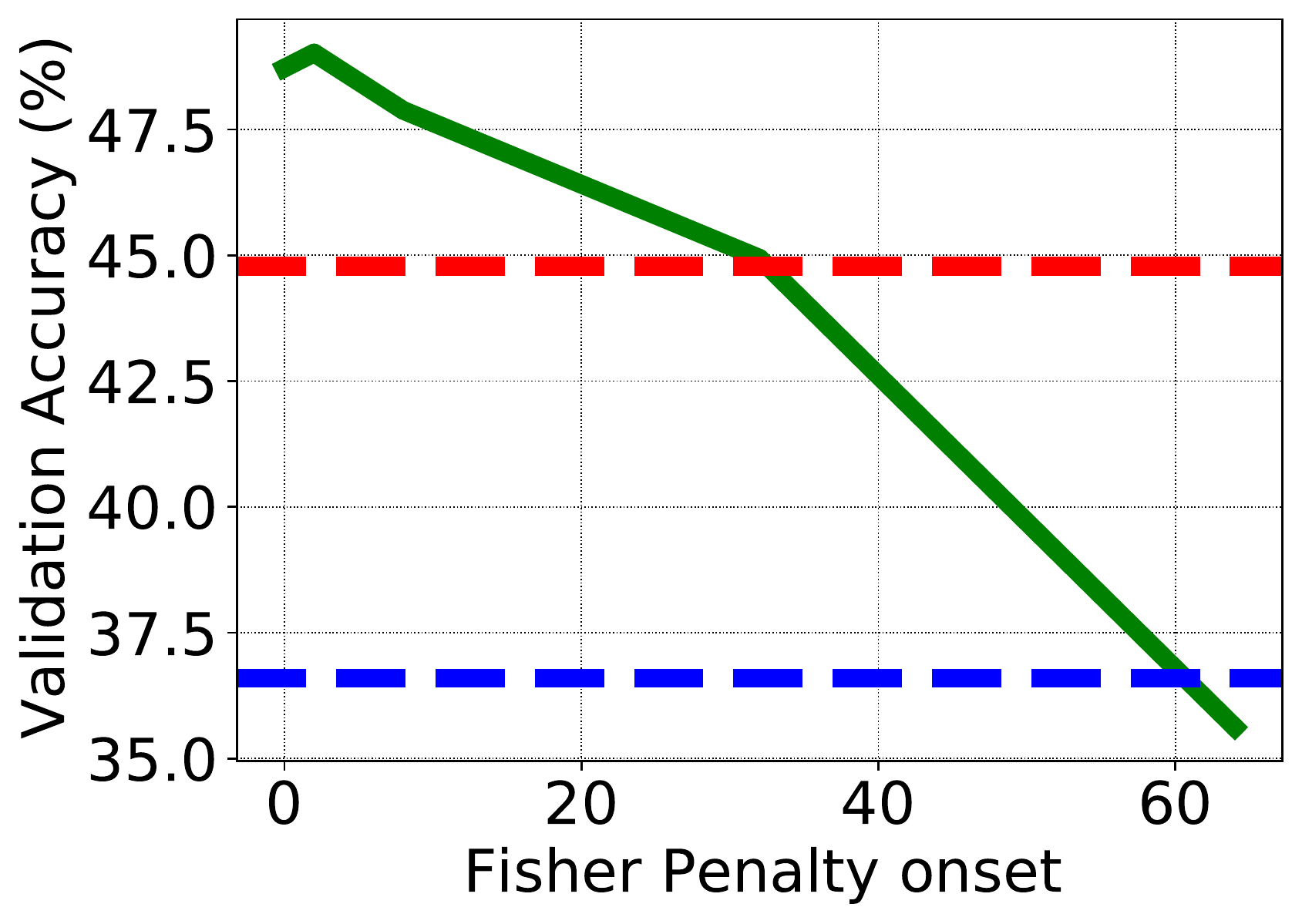}
\caption{VGG-11 on CIFAR-100}
    \end{subfigure}
        \begin{subfigure}[t]{0.236\textwidth}
       \includegraphics[width=1.0\columnwidth]{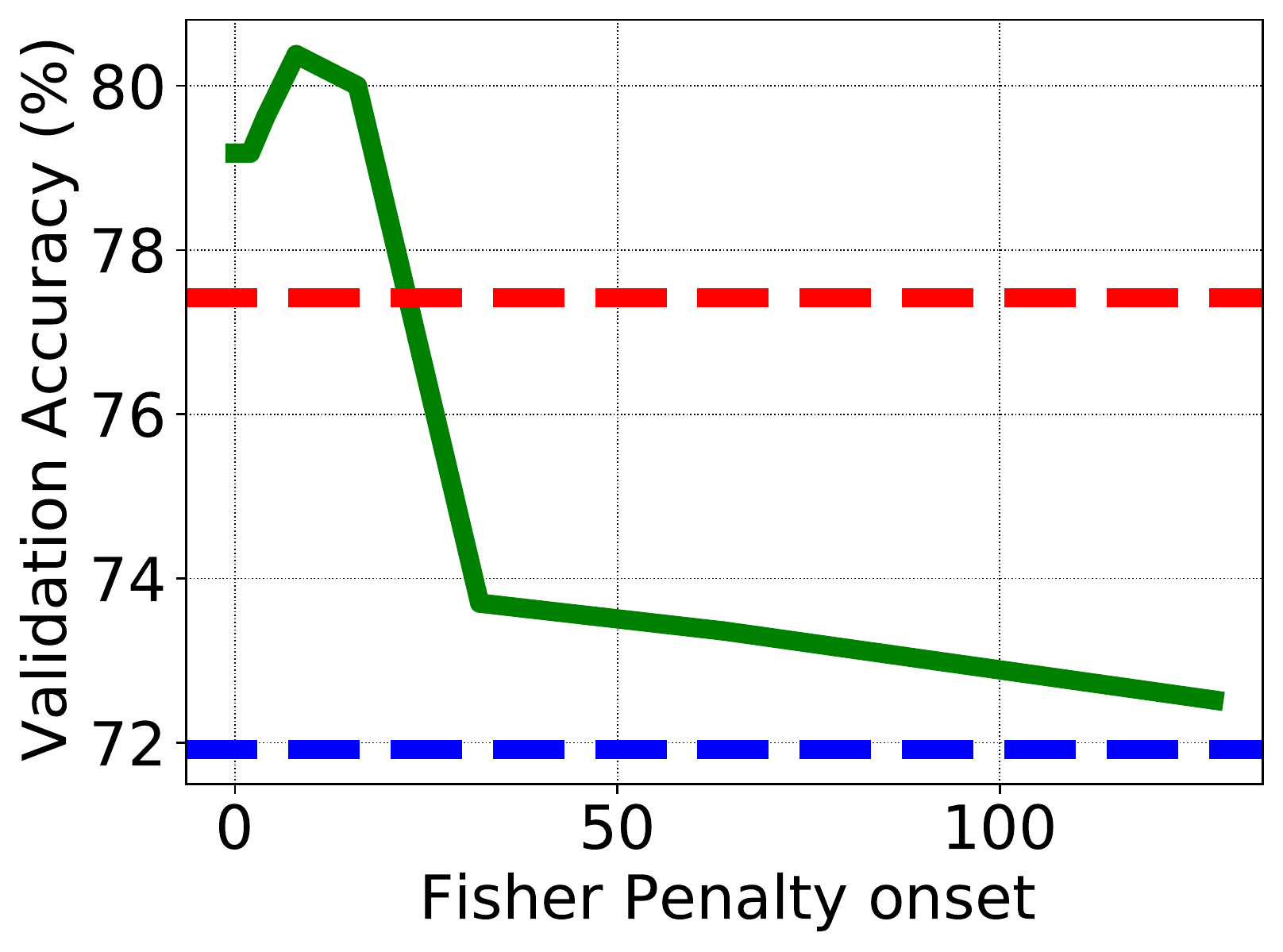}
\caption{Simple CNN on CIFAR-10}
    \end{subfigure}
        \begin{subfigure}[t]{0.236\textwidth}
              \includegraphics[width=1.0\columnwidth]{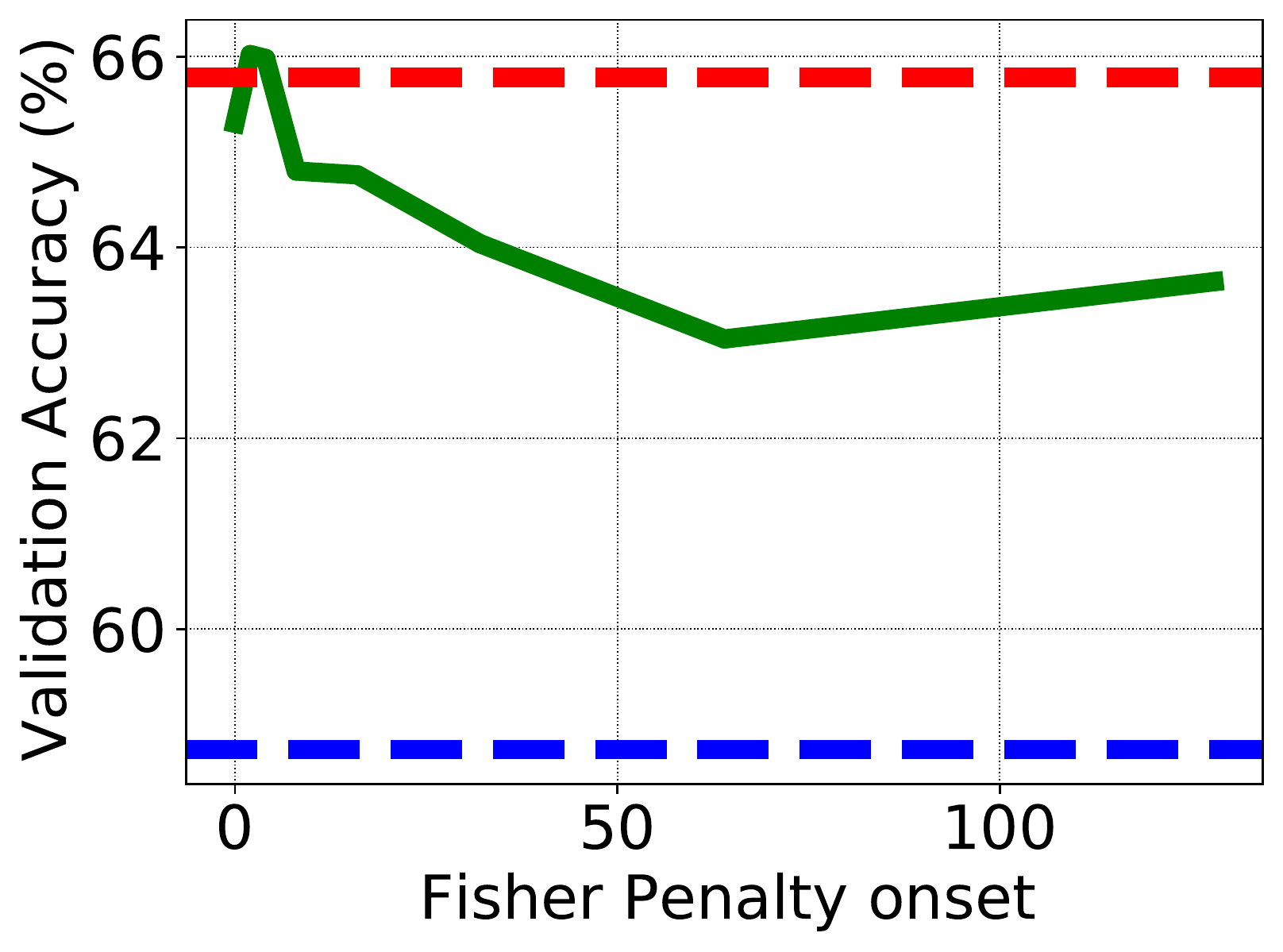}
\caption{DenseNet on CIFAR-100}
    \end{subfigure}
    \caption{Fisher Penalty has to be applied early in training to close the generalization gap to the optimal learning rate (c.f. the red horizontal line to the blue horizontal line). Each subplot summarizes an experiment in which we apply Fisher Penalty starting from a certain epoch (x axis) and measure the final test accuracy (y axis).}
    \label{fig:fisher_penalty_setting1_time}
\end{figure} 
\paragraph{Experimental setting}

We use a similar setting as in the previous section, but we include larger models. We run experiments using Wide ResNet~\citep{zagoruyko2016} (depth 44 and width 3, with or without BN layers), SimpleCNN (without BN layers), DenseNet (L=40, K=12)~\citep{huang_densely_2016} and VGG-11~\citep{Simonyan15}. We train these models on either the CIFAR-10 or the CIFAR-100 datasets. Due to larger computational cost, we replace ImageNet with the TinyImageNet dataset~\citep{Le2015TinyIV} (with images scaled to $32\times32$ resolution) in these experiments. 

 To investigate if the correlation between \TrFi and the final generalization holds more generally, we apply Fisher penalty in two settings. First, we use a learning rate 10-30x smaller than the optimal one, which both incur up to 9\% degradation in test accuracy and results in large value of \TrFi. We also remove data augmentation from the CIFAR-10 and the CIFAR-100 datasets to ensure that training a with small learning rate does not result in underfitting. In the second setting, we add Fisher penalty in training with an optimized learning rate using grid search ($\eta^*$) and train with data augmentation. 

\begin{figure*}
    \centering
    \begin{subfigure}[t]{0.48\textwidth}
    \centering
       \includegraphics[width=1.\columnwidth]{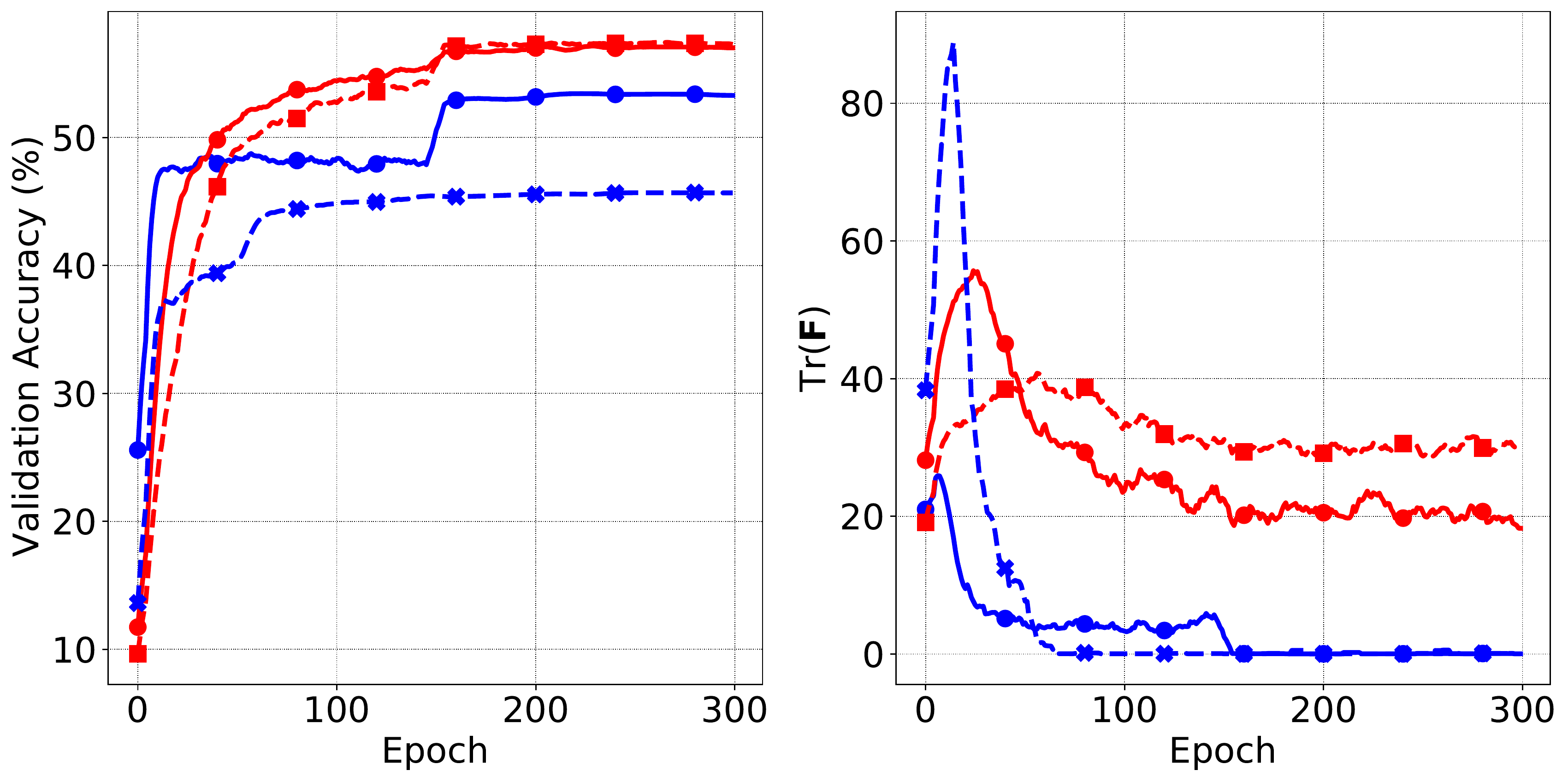}
        \vspace{-0.3cm}
       \includegraphics[width=0.6\columnwidth]{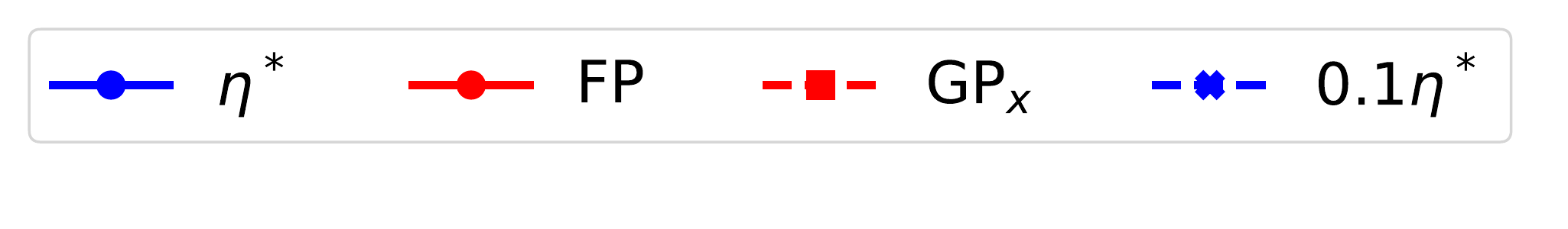}
\caption{Wide ResNet on CIFAR-100 (w/o aug.)}
    \end{subfigure}
    \begin{subfigure}[t]{0.48\textwidth}
    \centering
    \includegraphics[width=1.\columnwidth]{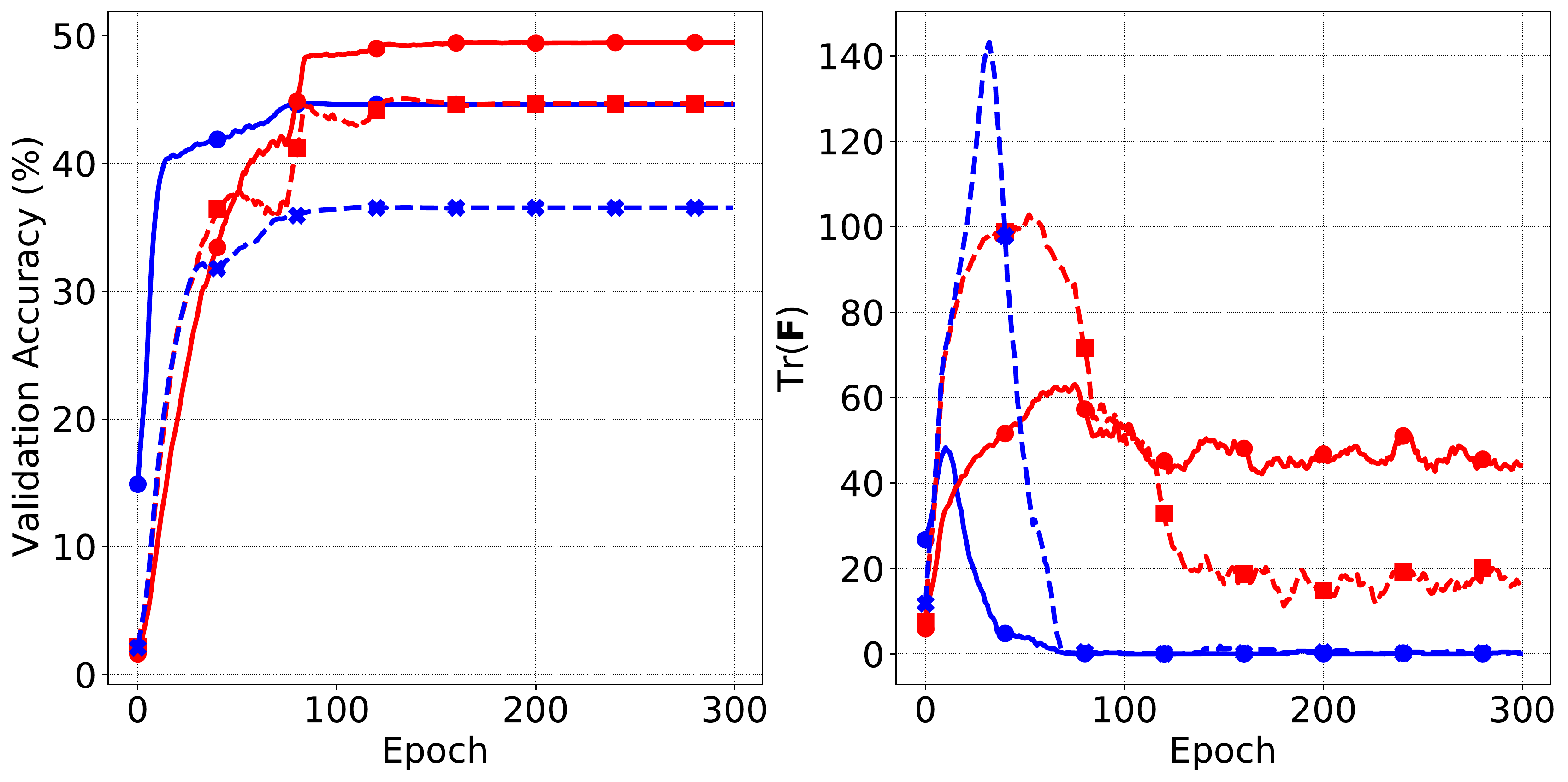}
    \vspace{-0.3cm}
       \includegraphics[width=0.6\columnwidth]{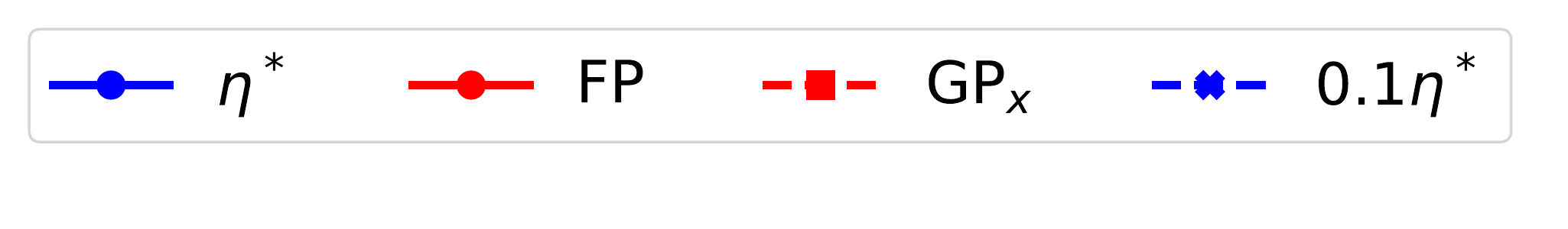}
\caption{VGG-11 on CIFAR-100 (w/o aug.)}
    \end{subfigure}
                \caption{Using Fisher Penalty (\FP) in training with a sub-optimal (small) learning rate drastically reduces the tendency to reach highly curved regions of the loss surface that arises when training with a small learning rate (compare the peak values of \TrF). At the same time, \FP also significantly improves generalization. Penalizing the input gradient norm (\GPx) also impacts \TrF but achieves worse generalization. Each subfigure shows validation accuracy (left) and \TrF (right). Curves were smoothed for clarity.}
    \label{fig:fisher_penalty_setting1_visualization}
\end{figure*}

Fisher penalty penalizes the gradient norm computed using labels sampled from $p_{\bm{\theta}}(y|\bm{x})$. We hypothesize that a similar, but weaker, effect can be introduced by other gradient norm regularizers. We compare \FP to: (1) penalizing the input gradient norm $\left\Vert \bm{g}_x \right\Vert^2 = \| \frac{\partial}{\partial \bm{x}} \ell(\bm{x},y) \|^2$, which we denote by \GPx~\citep{varga2018gradient,Rifai2011,Drucker1992}; (2) penalizing the vanilla mini-batch gradient~\cite{Gulrajani2017}, which we denote by \GP; and penalizing the mini-batch gradient computed with random labels $\left\Vert \bm{g_r} \right\Vert^2 = \| \frac{\partial}{\partial \bm{x}} \ell(\bm{x},\hat y) \|^2$ where $\hat y$ is sampled from a uniform distribution over the label set (\GPr). We are not aware of any prior work using \GP or \GPr in supervised training, with the exception of \citet{Alizadeh2020Gradient} where the authors penalized $\ell_1$ norm of gradients to compress the network towards the end of training, and concurrent work of \citet{barrett2020implicit}. We note that regularizing \GPx is related to regularizing the Jacobian input-output of the network~\citep{hoffman2019,chan2020}.

We tune the hyperparameters on the validation set. More specifically for $\alpha$, we test 10 different values spaced uniformly between $10^{-1} \times v$ and $10^{1} \times v$ on a logarithmic scale with $v \in \mathbb{R}_+$. For TinyImageNet, we evaluate 5 values spaced equally on a logarithmic scale. 
We include the remaining experimental details in the Supplement~\ref{app:fisher_penalty_experimental_details}.

\paragraph{Fisher Penalty improves generalization}

Table~\ref{tab:fisher_penalty_setting1} summarizes the results of the main experiment. First, we observe that a suboptimal learning rate (10-30x lower than the optimal) leads to dramatic overfitting. We observe a degradation of up to 9\% in test accuracy, while achieving approximately 100\% training accuracy (see Table~\ref{app:tab:fisher_penalty_setting1_acc} in the Supplement).

Fisher penalty closes the gap in test accuracy between the small and optimal learning rate, and even achieves better performance than the optimal learning rate. A similar performance was observed when minimizing $\left\|g_r\right\|^2$. We will come back to this observation in the next section. 

\GP and \GPx  reduce the early value of \TrF (see Table~\ref{app:tab:fisher_penalty_setting1_TrF} in the Supplement). They, however, generally perform worse than \TrF or \GPr and do not fully close the gap between small and optimal learning rate. We hypothesize they improve generalization by a similar but less direct mechanism than \TrF and \GPr, which we make more precise in Section~\ref{app:relationship_FP_and_gp} in the Supplement. We note that \GP was  proposed in \citet{barrett2020implicit,smith2021on} as the term that SGD implicitly regularizes (see Related Work for more details).

We also investigate whether similar conclusions hold in large batch size training. In experiments with CIFAR-10 and SimpleCNN, we find that we can close the generalization gap due to training with a large batch size by using Fisher Penalty. We provide further details in Supplement~\ref{app:sec:fisher_explosion_holds_in_lb}.

In the second experimental setting, we apply \FP to a network trained with the optimal learning rate $\eta^*$. According to Table~\ref{tab:fisher_penalty_setting2} (see Table~\ref{app:tab:fisher_penalty_setting2_acc} for training accuracies), Fisher Penalty improves generalization in 4 out of 5 settings. The gap between the baseline and \FP is relatively small in 3 out of 5 settings (below 2\%), which is natural given that we already regularize training implicitly by using the optimal $\eta$..

\begin{table}
\centering
\small
\caption{Fisher penalty (\FP) improves generalization in 4 out of 5  settings when applied with the optimal learning rate $\eta^*$ and trained using standard data augmentation. In 3 out of 5 settings the difference between \FP and $\eta^*$ is relatively small (below 2\%), which is expected given that \FP is aimed at reproducing the regularization effect of large $\eta$, and we compare to training with the optimal $\eta^*$. }
\label{tab:fisher_penalty_setting2}
\begin{tabular}{lll}
\toprule
                     Setting &          $\eta^*$ & \FP \\
                     \midrule
                     WResNet / TinyImageNet &           54.70$\pm$0.0\% &  \textbf{60.00$\pm$0.1\%} \\
                     \midrule
 DenseNet / C100 &  \textbf{74.41$\pm$0.5\%} &           74.19$\pm$0.5\% \\
            VGG11 / C100 &           59.82$\pm$1.2\% &  \textbf{65.08$\pm$0.5\%} \\
          WResNet / C100 &           69.48$\pm$0.3\% &  \textbf{71.53$\pm$1.2\%} \\
          \midrule
                         SimpleCNN / C10 &           87.16$\pm$0.2\% &  \textbf{87.52$\pm$0.5\%} \\
\bottomrule
\end{tabular}
\end{table}

\paragraph{Geometry and generalization in the early phase of training}
\label{sec:fisher_penalty_early}

 Here, we investigate if penalizing \TrF early in training matters for the final generalization. A positive answer would further strengthen the link between the early phase of training and implicit regularization effects in SGD. We run experiments on CIFAR-10 and CIFAR-100. 
 
First, we observe that all gradient-norm regularizers reduce the early value of \TrF closer to \TrF achieved when trained with the optimal learning rate $\eta^*$. We show this effect with Wide ResNet and VGG-11 on CIFAR-100 in Figure~\ref{fig:fisher_penalty_setting1_visualization}, and for other experimental settings in the Supplement. We also tabulate the maximum achieved values of \TrF over the optimization trajectory in Supplement~\ref{app:fisher_penalty_additional_results}.

To test the importance of explicitly penalizing \TrF early in training, we start applying it after a certain number of epochs $E \in \{1, 2, 4, 8, 16, 32, 64, 128\}$. We use the best hyperparameter set from the previous experiments. Figure~\ref{fig:fisher_penalty_setting1_time} summarizes the results. For both datasets, we observe a consistent pattern. When \FP is applied starting from a later epoch, final generalization is significantly worse, and the generalization gap arising from a suboptimal learning rate is not closed. Interestingly, there seems to be a benefit in applying \FP after some warm-up period, which might be related to the widely used trick to gradually increase the learning rate in the early phase~\citep{gotmare2018a}.

\subsection{Fisher Penalty Reduces Memorization}
\label{sec:fisher_penalty_prevents_memorization}

It is not self-evident why regularizing \TrF should influence generalization. In this section, we show that explicit penalization of \TrF improves learning with datasets with noisy labels. To study this, we replace labels of the examples in the CIFAR-100 dataset (25\% or 50\% of the training set) with labels sampled uniformly. We refer to these examples as \emph{noisy examples}. While label noise in real datasets is not uniform, methods that perform well with uniform label noise generally are more robust to label noise in real datasets~\citep{jiang2020}. We also know that datasets such as CIFAR-100 contain many labeling errors~\citep{song2020robust}. Hence, examining if \TrF reduces memorization of synthetic label noise provides an insight into why it improves generalization in our prior experiments.

We argue \FP should reduce memorization. Under certain assumptions, \TrF is equivalent to the norm of the noisy examples' gradient. As such, we would expect \TrF to decrease the norm of the gradient of the noisy examples compared to the gradient of the clean examples. Assuming that the gradient norm of a given group of examples is related to its learning speed~\citep{Chatterjee2020Coherent,fort_stiffness_2019}, \FP should promote learning first clean examples, before learning noisy examples. We make this argument more precise in the Supplement~\ref{app:FP_and_memorization}.

To study whether the above happens in practice, we compare \FP to \GPx, \GPr, and mixup~\citep{zhang2018mixup}. While mixup is not the state-of-the-art approach for learning with noisy labels, it is competitive among approaches that do not require additional data nor multiple stages of training. In particular, it is a component in several state-of-the-art methods~\citep{Li2020DivideMix,song2020robust}. For gradient norm based regularizers, we evaluate 6 different hyperparameter values spaced uniformly on a logarithmic scale, and for mixup we evaluate $\beta \in \{0.2, 0.4, 0.8, 1.6, 3.2, 6.4\}$. We experiment with the Wide ResNet and VGG-11 models. We describe remaining experimental details in  Supplement~\ref{app:fisher_penalty_prevents_memorization_experimental_details}.

\begin{table*}
\centering
\caption{Fisher Penalty (\FP) and \GPr both reduce memorization competitively to mixup. We measure test accuracy at the best validation point in training with either 25\% or 50\% examples with noisy labels in the CIFAR-100 dataset.}
\label{tab:fisher_penalty_noisy_data}
\begin{tabular}{lllll|ll}
\toprule
Label Noise &      Setting & Baseline &    Mixup & \GPx & \FP & \GPr \\
\midrule
 25\% &  VGG-11 / CIFAR-100 &  41.74\% &  52.31\% &                        45.94\% &  \textbf{60.18\%} &                        58.46\% \\
 &  ResNet-52 / CIFAR-100 &  53.30\% &  \textbf{61.61\%} &                        52.70\% &       58.31\% &                        57.60\% \\
 \midrule 
 50\% &  VGG-11 / CIFAR-100 &  30.05\% &  39.15\% &                        34.26\% &  \textbf{51.33\%} &                        50.33\% \\
 &  ResNet-52 / CIFAR-100 &  43.35\% &  \textbf{51.71\%} &                        42.99\% &       47.99\% &                        50.08\% \\
\bottomrule
\end{tabular}
\end{table*}

\paragraph{Results}

\begin{figure*}[t]
\centering
  \begin{subfigure}[t]{0.9\textwidth}
  \centering
    \includegraphics[width=1.0\columnwidth]{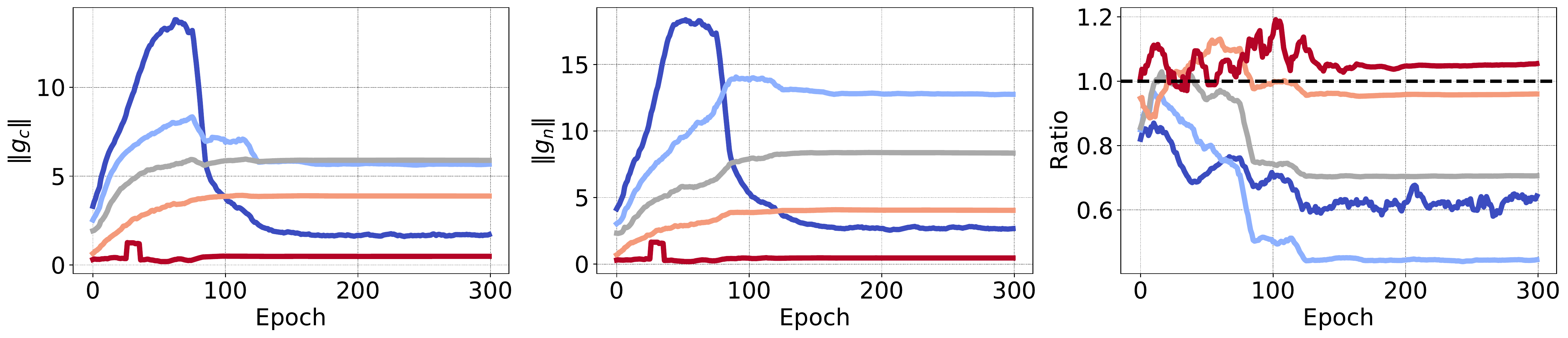}
\caption{Gradient norm on clean examples (left), noisy examples (middle), and their ratio (right); evaluated on the training set.}
     \end{subfigure}
        \begin{subfigure}[t]{0.9\textwidth}
    \centering
 \includegraphics[width=1.0\columnwidth]{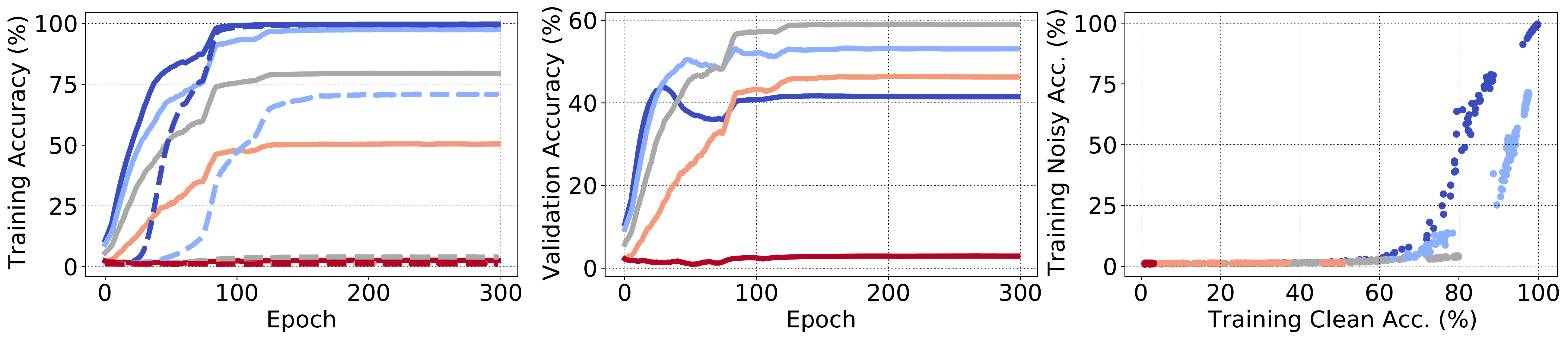}
 \caption{Training accuracy on clean and noisy examples (solid/dashed lined, left), validation accuracy (middle), and a scatter plot of training accuracy on clean vs noisy examples (right). }
 \end{subfigure}
 \caption{Fisher penalty reduces memorization by slowing down training on data with noisy labels more strongly than does training on clean data, as indicated by relative differences in gradient norm (top) and training accuracy (bottom) between the two groups of examples. Experiment run with VGG-11 on the CIFAR-100 dataset. Blue to red color represents increasing regularization coefficient (from $10^{-2}$ to $10^1$). }
 \label{fig:fisher_penalty_noisy_data_visualization}
\end{figure*}

To test whether \FP reduces the speed with which the noisy examples are learned, we track the training and validation accuracy, and the gradient norm. We compute these metrics separately on the noisy and clean examples. Figure~\ref{fig:fisher_penalty_noisy_data_visualization} summarizes the results for VGG-11, and we show results for ResNet-32 in Figure~\ref{app:fig:fisher_penalty_noisy_data_visualization} in the Supplement. 

We observe that \FP limits the ability of the model to memorize data more strongly than it limits its ability to learn from clean data. Figure~\ref{fig:fisher_penalty_noisy_data_visualization} confirms applying \FP results in training accuracy on noisy examples being lower for the same accuracy on clean examples, compared to the baseline.

We can further confirm our interpretation of the effect \TrF has on training by studying the gradient norm. As visible in the top panel of Figure~\ref{fig:fisher_penalty_noisy_data_visualization}, the gradient norm evaluated on noisy examples is larger than on clean examples, and the ratio is closer to 1 when \FP is applied with a larger coefficient. Interestingly, we also observe that the angle between the noisy and clean examples' gradients is negative early in training, which we plot in Figure~\ref{app:fig:fisher_penalty_angle} in the Supplement.

Finally, we summarize test accuracies (at the best validation epoch) in Table~\ref{tab:fisher_penalty_noisy_data}. Penalizing \TrF reduces memorization competitively to mixup, improving test accuracy by up to 21.28\%. Furthermore, \FP strongly outperforms penalizing \GPx, consistent with with trends in the prior Sections. We also again observe \FP to perform comparably to penalizing \GPr. Under the assumption that the model predictive distribution is close to random, \FP is equivalent to penalizing \GPr. Assuming this holds in the early phase of training, we interpret this as a further corroboration of the importance of applying \FP in the early phase (c.f. Section~\ref{sec:fisher_penalty_early}). In the Supplement, we also report training accuracy separately on the clean and noisy examples for the experiment with 25\% noisy examples. 

Taken together, the results suggest implicit or explicit regularization of \FP improves generalization at least in part by strengthening the bias of SGD to learn clean examples before learning noisy examples~\citep{arpit_closer_2017}. We also note that related conclusions about the effect using large learning rates were reached by~\citet{jastrzebski_three_2017,li2019}.

\section{Early \TrF influences final curvature}
\label{sec_trf_final_minima}

To provide further insight into why it is important to regularize \TrF during the early phase of training, we establish a connection between the early phase of training and the wide minima hypothesis~\citep{hochreiter1997flat,keskar_large-batch_2017} which states that \emph{flat} minima \textit{typically} correspond to better generalization. Here, we use \TrH as a measure of flatness. 

\paragraph{Experimental setting} 

We investigate how likely it is for an optimization trajectory to end up in a wide minimum in two scenarios: 1) when optimization exhibits small \TrF early on, and 2) when optimization exhibits large \TrF early on. We train two  ResNet-26 models for 20 epochs using high and low regularization configurations. At epoch 20 we record \TrF for each model. We then use these two models as initialization for 8 separate models each, and continue training using the low regularization configuration with different random seeds. The motivation behind this experiment is to investigate if the degree of regularization in the early phase biases the model towards minima with certain flatness (\TrH) even though no further high regularization configurations are used during the rest of the training. For all these runs, we record the best test accuracy along the optimization trajectory along with \TrH at the point corresponding to the best test accuracy. We describe the remaining experimental details in Supplement \ref{sec_trf_final_minima_details}.

\paragraph{Results} 

We present the result in Figure \ref{fig:histogram_c100} for the CIFAR-100 datasets, and for CIFAR-10 in Supplement \ref{sec_histogram_appendix}. A training run with a lower \TrF during the early phase is more likely to end up in a wider minimum as opposed to one that reaches large \TrF during the early phase. This happens despite that the late phases of both sets of models use the low regularization configuration. The latter runs always end up in sharper minima. In Supplement \ref{sec_trf_final_minima_details} we also show evolution of \TrH throughout training, which suggests that this behavior can be attributed to curvature stabilization happening early during training.

\begin{figure*}
  \centering
        \begin{subfigure}[t]{1\textwidth}
    \includegraphics[width=1.\columnwidth ,trim=0.1in 0.in 0.in 0.in,clip]{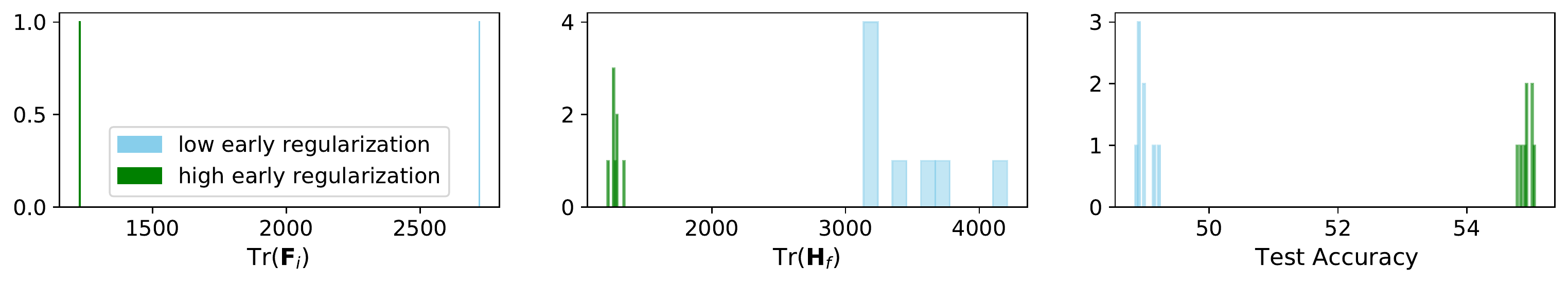}
\end{subfigure}
       \caption{Optimization trajectories passing through regions with low \TrF (\TrFi) during the early phase of training reach wider minima. Two ResNet-56 models are trained with two different levels of regularization for 20 epochs on CIFAR-100. Each model is then continued trained using the low regularization configuration with different random seeds. Left: \TrF at the end of the 20 epochs (\TrFi).  Middle: A histogram of \TrH at the best test accuracy epoch along the trajectory (\TrHf). Right: a histogram of test accuracy.}
    \label{fig:histogram_c100}
\end{figure*}

\section{Related Work}
\label{sec_related_work}

Implicit regularization effects are critical to the empirical success of DNNs~\citep{neyshabur2017,zhang_understanding_2016}. Much of it is attributed to the choice of hyperparameters in SGD \citep{keskar_large-batch_2017,smith2017understanding,li2019}, low complexity bias  induced by gradient descent~\citep{xu2018understanding,jacot2018,arora2019,hu2020surprising}, the cross-entropy loss function \citep{poggio2017theory,soudry2018implicit}, or the importance of the early phase of training~\citep{achille_critical_2017,Jastrzebski2020The,fort_deep_learning,Frankle2020The,golatkar2019,lewkowycz2020large}. However, developing a mechanistic understanding of how SGD implicitly regularizes DNNs remains a largely unsolved problem.

Many prior works have proposed explicit regularizers aimed at finding low curvature solutions~\citep{hochreiter1997flat}. \citet{chaudhari2019entropy} proposed a Langevin dynamics based algorithm. \citet{wen2018smoothout,izmailov2018averaging,foret2021sharpnessaware} propose finding wide minima through approximations involving averaging gradients or parameters at the neighborhood of the current parameter state. In contrast, our focus is on elucidating a mechanistic link behind the implicit regularization effects in SGD and the early phase of training. To corroborate our hypothesis, we develop Fisher Penalty, an efficient and novel explicit regularizer that aims to reproduce the regularization effect of training with a large learning rate. 

Our work contributes to a better understanding of the role of the FIM in training and generalization of deep neural networks. The FIM was also used to define complexity measures such as the Fisher-Rao norm~\citep{karakida2019,liang2019}, and can be seen as approximating the local curvature of the loss surface~\citep{martens2014,thomas2020interplay}. Most notably, the FIM defines the distance metric used in natural gradient descent~\citep{amari1998}.

Penalizing \TrF is related to penalizing the input gradient norm or the input-output Jacobian of the network, which were shown to be effective regularizers for deep neural networks~\citep{Drucker1992,varga2018gradient,hoffman2019,chan2020, moosavi2019robustness}. Most closely related is \citet{moosavi2019robustness} who regularize the curvature in the input space using a stochastic approximation scheme.  Our results suggest that penalizing \FP is a more effective regularizer than other tested variants of gradient norm penalities. However, we did not run an extensive comparison. In particular, we did not compare directly to \citet{moosavi2019robustness}.

\citet{Chatterjee2020Coherent,fort_stiffness_2019} show that SGD avoids memorization by extracting commonalities between examples due to following gradient descent directions shared between examples. Other works have also connected the implicit regularization effects of using large learning rates with preventing memorization~\citep{arpit_closer_2017,li2019,jastrzebski_three_2017}. \citet{liu2020early} was first to note the difference in gradient norms between noisy and clean examples that emerges in the early phase of training. Our work is complementary to these findings. We make a direct connection between memorization, the instability in the early phase of training, and implicit regularization effects arising from using large learning rates in SGD.

Concurrent works have also proposed that implicit regularization effects in SGD can be understood as a form of gradient penalty. \citet{barrett2020implicit,smith2021on} show that SGD implicitly penalizes the mini-batch gradient norm, with the strength controlled by the learning rate, and study \GP as an explicit regularizer. Analogously, we argue that SGD implicitly regularizes \TrF, a closely related quantity, which can be expressed as the squared gradient norm under labels sampled from $p_{\bm{\theta}}(y|\bm{x})$. We found that penalizing \GP did not always close the generalization gap due to using a small learning rate. Overall, our results suggest that the fact that optimization is \emph{close to instability}~\citep{Jastrzebski2020The} in the early phase is critical for understanding implicit regularization effects in SGD. 

\section{Conclusion}

The dramatic instability and changes in the curvature that happen in the early phase of training motivated us to probe its importance for generalization~\citep{Jastrzebski2020The,cohen2021gradient}. We investigated if these effects might explain some of the implicit regularization effects attributed to SGD such as the poorly understood generalization benefit of using large learning rates.

We showed evidence that using a large learning rate in SGD influences generalization by implicitly penalizing the trace of the Fisher Information Matrix (\TrF), a measure of the local curvature, from the beginning of training. We argued that (1) the value of early \TrF correlates with final generalization, and (2) explicitly regularizing \TrF can substantially improve generalization, and showed similar results for training with a small batch size. In the absence of implicit or explicit regularization, \TrF can attain  large values early in training, which we referred to as catastrophic Fisher explosion.

To better understand the mechanism by which penalizing \TrF improves generalization, we investigated training on a dataset with incorrectly labeled examples. Our key finding is that penalizing \TrF significantly reduces memorization by slowing down learning examples with incorrect labels. We hypothesize that penalizing the local curvature (by using large learning rates or penalizing \TrF) early in training improves generalization at least in part by strengthening the implicit bias of SGD to avoid learning noisy labels~\citep{arpit_closer_2017,li2019,liu2020early}. 

Developing theory that is fully consistent with our findings is an interesting topic for the future. Notably, existing theoretical works generally do not connect implicit regularization effects in SGD to the fact optimization is \emph{close to instability} in the early phase of training~\citep{li_stochastic_2017,chaudhari2018stochastic,smith2021on}. 

Another exciting topic for the future is connecting these findings to shortcut learning~\citep{Geirhos_2020}. The tendency of SGD to learn the simplest patterns in the datasets can be detrimental to the broader generalization of the model~\citep{nam2020learning}. We hope that by better understanding implicit regularization effects in SGD, our work will contribute to developing optimization methods that better optimize for both in and out of distribution generalization.

\section*{Limitations}

Our work has several limitations. We used mini-batch gradients to approximate \TrF in Fisher Penalty. While experiments suggest it is inconsequential for the main conclusions, more work would be needed to fully establish it.  We also experimented only with vision tasks. Finally, we observed that \FP slows down learning on clean examples. This feature makes it challenging to apply with very small learning rates, at which underfitting might start to be an issue.

\section*{Acknowledgments}
GK acknowledges support in part by the FRQNT Strategic Clusters Program (2020-RS4-265502 - Centre UNIQUE - Union Neurosciences \& Artificial Intelligence - Quebec).
KC was supported by Samsung Advanced Institute of Technology (Next Generation Deep Learning:
from pattern recognition to AI) and Samsung Research (Improving Deep Learning using Latent Structure). We thank Catriona C. Geras for proofreading the paper.

\bibliography{ref}

\begin{thebibliography}{77}
\providecommand{\natexlab}[1]{#1}
\providecommand{\url}[1]{\texttt{#1}}
\expandafter\ifx\csname urlstyle\endcsname\relax
  \providecommand{\doi}[1]{doi: #1}\else
  \providecommand{\doi}{doi: \begingroup \urlstyle{rm}\Url}\fi

\bibitem[Achille et~al.(2019)Achille, Rovere, and
  Soatto]{achille_critical_2017}
Alessandro Achille, Matteo Rovere, and Stefano Soatto.
\newblock Critical learning periods in deep networks.
\newblock In \emph{7th International Conference on Learning Representations,
  {ICLR} 2019, New Orleans, LA, USA, May 6-9, 2019}, 2019.

\bibitem[Alizadeh et~al.(2020)Alizadeh, Behboodi, van Baalen, Louizos,
  Blankevoort, and Welling]{Alizadeh2020Gradient}
Milad Alizadeh, Arash Behboodi, Mart van Baalen, Christos Louizos, Tijmen
  Blankevoort, and Max Welling.
\newblock Gradient $\ell_1$ regularization for quantization robustness.
\newblock In \emph{International Conference on Learning Representations}, 2020.

\bibitem[Amari(1998)]{amari1998}
Shun-Ichi Amari.
\newblock Natural gradient works efficiently in learning.
\newblock \emph{Neural Computation}, 1998.

\bibitem[Arora et~al.(2019)Arora, Cohen, Hu, and Luo]{arora2019}
Sanjeev Arora, Nadav Cohen, Wei Hu, and Yuping Luo.
\newblock Implicit regularization in deep matrix factorization.
\newblock In \emph{Advances in Neural Information Processing Systems 32: Annual
  Conference on Neural Information Processing Systems 2019, NeurIPS 2019,
  December 8-14, 2019, Vancouver, BC, Canada}, 2019.

\bibitem[Arpit et~al.(2017)Arpit, Jastrzebski, Ballas, Krueger, Bengio, Kanwal,
  Maharaj, Fischer, Courville, Bengio, and Lacoste{-}Julien]{arpit_closer_2017}
Devansh Arpit, Stanislaw Jastrzebski, Nicolas Ballas, David Krueger, Emmanuel
  Bengio, Maxinder~S. Kanwal, Tegan Maharaj, Asja Fischer, Aaron~C. Courville,
  Yoshua Bengio, and Simon Lacoste{-}Julien.
\newblock A closer look at memorization in deep networks.
\newblock In \emph{Proceedings of the 34th International Conference on Machine
  Learning, {ICML} 2017, Sydney, NSW, Australia, 6-11 August 2017}, Proceedings
  of Machine Learning Research, 2017.

\bibitem[Arpit et~al.(2019)Arpit, Campos, and Bengio]{arpit2019initialize}
Devansh Arpit, V{\'{\i}}ctor Campos, and Yoshua Bengio.
\newblock How to initialize your network? robust initialization for weightnorm
  {\&} resnets.
\newblock In \emph{Advances in Neural Information Processing Systems 32: Annual
  Conference on Neural Information Processing Systems 2019, NeurIPS 2019,
  December 8-14, 2019, Vancouver, BC, Canada}, 2019.

\bibitem[Barrett \& Dherin(2021)Barrett and Dherin]{barrett2020implicit}
David Barrett and Benoit Dherin.
\newblock Implicit gradient regularization.
\newblock In \emph{International Conference on Learning Representations}, 2021.

\bibitem[Bjorck et~al.(2018)Bjorck, Gomes, Selman, and
  Weinberger]{bjorck_understanding_2018}
Johan Bjorck, Carla~P. Gomes, Bart Selman, and Kilian~Q. Weinberger.
\newblock Understanding batch normalization.
\newblock In \emph{Advances in Neural Information Processing Systems 31: Annual
  Conference on Neural Information Processing Systems 2018, NeurIPS 2018,
  December 3-8, 2018, Montr{\'{e}}al, Canada}, 2018.

\bibitem[Chan et~al.(2020)Chan, Tay, Ong, and Fu]{chan2020}
Alvin Chan, Yi~Tay, Yew{-}Soon Ong, and Jie Fu.
\newblock Jacobian adversarially regularized networks for robustness.
\newblock In \emph{8th International Conference on Learning Representations,
  {ICLR} 2020, Addis Ababa, Ethiopia, April 26-30, 2020}, 2020.

\bibitem[Chatterjee(2020)]{Chatterjee2020Coherent}
Satrajit Chatterjee.
\newblock Coherent gradients: An approach to understanding generalization in
  gradient descent-based optimization.
\newblock In \emph{8th International Conference on Learning Representations,
  {ICLR} 2020, Addis Ababa, Ethiopia, April 26-30, 2020}, 2020.

\bibitem[Chaudhari \& Soatto(2018)Chaudhari and
  Soatto]{chaudhari2018stochastic}
Pratik Chaudhari and Stefano Soatto.
\newblock Stochastic gradient descent performs variational inference, converges
  to limit cycles for deep networks.
\newblock In \emph{6th International Conference on Learning Representations,
  {ICLR} 2018, Vancouver, BC, Canada, April 30 - May 3, 2018, Conference Track
  Proceedings}, 2018.

\bibitem[Chaudhari et~al.(2017)Chaudhari, Choromanska, Soatto, LeCun, Baldassi,
  Borgs, Chayes, Sagun, and Zecchina]{chaudhari2019entropy}
Pratik Chaudhari, Anna Choromanska, Stefano Soatto, Yann LeCun, Carlo Baldassi,
  Christian Borgs, Jennifer~T. Chayes, Levent Sagun, and Riccardo Zecchina.
\newblock Entropy-sgd: Biasing gradient descent into wide valleys.
\newblock In \emph{5th International Conference on Learning Representations,
  {ICLR} 2017, Toulon, France, April 24-26, 2017, Conference Track
  Proceedings}, 2017.

\bibitem[Chollet \& {others}(2015)Chollet and {others}]{chollet_keras_2015}
François Chollet and {others}.
\newblock \emph{Keras}.
\newblock 2015.

\bibitem[Cohen et~al.(2021)Cohen, Kaur, Li, Kolter, and
  Talwalkar]{cohen2021gradient}
Jeremy Cohen, Simran Kaur, Yuanzhi Li, J~Zico Kolter, and Ameet Talwalkar.
\newblock Gradient descent on neural networks typically occurs at the edge of
  stability.
\newblock In \emph{International Conference on Learning Representations}, 2021.

\bibitem[De \& Smith(2020)De and Smith]{de2020batch}
Soham De and Samuel~L. Smith.
\newblock Batch normalization biases residual blocks towards the identity
  function in deep networks.
\newblock In \emph{Advances in Neural Information Processing Systems 33: Annual
  Conference on Neural Information Processing Systems 2020, NeurIPS 2020,
  December 6-12, 2020, virtual}, 2020.

\bibitem[Deng et~al.(2009)Deng, Dong, Socher, Li, Li, and
  Li]{deng_imagenet_2009}
Jia Deng, Wei Dong, Richard Socher, Li{-}Jia Li, Kai Li, and Fei{-}Fei Li.
\newblock Imagenet: {A} large-scale hierarchical image database.
\newblock In \emph{2009 {IEEE} Computer Society Conference on Computer Vision
  and Pattern Recognition {(CVPR} 2009), 20-25 June 2009, Miami, Florida,
  {USA}}, 2009.

\bibitem[Dinh et~al.(2017)Dinh, Pascanu, Bengio, and Bengio]{dinh_sharp_2017}
Laurent Dinh, Razvan Pascanu, Samy Bengio, and Yoshua Bengio.
\newblock Sharp minima can generalize for deep nets.
\newblock In \emph{Proceedings of the 34th International Conference on Machine
  Learning, {ICML} 2017, Sydney, NSW, Australia, 6-11 August 2017}, Proceedings
  of Machine Learning Research, 2017.

\bibitem[{Drucker} \& {Le Cun}(1992){Drucker} and {Le Cun}]{Drucker1992}
H.~{Drucker} and Y.~{Le Cun}.
\newblock Improving generalization performance using double backpropagation.
\newblock \emph{IEEE Transactionsf on Neural Networks}, 1992.

\bibitem[Foret et~al.(2021)Foret, Kleiner, Mobahi, and
  Neyshabur]{foret2021sharpnessaware}
Pierre Foret, Ariel Kleiner, Hossein Mobahi, and Behnam Neyshabur.
\newblock Sharpness-aware minimization for efficiently improving
  generalization.
\newblock In \emph{International Conference on Learning Representations}, 2021.

\bibitem[Fort et~al.(2020{\natexlab{a}})Fort, Dziugaite, Paul, Kharaghani, Roy,
  and Ganguli]{fort_deep_learning}
Stanislav Fort, Gintare~Karolina Dziugaite, Mansheej Paul, Sepideh Kharaghani,
  Daniel~M. Roy, and Surya Ganguli.
\newblock Deep learning versus kernel learning: an empirical study of loss
  landscape geometry and the time evolution of the neural tangent kernel.
\newblock In \emph{Advances in Neural Information Processing Systems 33: Annual
  Conference on Neural Information Processing Systems 2020, NeurIPS 2020,
  December 6-12, 2020, virtual}, 2020{\natexlab{a}}.

\bibitem[Fort et~al.(2020{\natexlab{b}})Fort, Nowak, Jastrzebski, and
  Narayanan]{fort_stiffness_2019}
Stanislav Fort, Paweł~Krzysztof Nowak, Stanislaw Jastrzebski, and Srini
  Narayanan.
\newblock Stiffness: A new perspective on generalization in neural networks,
  2020{\natexlab{b}}.

\bibitem[Frankle et~al.(2020)Frankle, Schwab, and Morcos]{Frankle2020The}
Jonathan Frankle, David~J. Schwab, and Ari~S. Morcos.
\newblock The early phase of neural network training.
\newblock In \emph{8th International Conference on Learning Representations,
  {ICLR} 2020, Addis Ababa, Ethiopia, April 26-30, 2020}, 2020.

\bibitem[Geirhos et~al.(2020)Geirhos, Jacobsen, Michaelis, Zemel, Brendel,
  Bethge, and Wichmann]{Geirhos_2020}
Robert Geirhos, Jörn-Henrik Jacobsen, Claudio Michaelis, Richard Zemel,
  Wieland Brendel, Matthias Bethge, and Felix~A. Wichmann.
\newblock Shortcut learning in deep neural networks.
\newblock \emph{Nature Machine Intelligence}, 2020.

\bibitem[Golatkar et~al.(2019)Golatkar, Achille, and Soatto]{golatkar2019}
Aditya Golatkar, Alessandro Achille, and Stefano Soatto.
\newblock Time matters in regularizing deep networks: Weight decay and data
  augmentation affect early learning dynamics, matter little near convergence.
\newblock In \emph{Advances in Neural Information Processing Systems 32: Annual
  Conference on Neural Information Processing Systems 2019, NeurIPS 2019,
  December 8-14, 2019, Vancouver, BC, Canada}, 2019.

\bibitem[Gotmare et~al.(2019)Gotmare, Keskar, Xiong, and Socher]{gotmare2018a}
Akhilesh Gotmare, Nitish~Shirish Keskar, Caiming Xiong, and Richard Socher.
\newblock A closer look at deep learning heuristics: Learning rate restarts,
  warmup and distillation.
\newblock In \emph{7th International Conference on Learning Representations,
  {ICLR} 2019, New Orleans, LA, USA, May 6-9, 2019}, 2019.

\bibitem[Gulrajani et~al.(2017)Gulrajani, Ahmed, Arjovsky, Dumoulin, and
  Courville]{Gulrajani2017}
Ishaan Gulrajani, Faruk Ahmed, Mart{\'{\i}}n Arjovsky, Vincent Dumoulin, and
  Aaron~C. Courville.
\newblock Improved training of wasserstein gans.
\newblock In \emph{Advances in Neural Information Processing Systems 30: Annual
  Conference on Neural Information Processing Systems 2017, December 4-9, 2017,
  Long Beach, CA, {USA}}, 2017.

\bibitem[Gur-Ari et~al.(2018)Gur-Ari, Roberts, and Dyer]{gur-ari_gradient_2018}
Guy Gur-Ari, Daniel~A. Roberts, and Ethan Dyer.
\newblock Gradient descent happens in a tiny subspace, 2018.

\bibitem[He et~al.(2019)He, Huang, and Yuan]{He2019}
Haowei He, Gao Huang, and Yang Yuan.
\newblock Asymmetric valleys: Beyond sharp and flat local minima.
\newblock In \emph{Advances in Neural Information Processing Systems 32: Annual
  Conference on Neural Information Processing Systems 2019, NeurIPS 2019,
  December 8-14, 2019, Vancouver, BC, Canada}, 2019.

\bibitem[He et~al.(2016)He, Zhang, Ren, and Sun]{he_deep_2015}
Kaiming He, Xiangyu Zhang, Shaoqing Ren, and Jian Sun.
\newblock Deep residual learning for image recognition.
\newblock In \emph{2016 {IEEE} Conference on Computer Vision and Pattern
  Recognition, {CVPR} 2016, Las Vegas, NV, USA, June 27-30, 2016}, 2016.

\bibitem[Hochreiter \& Schmidhuber(1997)Hochreiter and
  Schmidhuber]{hochreiter1997flat}
Sepp Hochreiter and J{\"u}rgen Schmidhuber.
\newblock Flat minima.
\newblock \emph{Neural Computation}, 1997.

\bibitem[Hoffman et~al.(2019)Hoffman, Roberts, and Yaida]{hoffman2019}
Judy Hoffman, Daniel~A. Roberts, and Sho Yaida.
\newblock Robust learning with jacobian regularization.
\newblock 2019.

\bibitem[Hu et~al.(2020)Hu, Xiao, Adlam, and Pennington]{hu2020surprising}
Wei Hu, Lechao Xiao, Ben Adlam, and Jeffrey Pennington.
\newblock The surprising simplicity of the early-time learning dynamics of
  neural networks.
\newblock In \emph{Advances in Neural Information Processing Systems 33: Annual
  Conference on Neural Information Processing Systems 2020, NeurIPS 2020,
  December 6-12, 2020, virtual}, 2020.

\bibitem[Huang et~al.(2017)Huang, Liu, van~der Maaten, and
  Weinberger]{huang_densely_2016}
Gao Huang, Zhuang Liu, Laurens van~der Maaten, and Kilian~Q. Weinberger.
\newblock Densely connected convolutional networks.
\newblock In \emph{2017 {IEEE} Conference on Computer Vision and Pattern
  Recognition, {CVPR} 2017, Honolulu, HI, USA, July 21-26, 2017}, 2017.

\bibitem[Hutchinson(1990)]{hutchinson1990stochastic}
Michael~F Hutchinson.
\newblock A stochastic estimator of the trace of the influence matrix for
  laplacian smoothing splines.
\newblock \emph{Communications in Statistics-Simulation and Computation}, 1990.

\bibitem[Izmailov et~al.(2018)Izmailov, Podoprikhin, Garipov, Vetrov, and
  Wilson]{izmailov2018averaging}
Pavel Izmailov, Dmitrii Podoprikhin, Timur Garipov, Dmitry~P. Vetrov, and
  Andrew~Gordon Wilson.
\newblock Averaging weights leads to wider optima and better generalization.
\newblock In \emph{Proceedings of the Thirty-Fourth Conference on Uncertainty
  in Artificial Intelligence, {UAI} 2018, Monterey, California, USA, August
  6-10, 2018}, 2018.

\bibitem[Jacot et~al.(2018)Jacot, Hongler, and Gabriel]{jacot2018}
Arthur Jacot, Cl{\'{e}}ment Hongler, and Franck Gabriel.
\newblock Neural tangent kernel: Convergence and generalization in neural
  networks.
\newblock In \emph{Advances in Neural Information Processing Systems 31: Annual
  Conference on Neural Information Processing Systems 2018, NeurIPS 2018,
  December 3-8, 2018, Montr{\'{e}}al, Canada}, 2018.

\bibitem[Jastrzebski et~al.(2017)Jastrzebski, Kenton, Arpit, Ballas, Fischer,
  Bengio, and Storkey]{jastrzebski_three_2017}
Stanislaw Jastrzebski, Zachary Kenton, Devansh Arpit, Nicolas Ballas, Asja
  Fischer, Yoshua Bengio, and Amos~J. Storkey.
\newblock Three {Factors} {Influencing} {Minima} in {SGD}.
\newblock 2017.

\bibitem[Jastrzebski et~al.(2019)Jastrzebski, Kenton, Ballas, Fischer, Bengio,
  and Storkey]{jastrzebski_relation_2018}
Stanislaw Jastrzebski, Zachary Kenton, Nicolas Ballas, Asja Fischer, Yoshua
  Bengio, and Amos~J. Storkey.
\newblock On the relation between the sharpest directions of {DNN} loss and the
  {SGD} step length.
\newblock In \emph{7th International Conference on Learning Representations,
  {ICLR} 2019, New Orleans, LA, USA, May 6-9, 2019}, 2019.

\bibitem[Jastrzebski et~al.(2020)Jastrzebski, Szymczak, Fort, Arpit, Tabor,
  Cho, and Geras]{Jastrzebski2020The}
Stanislaw Jastrzebski, Maciej Szymczak, Stanislav Fort, Devansh Arpit, Jacek
  Tabor, Kyunghyun Cho, and Krzysztof Geras.
\newblock The break-even point on optimization trajectories of deep neural
  networks.
\newblock In \emph{8th International Conference on Learning Representations,
  {ICLR} 2020, Addis Ababa, Ethiopia, April 26-30, 2020}, 2020.

\bibitem[Jiang et~al.(2020{\natexlab{a}})Jiang, Huang, Liu, and
  Yang]{jiang2020}
Lu~Jiang, Di~Huang, Mason Liu, and Weilong Yang.
\newblock Beyond synthetic noise: Deep learning on controlled noisy labels.
\newblock In \emph{Proceedings of the 37th International Conference on Machine
  Learning, {ICML} 2020, 13-18 July 2020, Virtual Event}, Proceedings of
  Machine Learning Research, 2020{\natexlab{a}}.

\bibitem[Jiang et~al.(2020{\natexlab{b}})Jiang, Neyshabur, Mobahi, Krishnan,
  and Bengio]{jiang_fantastic_2020}
Yiding Jiang, Behnam Neyshabur, Hossein Mobahi, Dilip Krishnan, and Samy
  Bengio.
\newblock Fantastic generalization measures and where to find them.
\newblock In \emph{8th International Conference on Learning Representations,
  {ICLR} 2020, Addis Ababa, Ethiopia, April 26-30, 2020}, 2020{\natexlab{b}}.

\bibitem[Karakida et~al.(2019)Karakida, Akaho, and Amari]{karakida2019}
Ryo Karakida, Shotaro Akaho, and Shun{-}ichi Amari.
\newblock Universal statistics of fisher information in deep neural networks:
  Mean field approach.
\newblock In \emph{The 22nd International Conference on Artificial Intelligence
  and Statistics, {AISTATS} 2019, 16-18 April 2019, Naha, Okinawa, Japan},
  Proceedings of Machine Learning Research, 2019.

\bibitem[Keskar et~al.(2017)Keskar, Mudigere, Nocedal, Smelyanskiy, and
  Tang]{keskar_large-batch_2017}
Nitish~Shirish Keskar, Dheevatsa Mudigere, Jorge Nocedal, Mikhail Smelyanskiy,
  and Ping Tak~Peter Tang.
\newblock On large-batch training for deep learning: Generalization gap and
  sharp minima.
\newblock In \emph{5th International Conference on Learning Representations,
  {ICLR} 2017, Toulon, France, April 24-26, 2017, Conference Track
  Proceedings}, 2017.

\bibitem[Krizhevsky(2009)]{krizhevsky_learning_2009}
Alex Krizhevsky.
\newblock Learning multiple layers of features from tiny images.
\newblock Technical report, 2009.

\bibitem[Kunstner et~al.(2019)Kunstner, Hennig, and
  Balles]{kunstner2019limitations}
Frederik Kunstner, Philipp Hennig, and Lukas Balles.
\newblock Limitations of the empirical fisher approximation for natural
  gradient descent.
\newblock In \emph{Advances in Neural Information Processing Systems 32: Annual
  Conference on Neural Information Processing Systems 2019, NeurIPS 2019,
  December 8-14, 2019, Vancouver, BC, Canada}, 2019.

\bibitem[Le \& Yang(2015)Le and Yang]{Le2015TinyIV}
Y.~Le and X.~Yang.
\newblock Tiny imagenet visual recognition challenge.
\newblock 2015.

\bibitem[Leclerc \& Madry(2020)Leclerc and Madry]{leclerc2020regimes}
Guillaume Leclerc and Aleksander Madry.
\newblock The two regimes of deep network training, 2020.

\bibitem[LeCun et~al.(2012)LeCun, Bottou, Orr, and
  Müller]{lecun_efficient_2012}
Yann LeCun, Léon Bottou, Genevieve~B. Orr, and Klaus-Robert Müller.
\newblock Efficient {BackProp}.
\newblock In \emph{Neural {Networks}: {Tricks} of the {Trade} (2nd ed.)}. 2012.

\bibitem[Lewkowycz et~al.(2020)Lewkowycz, Bahri, Dyer, Sohl-Dickstein, and
  Gur-Ari]{lewkowycz2020large}
Aitor Lewkowycz, Yasaman Bahri, Ethan Dyer, Jascha Sohl-Dickstein, and Guy
  Gur-Ari.
\newblock The large learning rate phase of deep learning: the catapult
  mechanism, 2020.

\bibitem[Li et~al.(2020)Li, Socher, and Hoi]{Li2020DivideMix}
Junnan Li, Richard Socher, and Steven C.~H. Hoi.
\newblock Dividemix: Learning with noisy labels as semi-supervised learning.
\newblock In \emph{8th International Conference on Learning Representations,
  {ICLR} 2020, Addis Ababa, Ethiopia, April 26-30, 2020}, 2020.

\bibitem[Li et~al.(2017)Li, Tai, and E]{li_stochastic_2017}
Qianxiao Li, Cheng Tai, and Weinan E.
\newblock Stochastic modified equations and adaptive stochastic gradient
  algorithms.
\newblock In \emph{Proceedings of the 34th International Conference on Machine
  Learning, {ICML} 2017, Sydney, NSW, Australia, 6-11 August 2017}, Proceedings
  of Machine Learning Research, 2017.

\bibitem[Li et~al.(2019)Li, Wei, and Ma]{li2019}
Yuanzhi Li, Colin Wei, and Tengyu Ma.
\newblock Towards explaining the regularization effect of initial large
  learning rate in training neural networks.
\newblock In \emph{Advances in Neural Information Processing Systems 32: Annual
  Conference on Neural Information Processing Systems 2019, NeurIPS 2019,
  December 8-14, 2019, Vancouver, BC, Canada}, 2019.

\bibitem[Liang et~al.(2019)Liang, Poggio, Rakhlin, and Stokes]{liang2019}
Tengyuan Liang, Tomaso~A. Poggio, Alexander Rakhlin, and James Stokes.
\newblock Fisher-rao metric, geometry, and complexity of neural networks.
\newblock In \emph{The 22nd International Conference on Artificial Intelligence
  and Statistics, {AISTATS} 2019, 16-18 April 2019, Naha, Okinawa, Japan},
  Proceedings of Machine Learning Research, 2019.

\bibitem[Liu et~al.(2020)Liu, Niles-Weed, Razavian, and
  Fernandez-Granda]{liu2020early}
Sheng Liu, Jonathan Niles-Weed, Narges Razavian, and Carlos Fernandez-Granda.
\newblock Early-learning regularization prevents memorization of noisy labels.
\newblock \emph{Advances in Neural Information Processing Systems}, 2020.

\bibitem[Martens(2020)]{martens2014}
James Martens.
\newblock New insights and perspectives on the natural gradient method, 2020.

\bibitem[Moosavi{-}Dezfooli et~al.(2019)Moosavi{-}Dezfooli, Fawzi, Uesato, and
  Frossard]{moosavi2019robustness}
Seyed{-}Mohsen Moosavi{-}Dezfooli, Alhussein Fawzi, Jonathan Uesato, and Pascal
  Frossard.
\newblock Robustness via curvature regularization, and vice versa.
\newblock In \emph{{IEEE} Conference on Computer Vision and Pattern
  Recognition, {CVPR} 2019, Long Beach, CA, USA, June 16-20, 2019}, 2019.

\bibitem[Nam et~al.(2020)Nam, Cha, Ahn, Lee, and Shin]{nam2020learning}
Junhyun Nam, Hyuntak Cha, Sungsoo Ahn, Jaeho Lee, and Jinwoo Shin.
\newblock Learning from failure: Training debiased classifier from biased
  classifier.
\newblock In \emph{Advances in Neural Information Processing Systems}, 2020.

\bibitem[Neyshabur(2017)]{neyshabur2017}
Behnam Neyshabur.
\newblock Implicit regularization in deep learning.
\newblock 2017.

\bibitem[Poggio et~al.(2018)Poggio, Kawaguchi, Liao, Miranda, Rosasco, Boix,
  Hidary, and Mhaskar]{poggio2017theory}
Tomaso Poggio, Kenji Kawaguchi, Qianli Liao, Brando Miranda, Lorenzo Rosasco,
  Xavier Boix, Jack Hidary, and Hrushikesh Mhaskar.
\newblock Theory of deep learning iii: explaining the non-overfitting puzzle,
  2018.

\bibitem[Rahaman et~al.(2019)Rahaman, Baratin, Arpit, Draxler, Lin, Hamprecht,
  Bengio, and Courville]{Rahaman2018OnTS}
Nasim Rahaman, Aristide Baratin, Devansh Arpit, Felix Draxler, Min Lin, Fred~A.
  Hamprecht, Yoshua Bengio, and Aaron~C. Courville.
\newblock On the spectral bias of neural networks.
\newblock In \emph{Proceedings of the 36th International Conference on Machine
  Learning, {ICML} 2019, 9-15 June 2019, Long Beach, California, {USA}},
  Proceedings of Machine Learning Research, 2019.

\bibitem[Rifai et~al.(2011)Rifai, Vincent, Muller, Glorot, and
  Bengio]{Rifai2011}
Salah Rifai, Pascal Vincent, Xavier Muller, Xavier Glorot, and Yoshua Bengio.
\newblock Contractive auto-encoders: Explicit invariance during feature
  extraction.
\newblock In \emph{Proceedings of the 28th International Conference on Machine
  Learning, {ICML} 2011, Bellevue, Washington, USA, June 28 - July 2, 2011},
  2011.

\bibitem[Salakhutdinov(2014)]{goodfellow_deep_2016}
Ruslan Salakhutdinov.
\newblock Deep learning.
\newblock In \emph{The 20th {ACM} {SIGKDD} International Conference on
  Knowledge Discovery and Data Mining, {KDD} '14, New York, NY, {USA} - August
  24 - 27, 2014}, 2014.

\bibitem[Simonyan \& Zisserman(2015)Simonyan and Zisserman]{Simonyan15}
Karen Simonyan and Andrew Zisserman.
\newblock Very deep convolutional networks for large-scale image recognition.
\newblock In \emph{3rd International Conference on Learning Representations,
  {ICLR} 2015, San Diego, CA, USA, May 7-9, 2015, Conference Track
  Proceedings}, 2015.

\bibitem[Smith \& Le(2018)Smith and Le]{smith2017understanding}
Samuel~L. Smith and Quoc~V. Le.
\newblock A bayesian perspective on generalization and stochastic gradient
  descent.
\newblock In \emph{6th International Conference on Learning Representations,
  {ICLR} 2018, Vancouver, BC, Canada, April 30 - May 3, 2018, Conference Track
  Proceedings}, 2018.

\bibitem[Smith et~al.(2021)Smith, Dherin, Barrett, and De]{smith2021on}
Samuel~L Smith, Benoit Dherin, David Barrett, and Soham De.
\newblock On the origin of implicit regularization in stochastic gradient
  descent.
\newblock In \emph{International Conference on Learning Representations}, 2021.

\bibitem[Song et~al.(2020)Song, Dauphin, Auli, and Ma]{song2020robust}
Jiaming Song, Yann Dauphin, Michael Auli, and Tengyu Ma.
\newblock Robust and on-the-fly dataset denoising for image classification.
\newblock In \emph{Computer Vision -- ECCV 2020}, 2020.

\bibitem[Soudry et~al.(2018)Soudry, Hoffer, Nacson, and
  Srebro]{soudry2018implicit}
Daniel Soudry, Elad Hoffer, Mor~Shpigel Nacson, and Nathan Srebro.
\newblock The implicit bias of gradient descent on separable data.
\newblock In \emph{6th International Conference on Learning Representations,
  {ICLR} 2018, Vancouver, BC, Canada, April 30 - May 3, 2018, Conference Track
  Proceedings}, 2018.

\bibitem[Thomas et~al.(2020)Thomas, Pedregosa, van Merri{\"{e}}nboer, Manzagol,
  Bengio, and Roux]{thomas2020interplay}
Valentin Thomas, Fabian Pedregosa, Bart van Merri{\"{e}}nboer, Pierre{-}Antoine
  Manzagol, Yoshua Bengio, and Nicolas~Le Roux.
\newblock On the interplay between noise and curvature and its effect on
  optimization and generalization.
\newblock In \emph{The 23rd International Conference on Artificial Intelligence
  and Statistics, {AISTATS} 2020, 26-28 August 2020, Online [Palermo, Sicily,
  Italy]}, Proceedings of Machine Learning Research, 2020.

\bibitem[Tsuzuku et~al.(2020)Tsuzuku, Sato, and
  Sugiyama]{tsuzuku2019normalized}
Yusuke Tsuzuku, Issei Sato, and Masashi Sugiyama.
\newblock Normalized flat minima: Exploring scale invariant definition of flat
  minima for neural networks using pac-bayesian analysis.
\newblock In \emph{Proceedings of the 37th International Conference on Machine
  Learning, {ICML} 2020, 13-18 July 2020, Virtual Event}, Proceedings of
  Machine Learning Research, 2020.

\bibitem[Varga et~al.(2018)Varga, Csiszárik, and Zombori]{varga2018gradient}
Dániel Varga, Adrián Csiszárik, and Zsolt Zombori.
\newblock Gradient regularization improves accuracy of discriminative models,
  2018.

\bibitem[Wen et~al.(2018)Wen, Wang, Yan, Xu, Wu, Chen, and
  Li]{wen2018smoothout}
Wei Wen, Yandan Wang, Feng Yan, Cong Xu, Chunpeng Wu, Yiran Chen, and Hai Li.
\newblock Smoothout: Smoothing out sharp minima to improve generalization in
  deep learning, 2018.

\bibitem[Xu(2018)]{xu2018understanding}
Zhiqin~John Xu.
\newblock Understanding training and generalization in deep learning by fourier
  analysis, 2018.

\bibitem[Yoshida \& Miyato(2017)Yoshida and Miyato]{yoshida2017spectral}
Yuichi Yoshida and Takeru Miyato.
\newblock Spectral norm regularization for improving the generalizability of
  deep learning, 2017.

\bibitem[Zagoruyko \& Komodakis(2016)Zagoruyko and Komodakis]{zagoruyko2016}
Sergey Zagoruyko and Nikos Komodakis.
\newblock Wide residual networks.
\newblock In \emph{Proceedings of the British Machine Vision Conference 2016,
  {BMVC} 2016, York, UK, September 19-22, 2016}, 2016.

\bibitem[Zhang et~al.(2017)Zhang, Bengio, Hardt, Recht, and
  Vinyals]{zhang_understanding_2016}
Chiyuan Zhang, Samy Bengio, Moritz Hardt, Benjamin Recht, and Oriol Vinyals.
\newblock Understanding deep learning requires rethinking generalization.
\newblock In \emph{5th International Conference on Learning Representations,
  {ICLR} 2017, Toulon, France, April 24-26, 2017, Conference Track
  Proceedings}, 2017.

\bibitem[Zhang et~al.(2018)Zhang, Ciss{\'{e}}, Dauphin, and
  Lopez{-}Paz]{zhang2018mixup}
Hongyi Zhang, Moustapha Ciss{\'{e}}, Yann~N. Dauphin, and David Lopez{-}Paz.
\newblock mixup: Beyond empirical risk minimization.
\newblock In \emph{6th International Conference on Learning Representations,
  {ICLR} 2018, Vancouver, BC, Canada, April 30 - May 3, 2018, Conference Track
  Proceedings}, 2018.

\bibitem[Zhang et~al.(2019)Zhang, Dauphin, and Ma]{zhang2019fixup}
Hongyi Zhang, Yann~N. Dauphin, and Tengyu Ma.
\newblock Residual learning without normalization via better initialization.
\newblock In \emph{International Conference on Learning Representations}, 2019.

\end{thebibliography}
\bibliographystyle{iclr2021_conference}

\newpage
\setcounter{page}{1}
\appendix
\onecolumn
\section*{Apppendix}

\section{Additional results}
\label{sm_additional_results}

\subsection{Early phase \TrF correlates with final generalization}
\label{sec_early_final_appendix}

\begin{figure}[H]
    \centering
        \begin{subfigure}[t]{0.32\textwidth}
       \includegraphics[width=0.9\columnwidth ,trim=0.1in 0.in 0.in 0.in,clip]{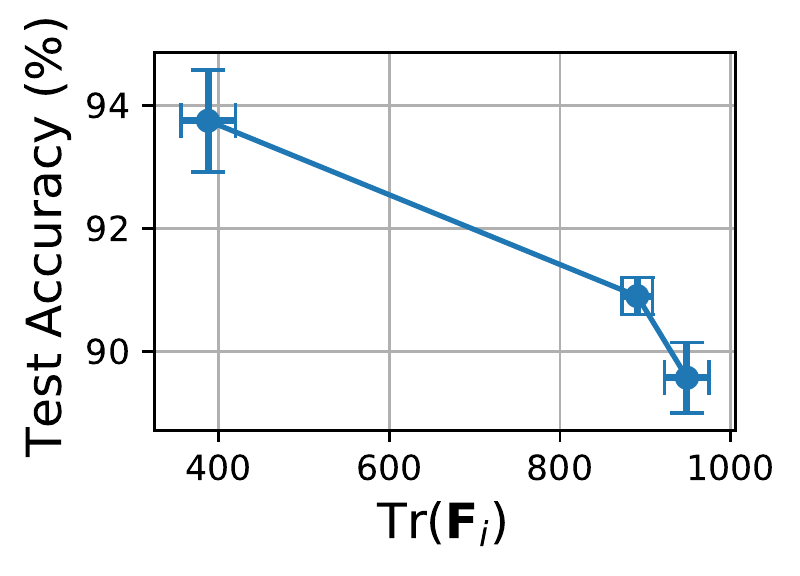}
    \caption{CIFAR-10 (with aug.)}
    \end{subfigure}
    \begin{subfigure}[t]{0.32\textwidth}
      \includegraphics[width=0.9\columnwidth ,trim=0.1in 0.in 0.in 0.in,clip]{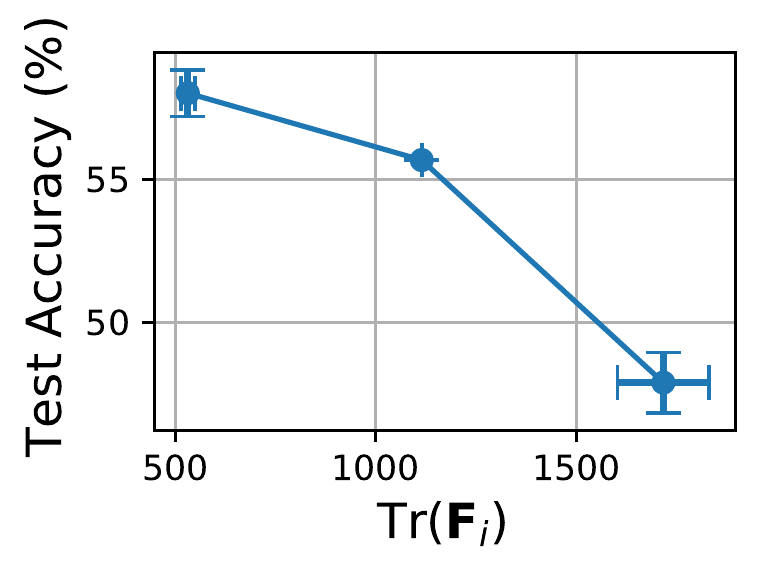}
    \caption{CIFAR-100 (w/o aug.)}
    \end{subfigure}
    \caption{Association between early phase values of \TrF and generalization holds on the CIFAR-10 and CIFAR-100 datasets. Each point corresponds to multiple runs with randomly chosen seeds and a specific value of batch size. $\text{Tr}\mathbf{F}_i$ is recorded during the early phase (2-7 epochs, see main text for details), while the test accuracy is the maximum value along the entire optimization path (averaged across runs with the same batch size). The horizontal and vertical error bars show the standard deviation of values across runs. The plots show that early phase \TrF is predictive of final generalization.}
    \label{fig:IF_LR_trF_bs}
\end{figure}

\begin{figure}[H]
    \centering
        \begin{subfigure}[t]{0.32\textwidth}
       \includegraphics[width=0.9\columnwidth ,trim=0.1in 0.in 0.in 0.in,clip]{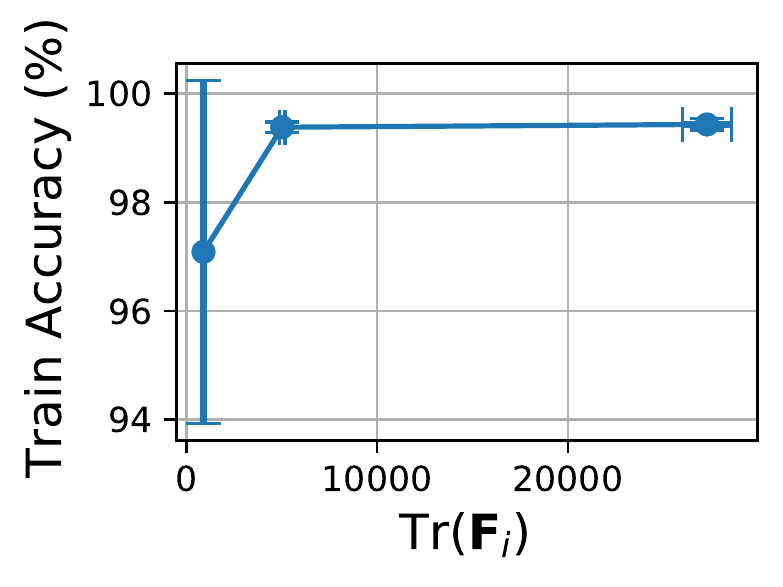}
\caption{ImageNet (w/o augmentation)}
    \end{subfigure}
    \hfill
        \begin{subfigure}[t]{0.32\textwidth}
       \includegraphics[width=0.9\columnwidth ,trim=0.1in 0.in 0.in 0.in,clip]{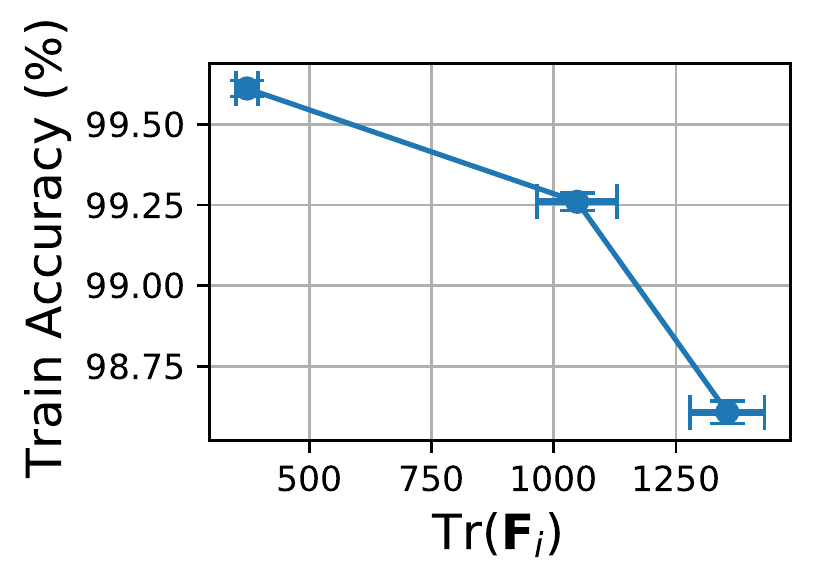}
    \caption{CIFAR-10 (with augmentation)}
    \end{subfigure}
    \hfill
    \begin{subfigure}[t]{0.32\textwidth}
      \includegraphics[width=0.97\columnwidth ,trim=0.1in 0.in 0.in 0.in,clip]{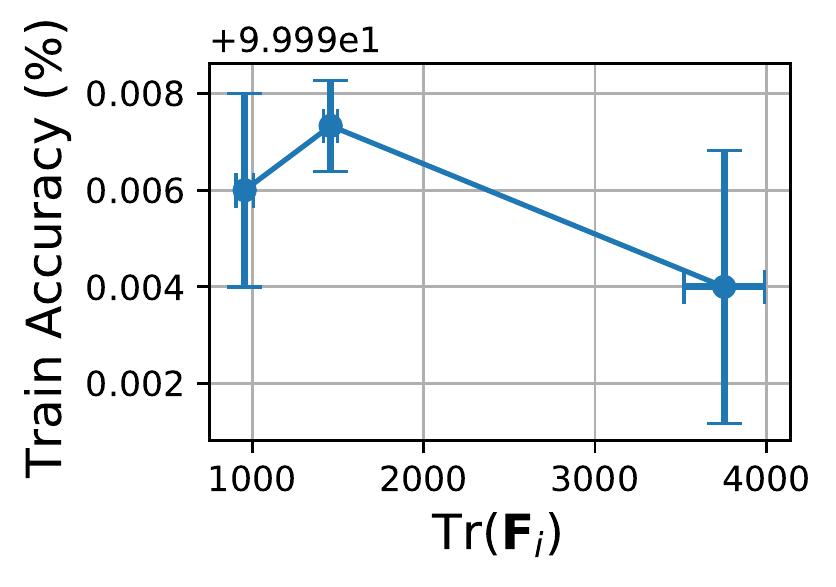}
    \caption{CIFAR-100 (w/o augmentation)}
    \end{subfigure}
    \caption{Training accuracy for the runs corresponding to Figure \ref{fig:IF_LR_trF}. Each point corresponds to multiple seeds and a specific value of learning rate. \TrFi is recorded during the early phase of training (2-7 epochs, see the main text for details), while the training accuracy is the maximum value along the entire optimization path (averaged across runs with the same learning rate).}
    \label{fig:IF_LR_trF_train}
\end{figure}

\begin{figure}[H]
    \centering
        \begin{subfigure}[t]{0.32\textwidth}
       \includegraphics[width=0.9\columnwidth ,trim=0.1in 0.in 0.in 0.in,clip]{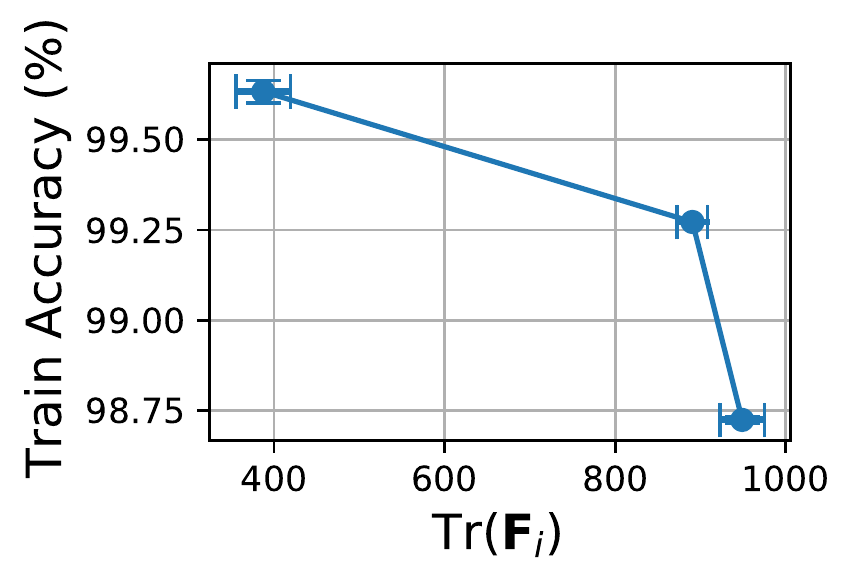}
    \caption{CIFAR-10 (with aug.)}
    \end{subfigure}
    \begin{subfigure}[t]{0.32\textwidth}
      \includegraphics[width=0.9\columnwidth ,trim=0.1in 0.in 0.in 0.in,clip]{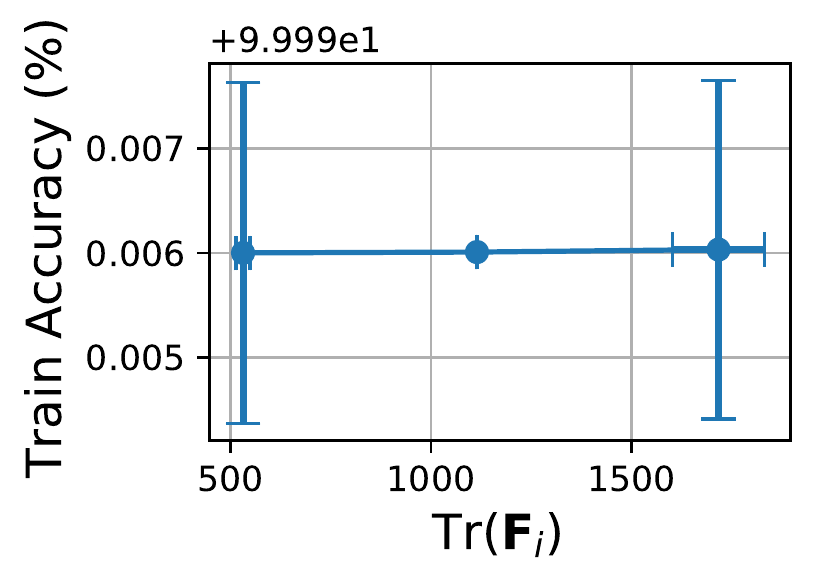}
    \caption{CIFAR-100 (w/o aug.)}
    \end{subfigure}
    \caption{Training accuracy for the runs corresponding to Figure \ref{fig:IF_LR_trF_bs}. Each point corresponds to multiple runs with randomly chosen seeds and a specific value of batch size. $\text{Tr}\mathbf{F}_i$ is recorded during the early phase (2-7 epochs, see main text for details), while the training accuracy is the maximum value along the entire optimization path (averaged across runs with the same batch size).}
    \label{fig:IF_LR_trF_bs_train}
\end{figure}

In this section, we present the additional experimental results for Section~\ref{sec_trf_generalization}. Figure~\ref{fig:IF_LR_trF_bs} shows the experiments with varying batch size for CIFAR-100 and CIFAR-10. The conclusions are the same as discussed in the main text in Section~\ref{sec_trf_generalization}. We also show the training accuracy for all the experiments performed in Figure \ref{fig:IF_LR_trF} and Figure~\ref{fig:IF_LR_trF_bs}. They are shown in Figure \ref{fig:IF_LR_trF_train} and Figure~\ref{fig:IF_LR_trF_bs_train} respectively. Most runs in all these experiments reach training accuracy  $\sim$99\% and above.

\subsection{Fisher Penalty}
\label{app:fisher_penalty_additional_results}

We first show additional metrics for experiments summarized in Table~\ref{tab:fisher_penalty_setting1} in the main text. Table~\ref{app:tab:fisher_penalty_setting1_acc} summarizes the final training accuracy, showing that the baseline experiments were trained until approximately 100\% training accuracy was reached. Table~\ref{app:tab:fisher_penalty_setting1_TrF} supports the claim that all gradient norm regularizers reduce the maximum value of \TrF (we measure \TrF starting from after one epoch of training because \TrF explodes in networks with batch normalization layers at initialization). Finally, Table~\ref{app:tab:fisher_penalty_setting1_time} confirms that the regularizers incurred a relatively small additional computational cost.

\begin{figure}[H]
    \centering
    \begin{subfigure}[t]{0.48\textwidth}
    \centering
       \includegraphics[width=1.\columnwidth]{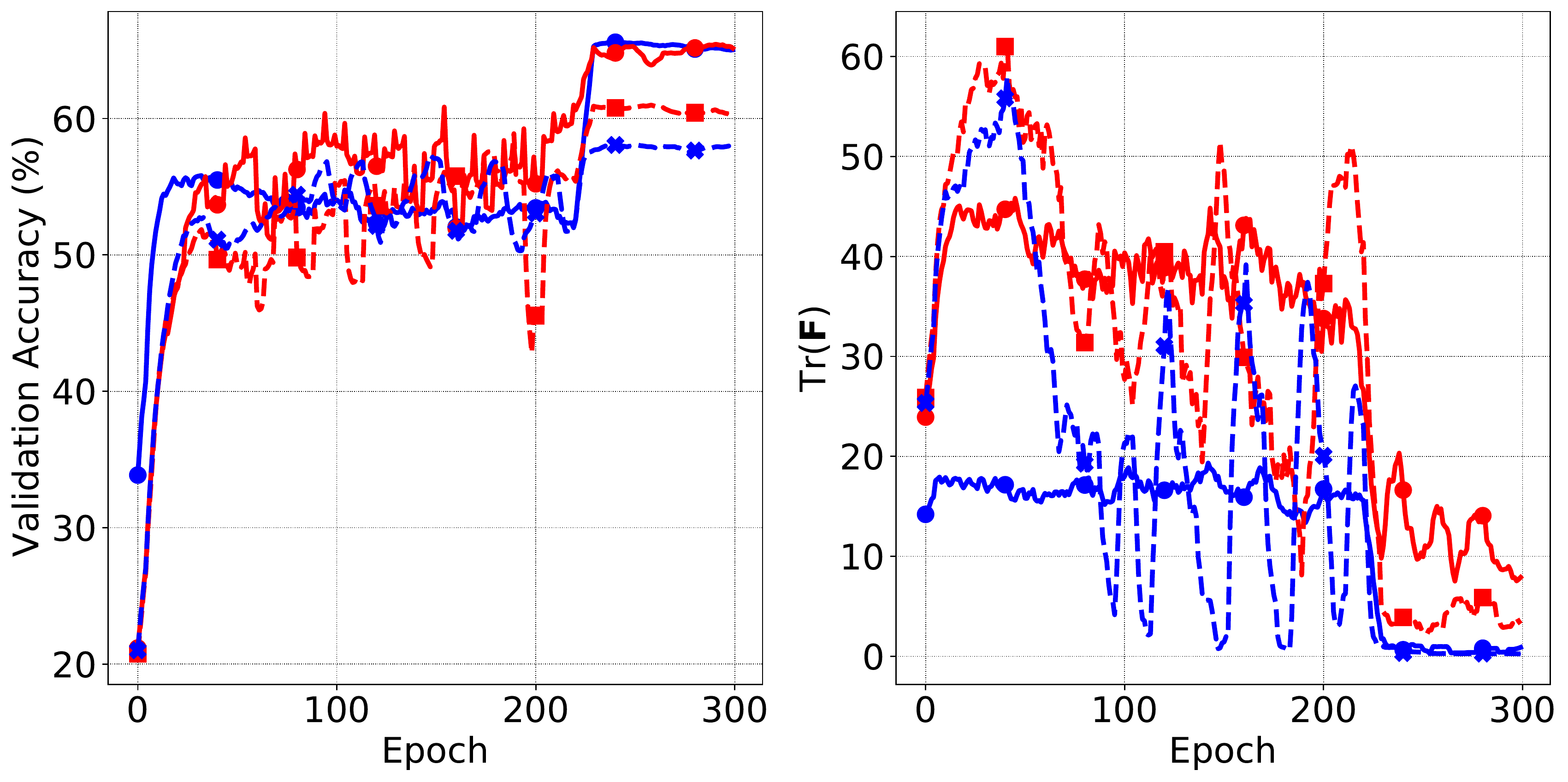}
        \vspace{-0.3cm}
       \includegraphics[width=0.6\columnwidth]{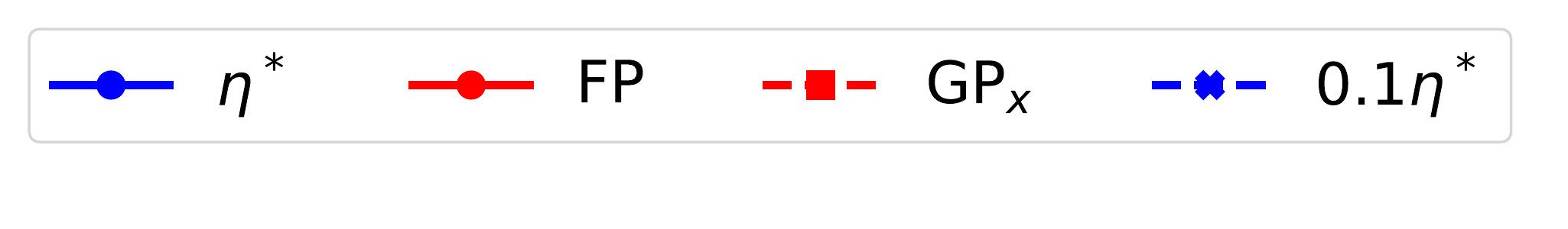}
\caption{DenseNet on CIFAR-100 (w/o aug.)}
    \end{subfigure}
    \begin{subfigure}[t]{0.48\textwidth}
    \centering
       \includegraphics[width=1.\columnwidth]{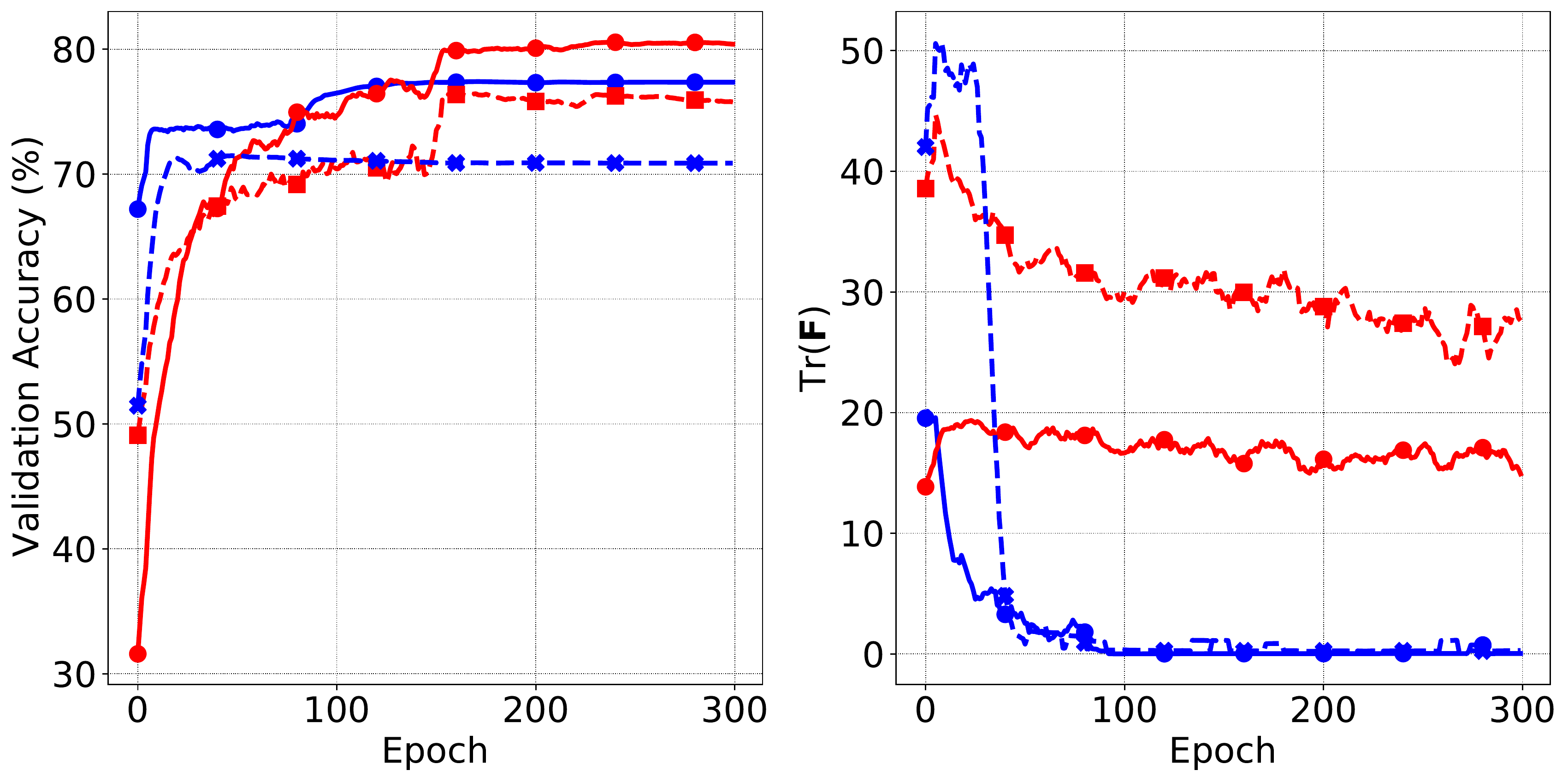}
        \vspace{-0.3cm}
       \includegraphics[width=0.6\columnwidth]{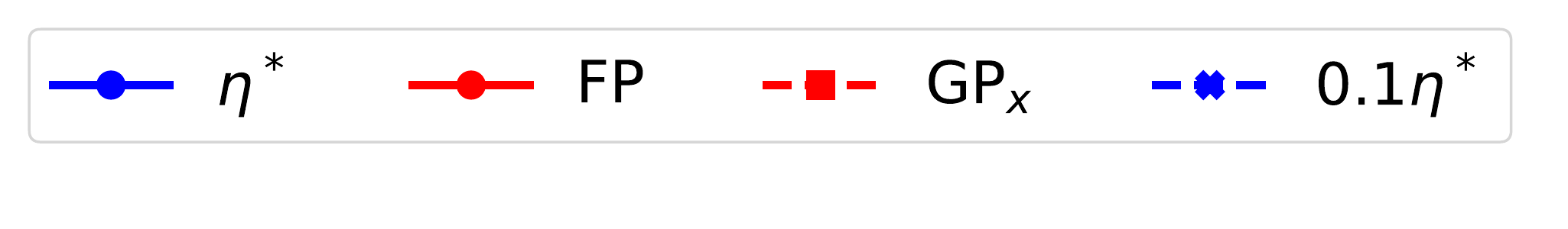}
\caption{SimpleCNN on CIFAR-10 (w/o aug.)}
    \end{subfigure}
        \caption{Same as Figure~\ref{fig:fisher_penalty_setting1_visualization}, but for DenseNet on CIFAR-100, and SimpleCNN on CIFAR-10. Curves were smoothed for visual clarity.}
    \label{app:fig:fisher_penalty_setting1_visualization}
\end{figure}

Figure~\ref{app:fig:fisher_penalty_setting1_visualization} complements Figure~\ref{fig:fisher_penalty_setting1_visualization} for the other two models on the CIFAR-10 and CIFAR-100 datasets. The figures are in line with the results of the main text.

Lastly, in Table~\ref{app:tab:fisher_penalty_setting2_acc} we report the final training accuracy reached by runs reported in Table~\ref{tab:fisher_penalty_setting2} in the main text.

\begin{table}[H]
\centering
\small
\caption{The maximum value of \TrF along the optimization trajectory for experiments on CIFAR-10 or CIFAR-100 included in Table~\ref{tab:fisher_penalty_setting1}.}
\label{app:tab:fisher_penalty_setting1_TrF}
\begin{tabular}{cll|ll|ll}
\toprule
              Setting & $\eta^*$ & Baseline & \GPx & \GP &  \FP & \GPr \\
              \midrule
        DenseNet/C100 (w/o aug.) &    24.68 &       98.17 &                83.64 &              64.33 &                     66.24 &                73.66 \\
           VGG11/C100 (w/o aug.) &    50.88 &      148.19 &               102.95 &              58.53 &                     64.93 &                62.96 \\
         WResNet/C100 (w/o aug.) &    26.21 &       91.39 &                41.43 &              40.94 &                     56.53 &                39.31 \\
          \midrule
             SCNN/C10 (w/o aug.) &    24.21 &       52.05 &                47.96 &              25.03 &                     19.63 &                25.35 \\
\bottomrule
\end{tabular}
\end{table}

\begin{table}[H]
\centering
\small
\caption{Time per epoch (in seconds) for experiments in Table~\ref{tab:fisher_penalty_setting1}. }
\label{app:tab:fisher_penalty_setting1_time}
\begin{tabular}{cll|ll|ll}
\toprule
             Setting & $\eta^*$ & Baseline & \GPx & \GP &  \FP & \GPr \\
                       \toprule
 WResNet/TinyImageNet (aug.) &   214.45 &      142.69 &               233.14 &             143.78 &                    208.62 &               371.74 \\
 \midrule
        DenseNet/C100 (w/o aug.) &    78.88 &       57.40 &                77.89 &              78.66 &                     97.25 &                75.96 \\
           VGG11/C100 (w/o aug.) &    30.50 &       35.27 &                31.54 &              32.52 &                     43.41 &                42.40 \\
         WResNet/C100 (w/o aug.) &    49.64 &       47.99 &                71.33 &              61.36 &                     76.93 &                53.25 \\
         
          \midrule
             SCNN/C10 (w/o aug.) &    18.64 &       19.51 &                26.09 &              19.91 &                     21.21 &                20.55 \\
\bottomrule
\end{tabular}
\end{table}

\begin{table}[H]
\centering
\small
\caption{The final epoch training accuracy for experiments shown in Table~\ref{tab:fisher_penalty_setting1}. Experiments with small learning rate reach no lower accuracy than experiments corresponding to a large learning rate $\eta^*$.}
\label{app:tab:fisher_penalty_setting1_acc}
\begin{tabular}{cll|ll|ll}
\toprule
              Setting & $\eta^*$ & Baseline & \GPx & \GP &  \FP & \GPr \\
                       \midrule
 WResNet/TinyImageNet (aug.) &            99.84\% &            99.96\% &     {99.97\%} &            93.84\% &                   81.05\% &              86.46\% \\
 \midrule
        DenseNet/C100 (w/o aug.) &   {99.98\%} &            99.97\% &              99.96\% &            99.91\% &                   99.91\% &              99.39\% \\
           VGG11/C100 (w/o aug.) &   {99.98\%} &   {99.98\%} &              99.85\% &            99.62\% &                   97.73\% &              86.32\% \\
         WResNet/C100 (w/o aug.) &            99.98\% &            99.98\% &              99.97\% &            99.96\% &          {99.99\%} &              99.94\% \\
          \midrule
             SCNN/C10 (w/o aug.) &  {100.00\%} &  {100.00\%} &              97.79\% &  {100.00\%} &                   93.80\% &              94.64\% \\
\bottomrule
\end{tabular}
\end{table}

\begin{table}[H]
\centering
\small
\caption{The final epoch training accuracy for experiments shown in Table~\ref{tab:fisher_penalty_setting2}. }
\label{app:tab:fisher_penalty_setting2_acc}
\begin{tabular}{cll}
\toprule
                     Setting &                   $\eta^*$ &              $\mathrm{FP}$ \\
\midrule
         DenseNet/C100 (aug) &  {98.75$\pm$0.27\%} &           97.53$\pm$1.75\% \\
              SCNN/C10 (aug) &           97.52$\pm$1.98\% &  {98.94$\pm$0.08\%} \\
            VGG11/C100 (aug) &  {98.64$\pm$0.06\%} &           93.06$\pm$0.01\% \\
          WResNet/C100 (aug) &  {99.97$\pm$0.01\%} &           99.97$\pm$0.00\% \\
 WResNet/Tiny ImageNet (aug) &  {99.86$\pm$0.02\%} &           93.65$\pm$5.85\% \\
\bottomrule
\end{tabular}
\end{table}

\subsection{Fisher Penalty Reduces Memorization}
\label{app:sec:fisher_penalty_memorization}

\begin{figure*}[t]
\centering
  \begin{subfigure}[t]{0.9\textwidth}
  \centering
    \includegraphics[width=1.0\columnwidth]{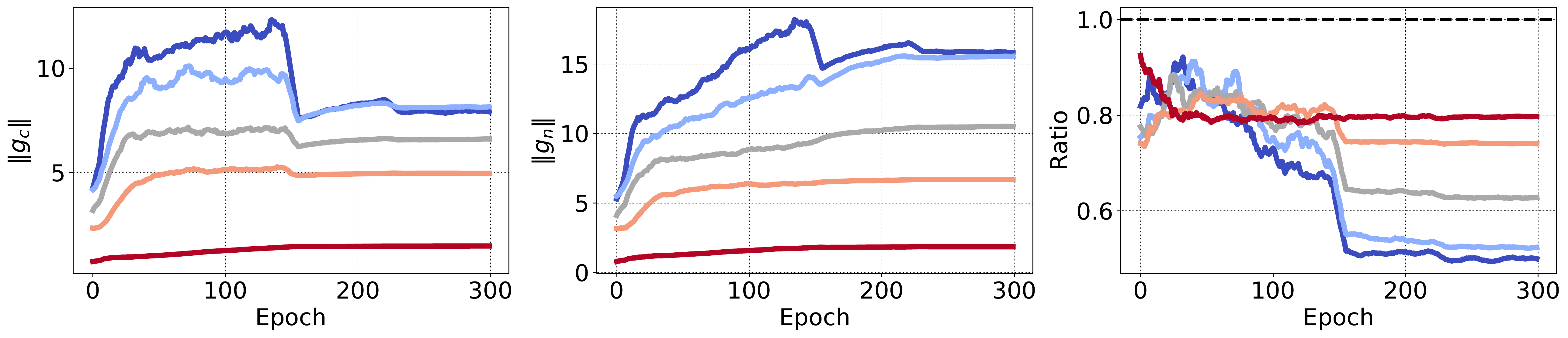}
\caption{Gradient norm on clean examples (left, denoted as $g_c$), noisy examples (middle, denoted as $g_n$), and their ratio (right); evaluated on the training set.}
     \end{subfigure}
         \begin{subfigure}[t]{0.9\textwidth}
    \centering
 \includegraphics[width=1.0\columnwidth]{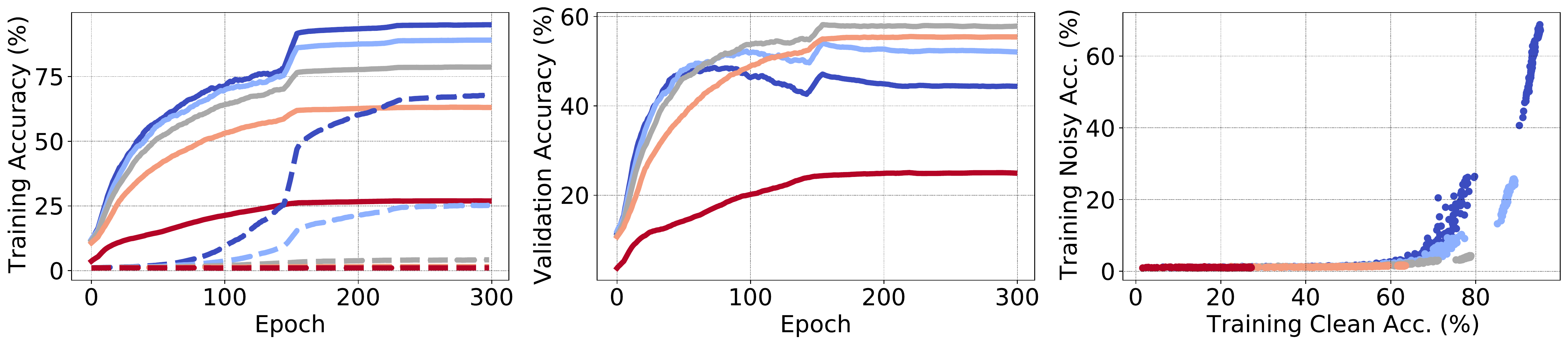}
 \caption{Training accuracy on clean and noisy examples (solid/dashed lined, left), validation accuracy (middle), and a scatter plot of training accuracy on clean vs noisy examples (right). }
 \end{subfigure}
 \caption{Same as Figure~\ref{fig:fisher_penalty_noisy_data_visualization}, but for ResNet-50 trained on the CIFAR-100 dataset.}
 \label{app:fig:fisher_penalty_noisy_data_visualization}
\end{figure*}

\begin{figure}[H]
\centering
     \begin{subfigure}[t]{0.3\textwidth}
    \centering
 \includegraphics[width=1.0\columnwidth]{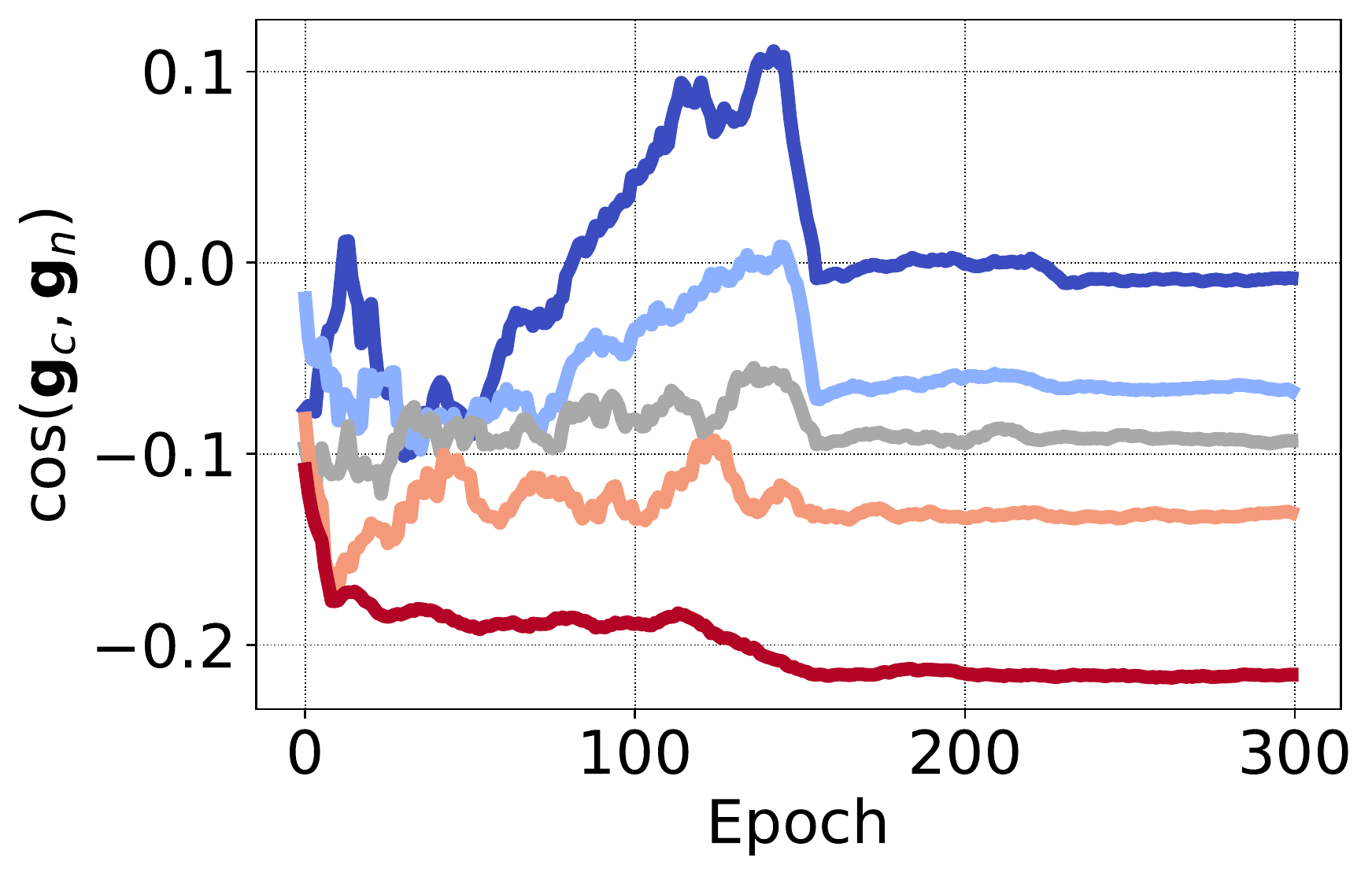}
 \caption{\GPr}
 \end{subfigure}
      \begin{subfigure}[t]{0.3\textwidth}
    \centering
 \includegraphics[width=1.0\columnwidth]{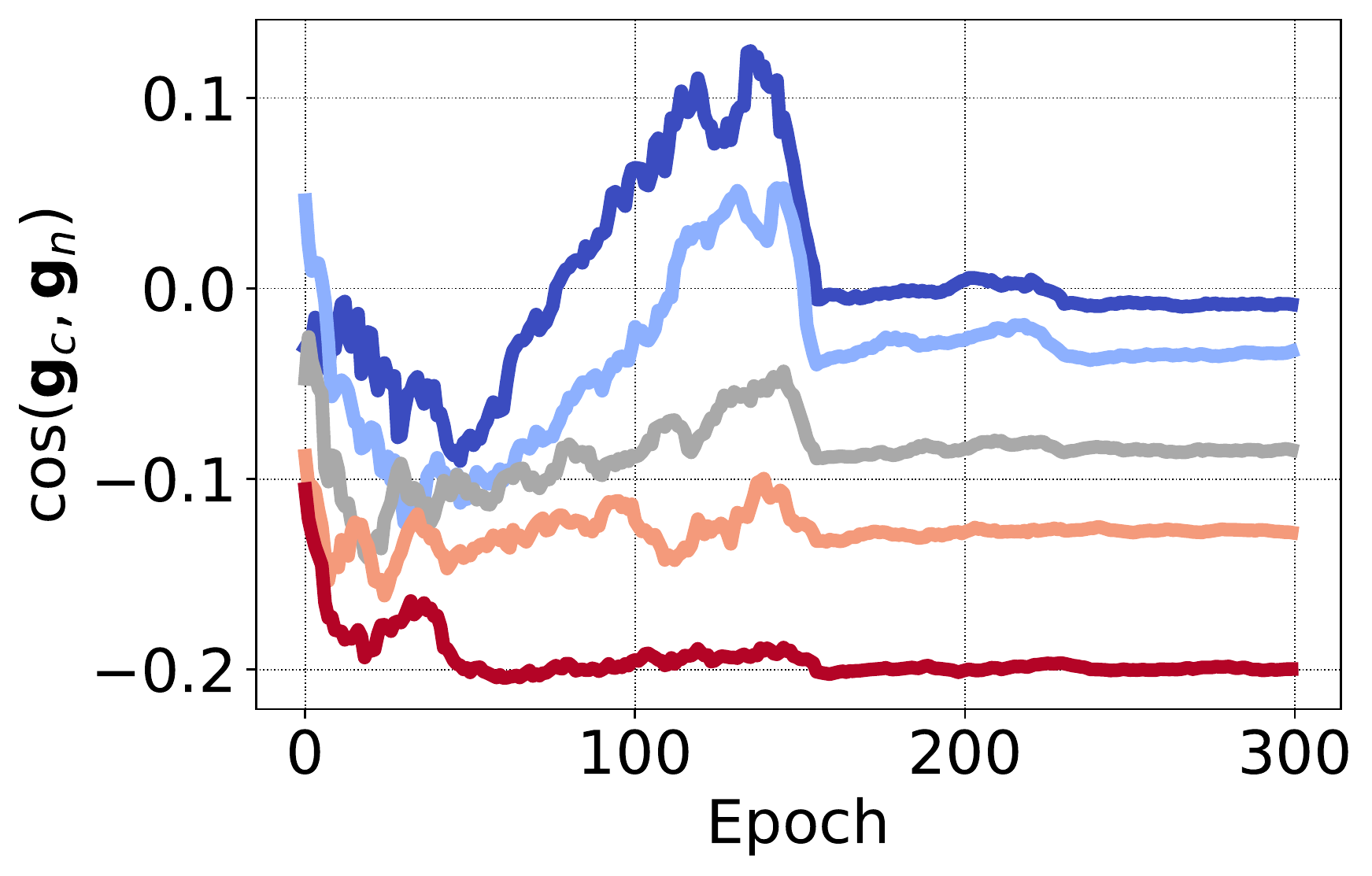}
 \caption{\FP}
 \end{subfigure}
      \begin{subfigure}[t]{0.3\textwidth}
    \centering
 \includegraphics[width=1.0\columnwidth]{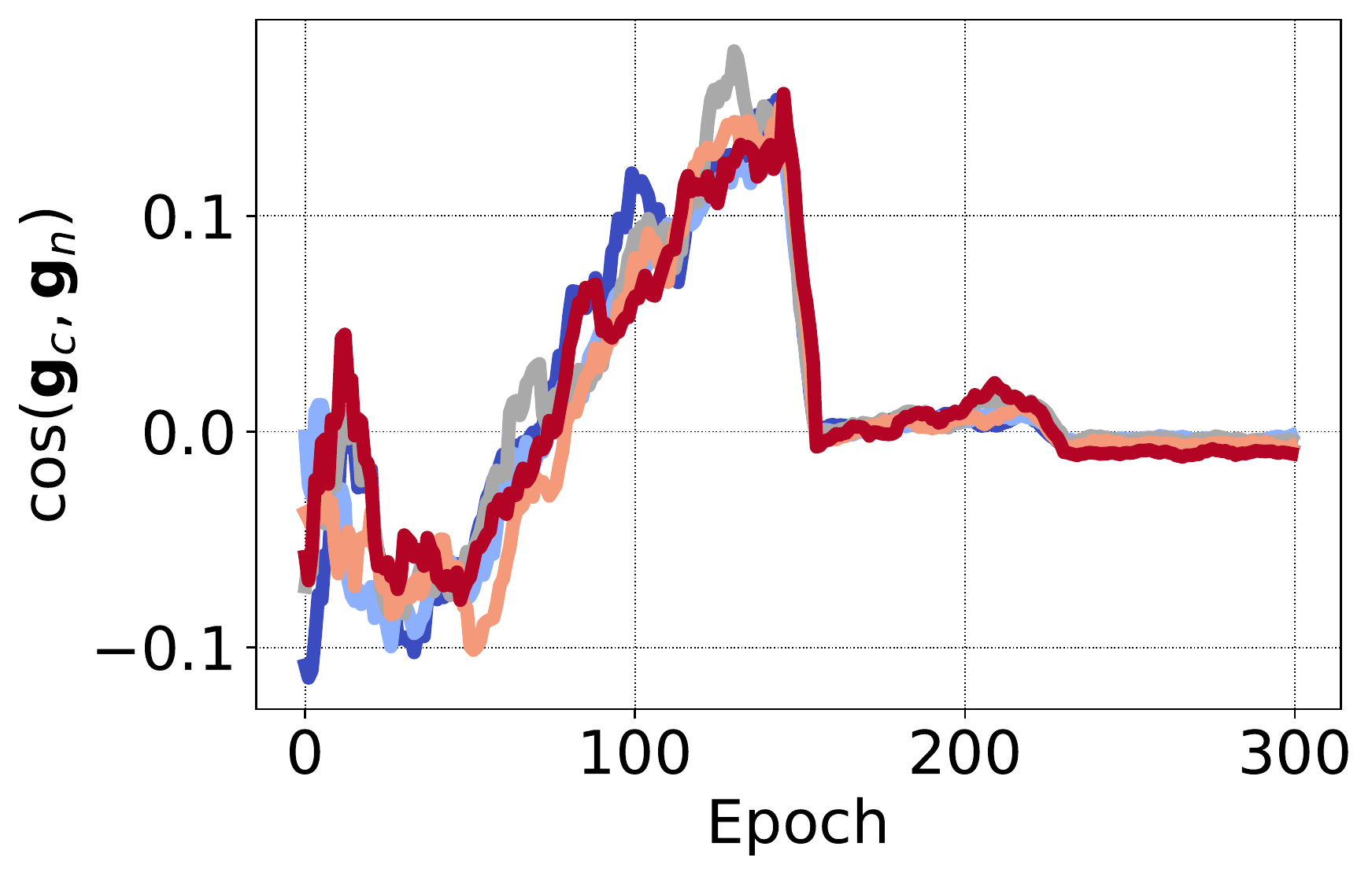}
 \caption{\GPx}
 \end{subfigure}
 \caption{The cosine between the mini-batch gradients computed on the noisy ($\mathbf{g}_n$) and clean ($\mathbf{g}_c$) data, both measured on the training set. We observe that in the early phase of training the angle is negative. Furthermore, stronger regularization with \GPr or \FP correlates with a more negative angle.}
 \label{app:fig:fisher_penalty_angle}
\end{figure}

In this section, we include additional experimental results for Section~\ref{sec:fisher_penalty_prevents_memorization}.  Figure~\ref{app:fig:fisher_penalty_noisy_data_visualization} is the same as Figure~\ref{fig:fisher_penalty_noisy_data_visualization}, but for ResNet-50. Finally, we show additional metrics for the experiments involving 25\% noisy examples. Figure~\ref{app:fig:fisher_penalty_angle} shows the cosine between the mini-batch gradients computed on the noisy and clean data. In Table~\ref{app:tab:fisher_penalty_noisy_data_acc_noisy} and Table~\ref{app:tab:fisher_penalty_noisy_data_acc_clean} we show training accuracy on the noisy and clean examples in the final epoch of training.

\begin{table*}
\centering
\caption{Training accuracy on the clean examples int the final epoch, for experiments reported in Table~\ref{tab:fisher_penalty_noisy_data} (with 25\% examples with noisy labels).}
\label{app:tab:fisher_penalty_noisy_data_acc_clean}
\begin{tabular}{lllll|ll}
\toprule
Label Noise &      Setting & Baseline &    Mixup & \GPx & \FP & \GPr \\
\midrule
   25\% &  VGG-11/C100 &  99.79\% &  73.14\% &                        97.46\% &       79.52\% &                        81.75\%   \\
     &  ResNet-52/C100 &  95.87\% &  77.71\% &                        95.88\% &       78.72\% &                        74.21\% \\
\bottomrule
\end{tabular}
\end{table*}

\begin{table*}
\centering
\caption{Training accuracy on the noisy examples in the final epoch, for experiments reported in Table~\ref{tab:fisher_penalty_noisy_data} (with 25\% examples with noisy labels).}
\label{app:tab:fisher_penalty_noisy_data_acc_noisy}
\begin{tabular}{lllll|ll}
\toprule
Label Noise &      Setting & Baseline &    Mixup & \GPx & \FP & \GPr \\
\midrule
   25\% &  VGG-11/C100 &  99.56\% &  8.29\% &                        89.23\% &        4.10\% &                         4.96\%   \\
     & ResNet-52/C100 &  73.67\% &  4.22\% &                        72.67\% &        4.14\% &                         2.86\% \\
\bottomrule
\end{tabular}
\end{table*}

\subsection{Early \TrF influences final curvature}
\label{sec_histogram_appendix}

\begin{figure}
\vspace{-1pt}
  \centering
        \begin{subfigure}[t]{1\textwidth}
    \includegraphics[width=1.\columnwidth ,trim=0.1in 0.in 0.in 0.in,clip]{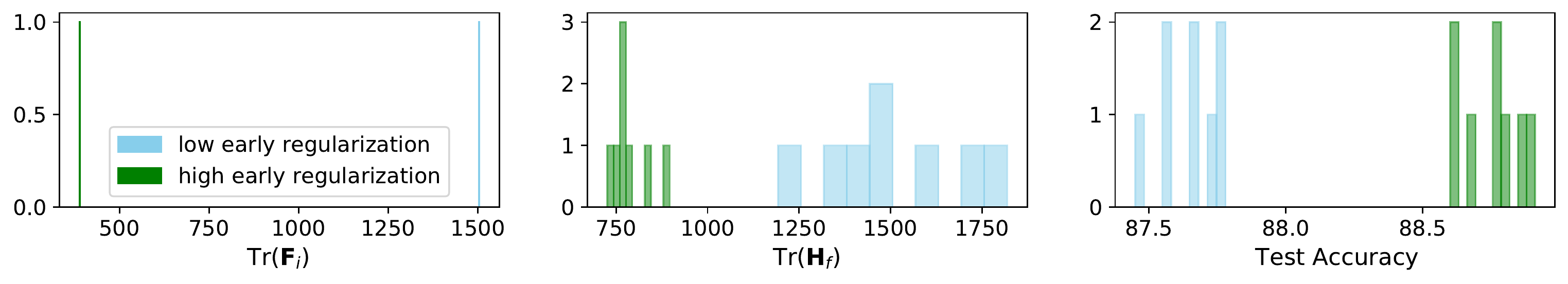}
\end{subfigure}
    \caption{Small \TrF during the early phase of training is more likely to reach wider minima as measured by \TrH. Left: 2 models are trained with different levels of regularization for 20 epochs on CIFAR-10. \TrF at the end of 20 epochs (denoted as \TrFi) is shown. Middle: Each model is then used as initialization and trained until convergence using the low regularization configuration with different random seeds. A histogram of \TrH at the point corresponding to the best test accuracy along the trajectory (denoted by \TrHf) is shown. Right: a histogram of the best test accuracy corresponding to middle figure is shown.}
    \label{fig:histogram_c10}
\end{figure}

In this section, we present additional experimental results for Section~\ref{sec_trf_final_minima}.  The experiment on CIFAR-10 is shown in Figure~\ref{fig:histogram_c10}. The conclusions are the same as discussed in the main text in Section~\ref{sec_trf_final_minima}.

\begin{figure}[H]
   \centering
    \begin{subfigure}[t]{0.47\textwidth}
    \includegraphics[width=1.\columnwidth ,trim=0.1in 0.in 0.in 0.in,clip]{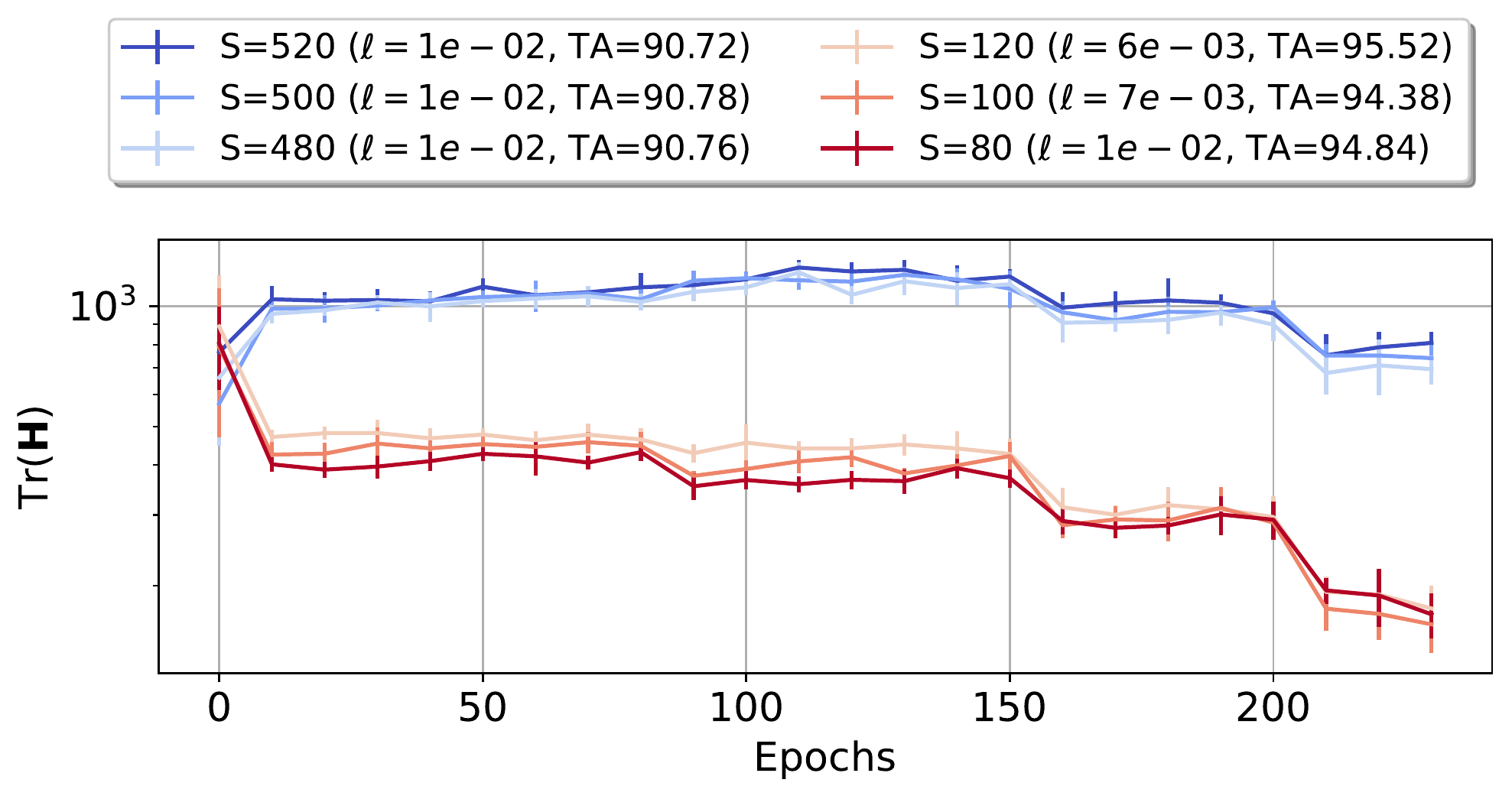}
\caption{CIFAR-10 (w/ augmentation)}
    \end{subfigure}
    \hfill
    \begin{subfigure}[t]{0.47\textwidth}
      \includegraphics[width=1.05\columnwidth ,trim=0.1in 0.in 0.in 0.in,clip]{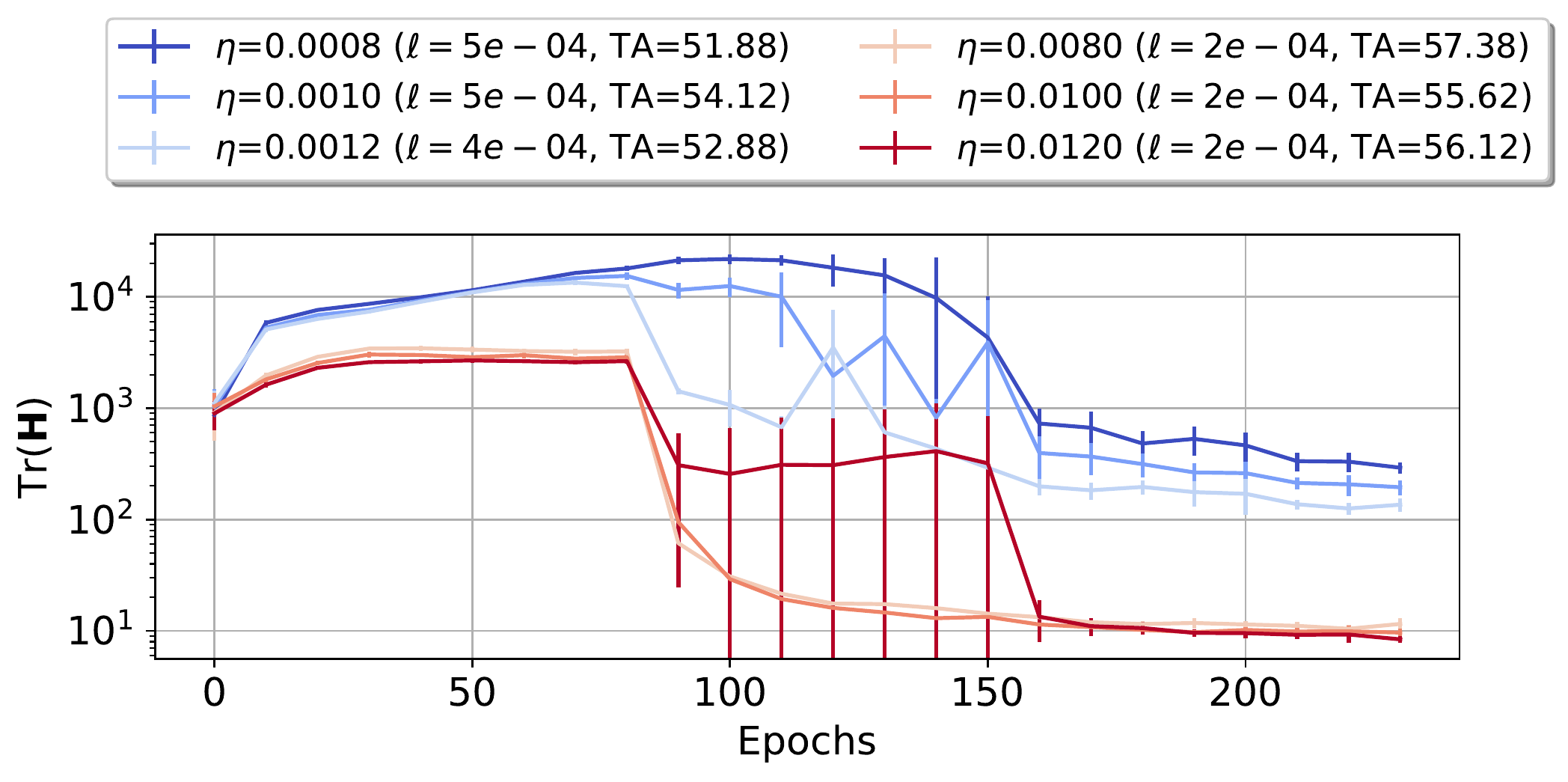}
    \caption{CIFAR-100 (w/o augmentation)}
    \end{subfigure}
    \caption{The value of \TrH over the course of training. Each point corresponds to runs with different seeds and a specific value of learning rate $\eta$ and batch size $S$. $\ell$ and TA respectively denote the minimum training loss and the maximum test accuracy along the entire trajectory for the corresponding runs (averaged across seeds). The plots show that flatter optimization trajectories become biased towards flatter minima early during training, at a coarse scale of hyper-parameter values (red vs blue).}
    \label{fig:IF_GT_appendix}
\vspace{-5pt}
\end{figure}

Next, to understand why smaller \TrF during the early phase is more likely to end up in a wider final minimum, we track \TrH during the entire course of training and show that it stabilizes early on. In this experiment, we create two sets of hyper-parameters: coarse-grained and fine-grained. For CIFAR-10, we use batch size $S \in A\cup B$, where $A=\{480,500,520\}$ and $B = \{80,100,120\}$. For all batch size configurations, a learning rate of 0.02 is used. Overloading the symbols $A$ and $B$ for CIFAR-100, we use learning rate $\eta \in A\cup B$, where $A = \{0.0008,0.001, 0.0012\}$ and $B=\{0.008, 0.01, 0.012\}$. For all learning rate configurations, a batch size of 100 is used. In both cases, the elements within each set ($A$ and $B$) vary on a fine-grained scale, while the elements across the two sets vary on a coarse-grained scale. The remaining details and additional experiments can be found in Appendix \ref{sec_trf_final_minima_details}. The experiments are shown in Figure \ref{fig:IF_GT_appendix}. Notice that after initialization (index 0 on the horizontal axis), the first value is computed at epoch 10 (at which point the experiments show that entanglement has started to hold in alignment with the late phase). 

We make three observations in this experiment. First, the relative ordering of \TrH values for runs between sets $A$ vs $B$ stays the same after the first 10 epochs. Second, the degree of entanglement is higher between any two epochs when looking at runs across sets $A$ and $B$, while it is weaker when looking at runs within any one of the sets. Finally, test accuracies for set $B$ runs are always higher than those of set $A$ runs, but this trend is not strong for runs within any one set. Note that the minimum loss values are roughly at a similar scale for each dataset and they are all at or below $10^{-2}$.

\section{Computation of \TrH }
\label{app:computation_of_TrH}
We computed \TrH in our experiments using the Hutchinson's estimator \cite{hutchinson1990stochastic},
\begin{align*}
    Tr(\mathbf{H}) &= Tr(\mathbf{H} \cdot \mathbf{I})\\
    &= Tr(\mathbf{H} \cdot \mathbb{E}[\mathbf{z}\mathbf{z}^T])\\
    &= \mathbb{E}[Tr(\mathbf{H} \cdot \mathbf{z}\mathbf{z}^T)]\\
    &= \mathbb{E}[\mathbf{z}^T\mathbf{H} \cdot \mathbf{z}]\\
    &\approx \frac{1}{M} \sum_{i=1}^{M} \mathbf{z}_i^T\mathbf{H} \cdot \mathbf{z}_i\\
    &= \frac{1}{M} \sum_{i=1}^{M} \mathbf{z}_i^T \frac{\partial }{\partial \theta } \left( \frac{\partial \ell}{ \partial \theta^T} \right) \cdot \mathbf{z}_i\\
    &= \frac{1}{M} \sum_{i=1}^{M} \mathbf{z}_i^T \frac{\partial }{\partial \theta } \left( \frac{\partial \ell}{ \partial \theta}^T \mathbf{z}_i \right),
\end{align*}
where $\mathbf{I}$ is the identity matrix, $\mathbf{z}$ is a multi-variate standard Gaussian random variable, and $\mathbf{z}_i$'s are i.i.d. instances of $\mathbf{z}$. The larger the value of $M$, the more accurate the approximation is. We used $M=30$. To make the above computation efficient, note that the gradient $\frac{\partial \ell}{ \partial \theta}$ only needs to be computed once and it can be re-used in the summation over the $M$ samples.

\section{Approximations in Fisher Penalty}
\label{app:approx_in_FP}

In this section, we describe in detail the approximations made when computing the Fisher Penalty. Recall that \TrF can be expressed as

\begin{equation}
\mathrm{Tr}(\mathbf{F}) = \mathbb{E}_{x \sim \mathcal{X},\hat y \sim p_{\theta}(y|\bm{x})} \left[ \Vert \frac{\partial}{\partial \mathbf{\theta}}\ell(\bm{x},\hat y) \Vert_2^2 \right].
\end{equation}

In the preliminary experiments, we found that we can use the norm of the expected gradient rather than the expected norm of the gradient, which is a more direct expression of \TrF: 
\begin{align*}
    \nabla 
    \mathbb{E}_{x \sim \mathcal{X},\hat y \sim p_{\theta}(y|\bm{x})} \left[ \left\| \frac{\partial}{\partial \mathbf{\theta}}\ell(\bm{x},\hat y) \right\|_2^2 \right]
    &\approx
    \frac{1}{N}
    \sum_{n=1}^N
    \frac{1}{M}
    \sum_{m=1}^M
    \nabla 
    \left\|
    \frac{\partial}{\partial \mathbf{\theta}}\ell(\bm{x}_n,\hat y_{nm})
    \right\|^2_2
    \\
    &\geq
    \nabla 
    \left\|
    \frac{1}{NM}
    \sum_{n=1}^N
    \sum_{m=1}^M
    \frac{\partial}{\partial \mathbf{\theta}}\ell(\bm{x}_n,\hat y_{nm})
    \right\|^2_2,
    \end{align*}
    
where $N$ and $M$ are the minibatch size and the number of samples from $p_{\theta}(y|\bm{x}_n)$, respectively. This greatly improves the computational efficiency. With $N=B$ and $M=1$, we end up with the following learning objective function:

\begin{equation}
\label{app:eq_fisher_objective}
\ell'(\bm{x}_{1:B}, y_{1:B}; \bm{\theta}) = \frac{1}{B} \sum_{i=1}^B \ell(\bm{x}_i,y_i; \bm{\theta}) + \alpha \left\| \frac{1}{B} \sum_{i=1}^B g(\bm{x}_i, \hat y_i) \right\|^2.
\end{equation}

\begin{figure}[H]
\centering
      \includegraphics[width=0.5\columnwidth]{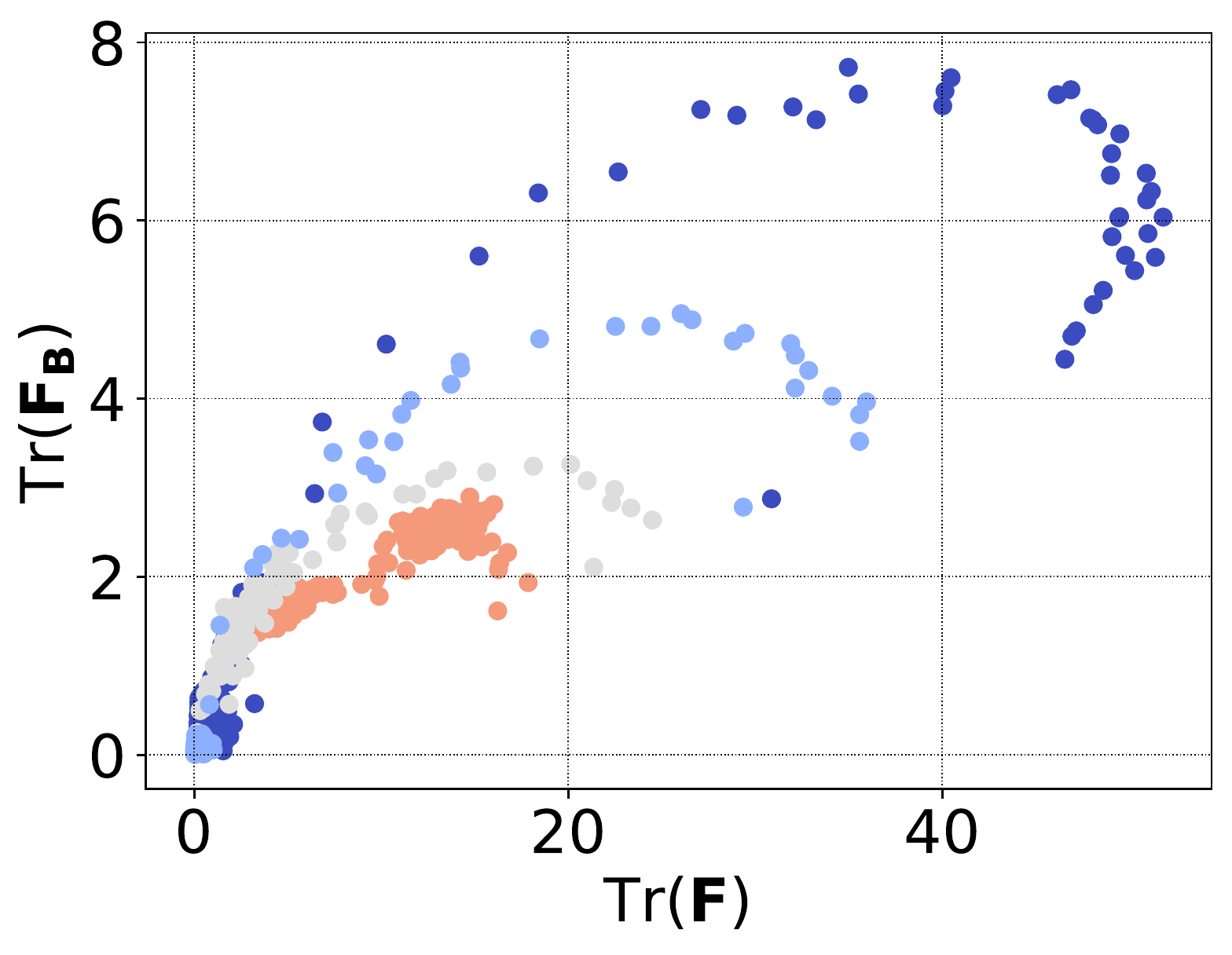}
  \caption{Correlation between \TrF and $\mathrm{Tr}(\mathbf{F}^B)$ for SimpleCNN trained on the CIFAR-10 dataset. Blue to red color denotes learning rates from $10^{-3}$ to $10^{-1}$. The value of \TrF and $\mathrm{Tr}(\mathbf{F}^B)$  correlate strongly for the most of the training trajectory. Using large learning rate reduces both \TrF and $\mathrm{Tr}(\mathbf{F}^B)$. }
\label{app:fig:correlationTrBFvTrF}
\end{figure}

We found empirically that $\left\| \frac{1}{B} \sum_{i=1}^B g(\bm{x}_i, \hat y_i) \right\|^2$, which we denote by $\mathrm{Tr}(\mathbf{F}_B)$, and \TrF correlate well during training. To demonstrate this, we train SimpleCNN on the CIFAR-10 dataset with 5 different learning rates (from $10^{-3}$ to $10^{-1}$). The outcome is shown in Figure~\ref{app:fig:correlationTrBFvTrF}. We see that for most of the training, with the exception of the final phase, $\mathrm{Tr}(\mathbf{F}^B)$ and $\mathrm{Tr}(\mathbf{F})$ correlate extremely well. Equally importantly, we find that using a large learning affects both $\mathrm{Tr}(\mathbf{F}_B)$ and  \TrF, which further suggests the two are closely connected.

\begin{figure}[H]
\centering
    \begin{subfigure}[t]{0.5\textwidth}
      \includegraphics[width=0.45\columnwidth]{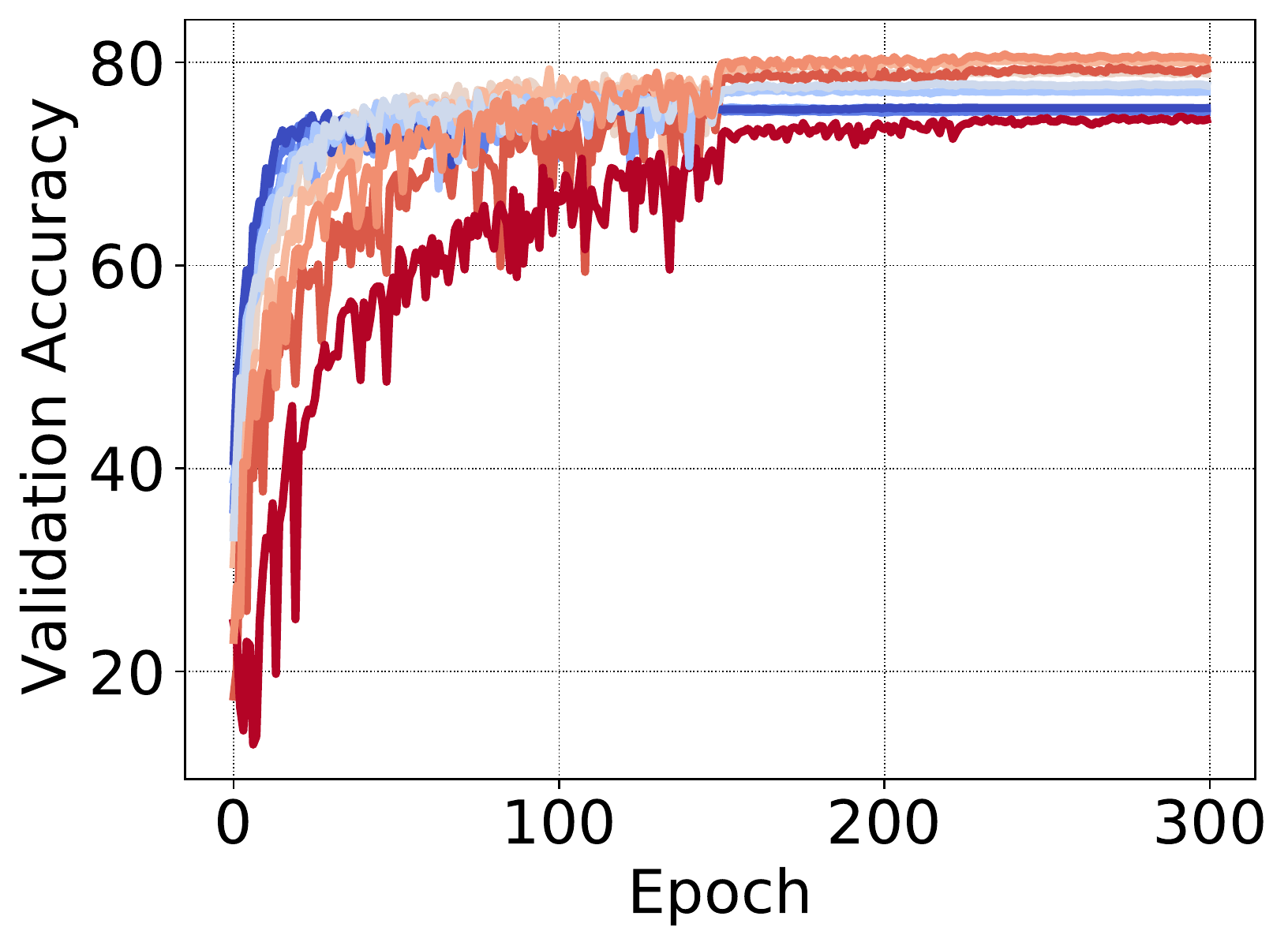}
      \includegraphics[width=0.45\columnwidth]{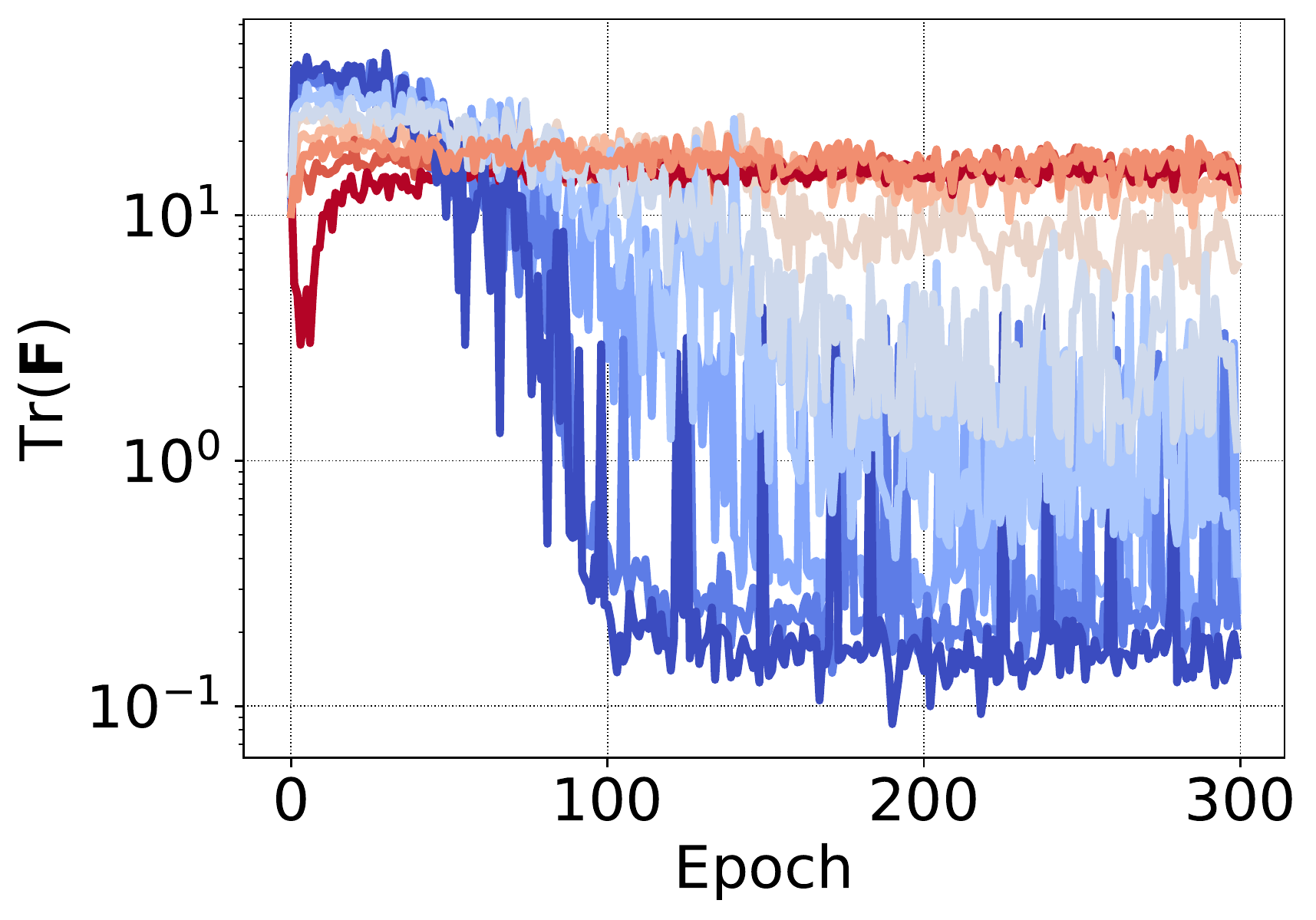}
\caption{Fisher Penalty with $f=10$}
\end{subfigure}%
\begin{subfigure}[t]{0.5\textwidth}
      \includegraphics[width=0.45\columnwidth]{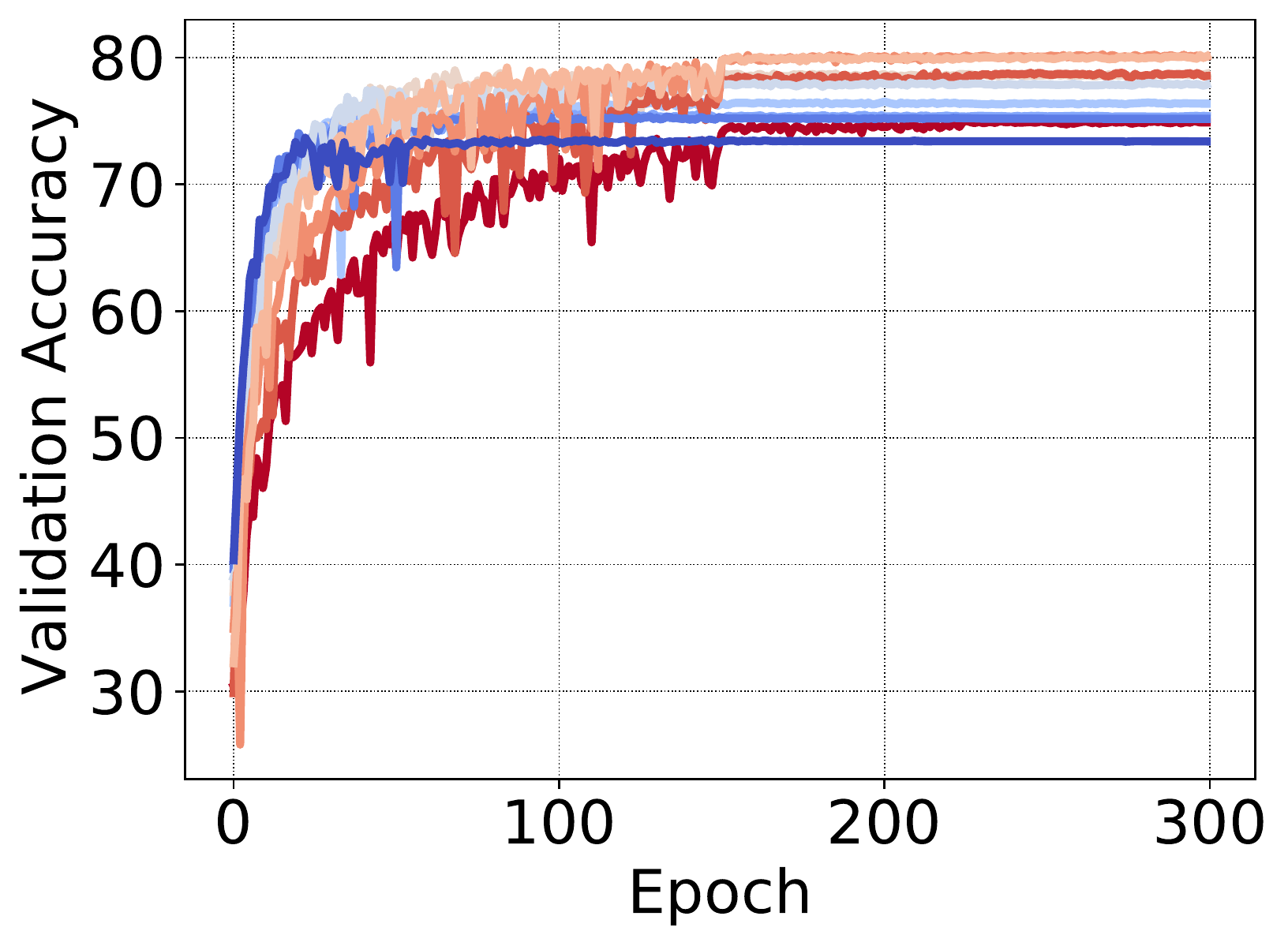}
      \includegraphics[width=0.45\columnwidth]{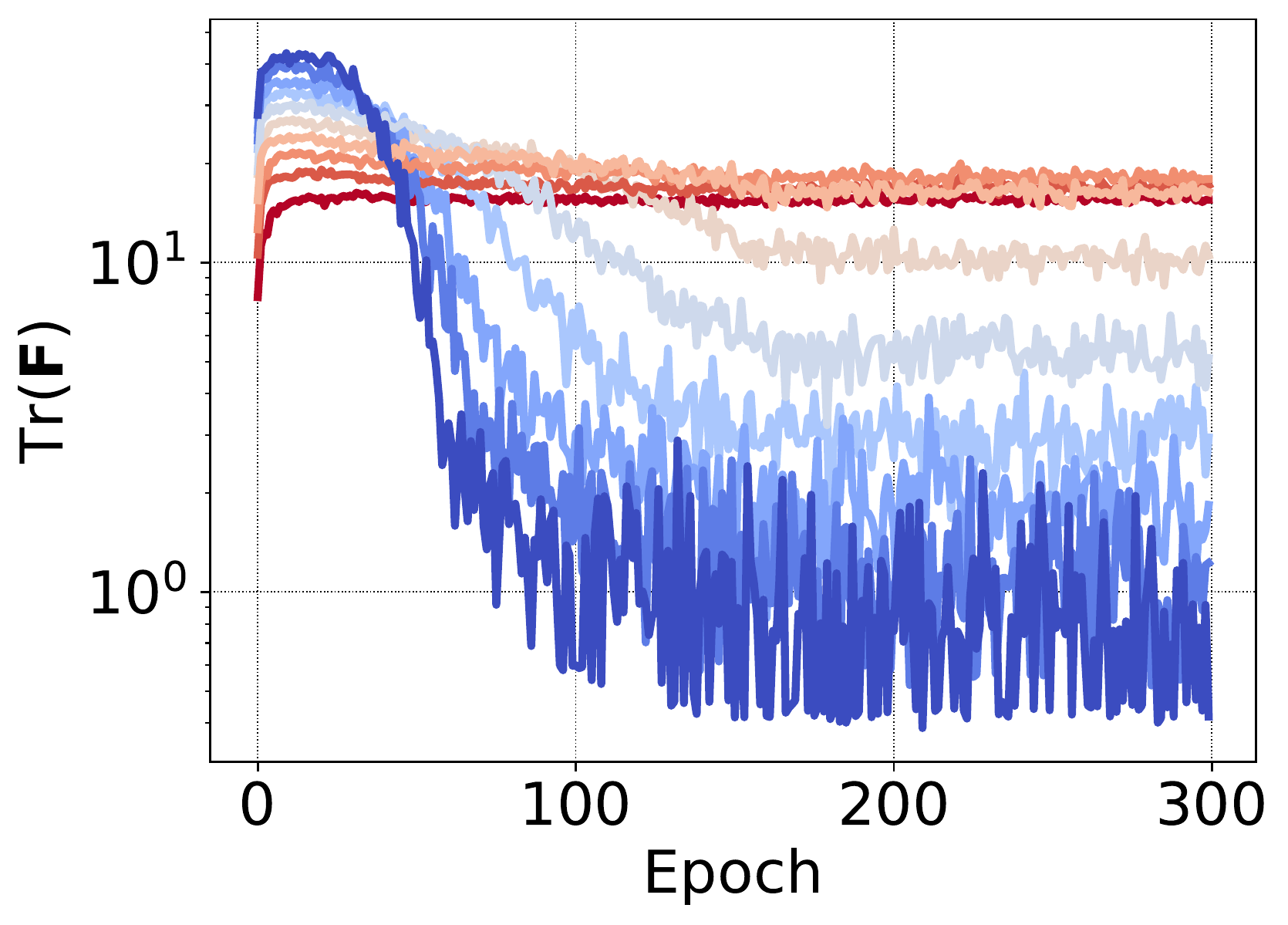}
\caption{Fisher Penalty with $f=1$}
    \end{subfigure}
\caption{A comparison between the effect of recomputing Fisher Penalty gradient every 10 iterations (left) or every iteration (right), with respect to validation accuracy and \TrF. We denote by $f$ the frequency with which we update the gradient. Both experiments result in approximately 80\% test accuracy with the best configuration.}
\label{app:fig:frequency_in_FP}
\end{figure}

\begin{figure}[H]
\centering
      \includegraphics[width=0.45\columnwidth]{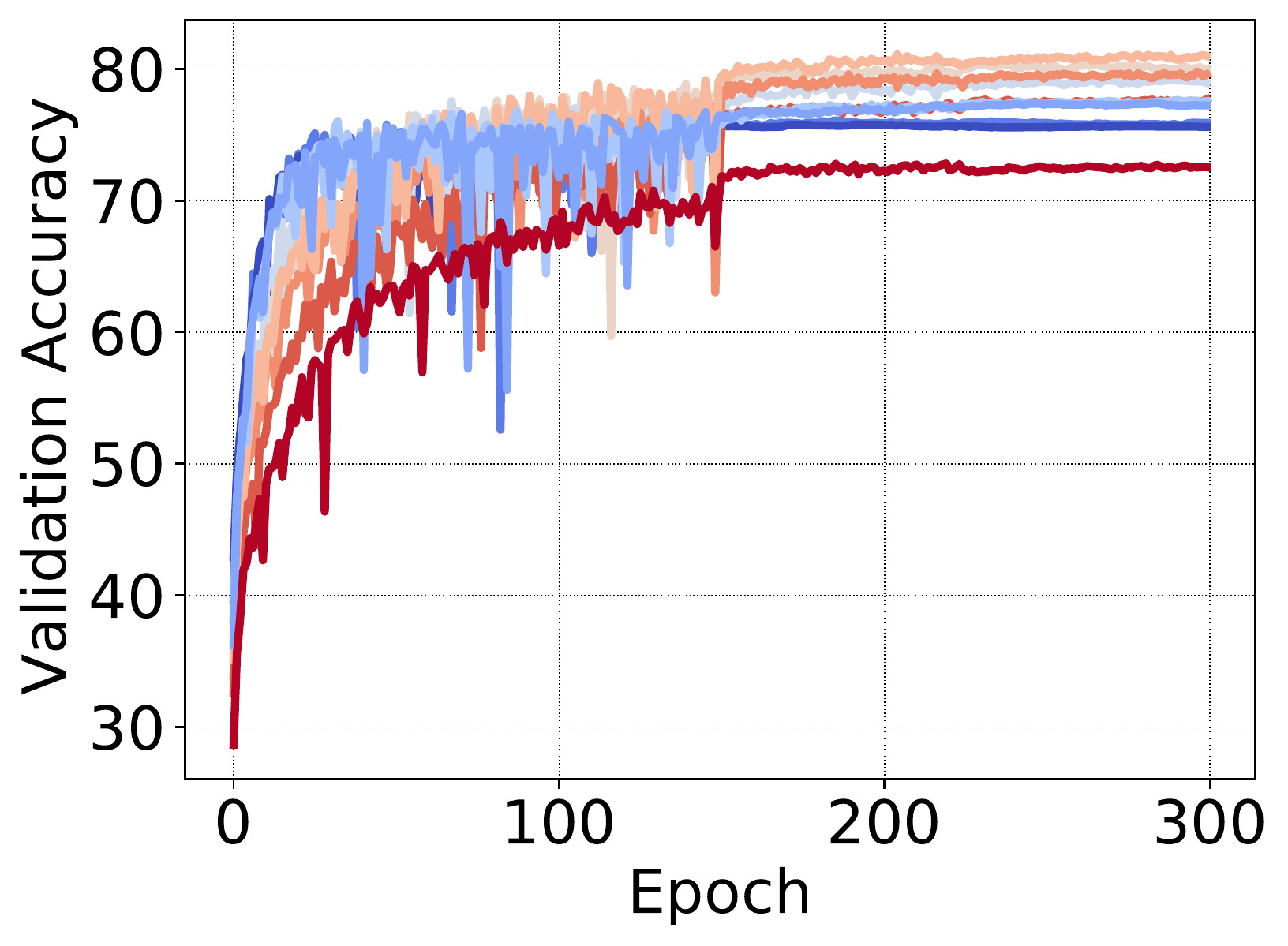}
      \includegraphics[width=0.45\columnwidth]{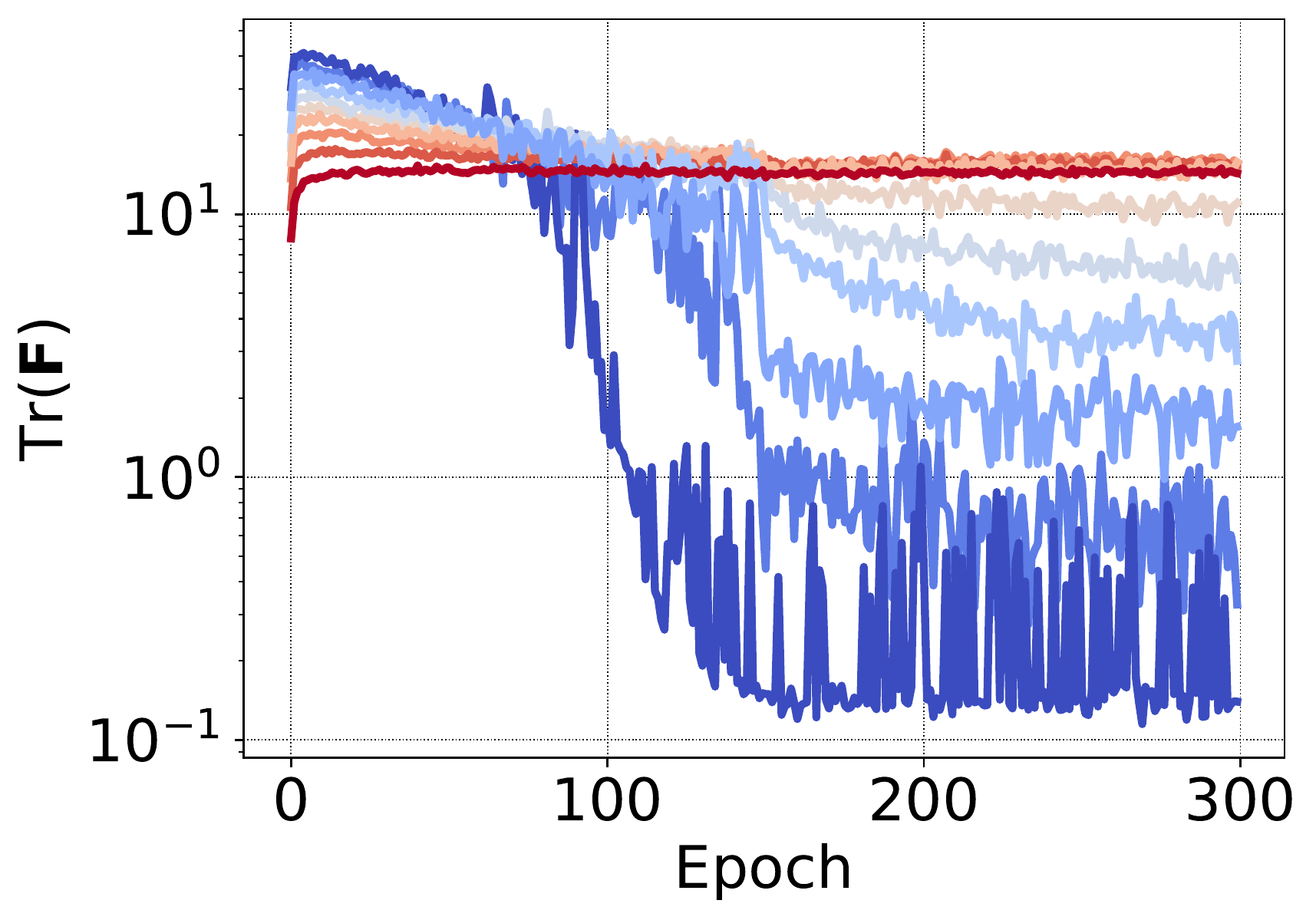}
\caption{Using Fisher Penalty without the approximation results in a similar generalization performance. We penalize the norm of the gradient rather than norm of the mini-batch gradient (as in Equation~\ref{eq_fisher_objective}). We observe that this variant of Fisher Penalty improves generalization to a similar degree as the version of Fisher Penalty used in the paper (c.f. Figure~\ref{app:fig:frequency_in_FP}.), achieving 79.7\% test accuracy.}
\label{app:fig:exact_FP}
\end{figure}

We also update the gradient of $\mathrm{Tr}(\mathbf{F}^B)$ only every 10 optimization steps. We found empirically it does not affect generalization performance nor the ability to regularize \TrF in our setting. However, we acknowledge that it is plausible that this choice would have to be reconsidered in training with very large learning rates or with larger models. 

Figure~\ref{app:fig:frequency_in_FP} compares learning curves of training with \FP recomputed every optimization step with every 10th optimization step. For each, we tune the hyperparameter $\alpha$, checking 10 values equally spaced between $10^{-2}$ and $10^{0}$ on a logarithmic scale. We observe that for the optimal value of $\alpha$, both validation accuracy and \TrF are similar between the two runs. Both experiments achieve approximately 80\% test accuracy.

Finally, to ensure that using the approximation in Equation~\ref{eq_fisher_objective} does not negatively affect how Fisher Penalty improves generalization or reduces the value of \TrF, we experiment with a variant of Fisher Penalty without the approximation. Please recall that we always measure \TrF (i.e. we do not use approximations in computing \TrF that is reported in the plots), regardless of what variant of penalty is used in regularizing the training.

Specifically, we augment the loss function with the norm of the gradient computed on the first example in the mini-batch as follows

\begin{equation}
\label{eq_fisher_objective_exact}
\ell'(\bm{x}_{1:B}, y_{1:B}; \bm{\theta}) = \frac{1}{B} \sum_{i=1}^B \ell(\bm{x}_i,y_i; \bm{\theta}) + \alpha \left\| g(\bm{x}_1, \hat y_1) \right\|^2.
\end{equation}

We apply this penalty in each optimization step. We tune the hyperparameter $\alpha$, checking 10 values equally spaced between $10^{-4}$ and $10^{-2}$ on a logarithmic scale.

Figure~\ref{app:fig:exact_FP} summarizes the results. We observe that the best value of $\alpha$ yields 79.7\% test accuracy, compared to 80.02\% test accuracy yielded by the Fisher Penalty. The effect on \TrF is also very similar. We observe that the best run corresponds to a maximum value of \TrF of 24.16, compared to that of 21.38 achieved by Fisher Penalty. These results suggest that the approximation used in this paper's version of the Fisher Penalty only improves the generalization and flattening effects of Fisher Penalty.

\section{A closer look at the surprising effect of learning rate on the loss geometry in the early phase of training}
\label{app:closer_look}

It is intuitive to hypothesize that the catastrophic Fisher explosion (the initial growth of the value of \TrF) occurs during training with a large learning rate but is overlooked due to not sufficiently fine-grained computation of \TrF. In this section, we show evidence against this hypothesis based on the literature mentioned in the main text. We also run additional experiments in which we compute the value of \TrF at each iteration.

The surprising effect of the learning rate on the geometry of the loss surface (e.g. the value of \TrF) was demonstrated in prior works~\citep{jastrzebski_relation_2018,golatkar2019,lewkowycz2020large,leclerc2020regimes}. In particular, \citet{Jastrzebski2020The,lewkowycz2020large}
 show that training with a large learning rate rapidly escapes regions of high curvature, where curvature is understood as the spectral norm of the Hessian evaluated at the current point of the loss surface. 
Perhaps the most direct experimental data against this hypothesis can be found in \cite{cohen2021gradient} in Figure 1, where training with Gradient Descent finds regions of the loss surface with large curvature for small learning rate rapidly in the early phase of training. 

We also run the following experiment to provide further evidence against the hypothesis. We train SimpleCNN on the CIFAR-10 dataset using two different learning rates, while computing the value of \TrF for every mini-batch. We use 128 random samples in each iteration to estimate \TrF. 

\begin{figure}[H]
\centering
      \includegraphics[width=0.45\columnwidth]{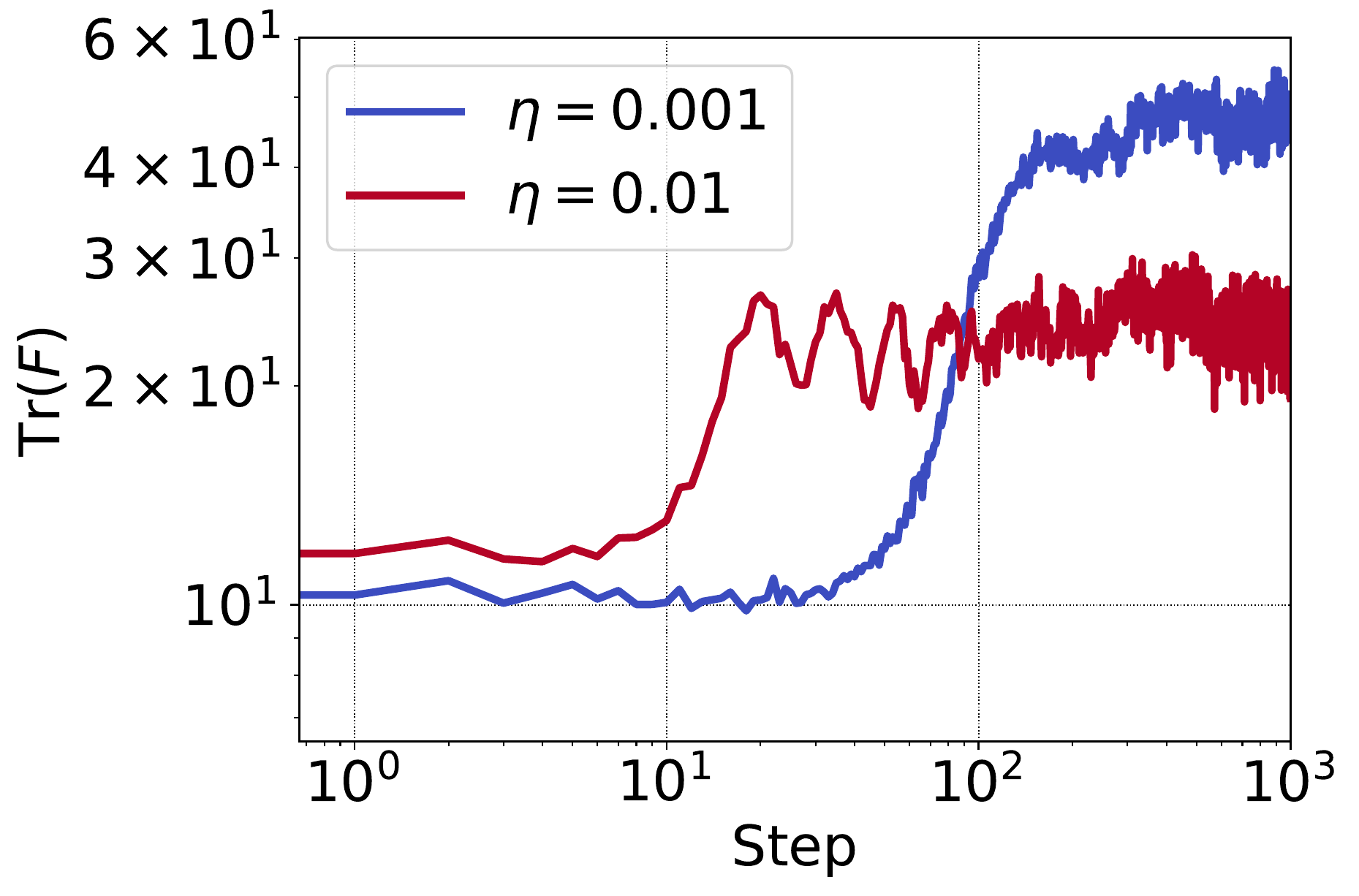}
      \includegraphics[width=0.45\columnwidth]{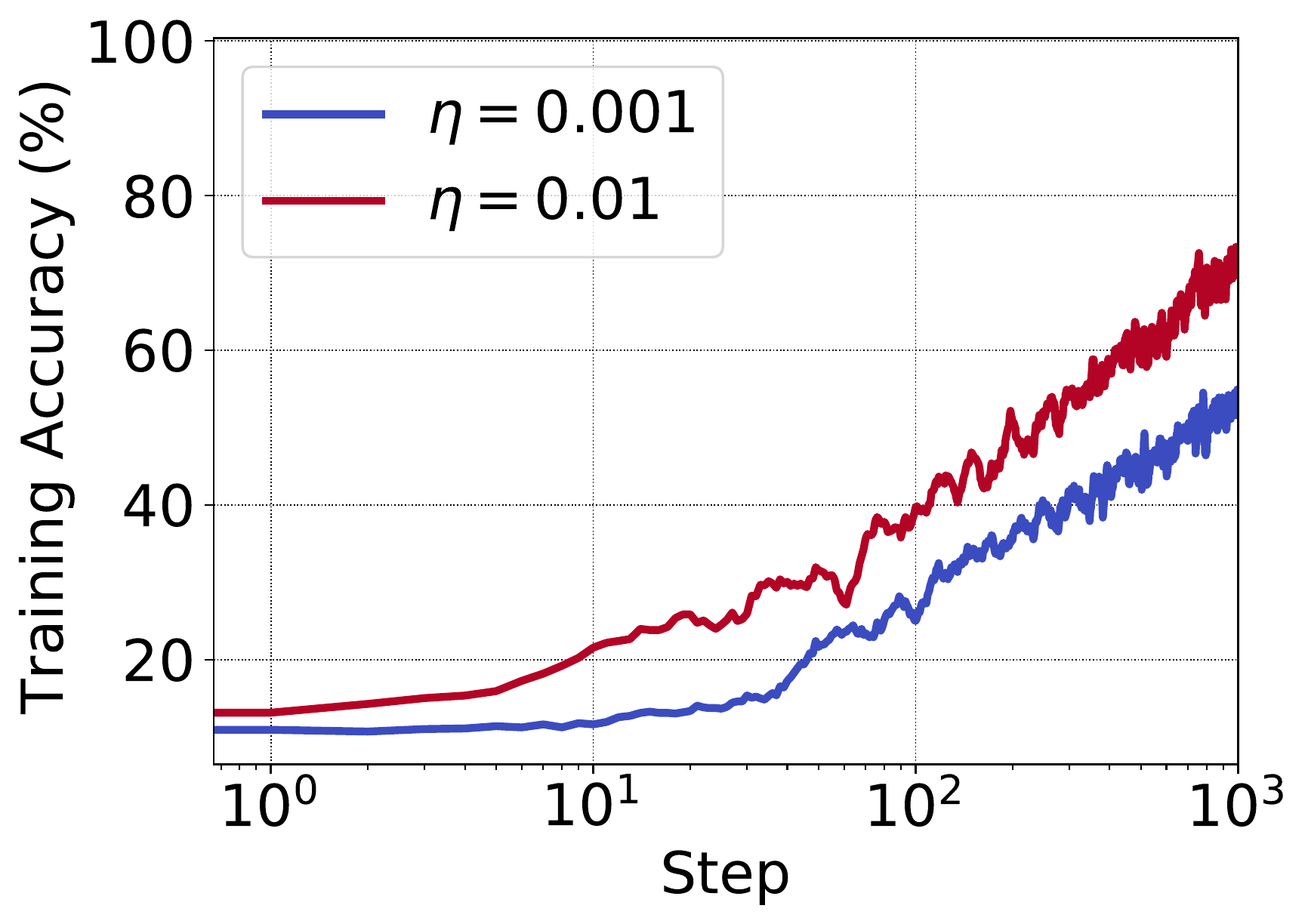}
\caption{Training with a large learning rate never (even for a single optimization step) enters a region with as large value of \TrF as the maximum value of \TrF reached during training with a small learning rate. We run the experiment using SimpleCNN on the CIFAR-10 dataset with two different learning rates. The left plot shows the value of \TrF computed at each iteration, and the right plot shows training accuracy computed on the current mini-batch (curve has been smoothed for clarity). }
\label{app:fig:closer_look_early_phase_lr}
\end{figure}

We find that training with a large learning rate never (even for a single optimization step) enters a region where the value of \TrF is as large as what is reached during training with a small learning rate.
 Figure~\ref{app:fig:closer_look_early_phase_lr} shows the experimental data. 
 
 We also found similar to hold when varying the batch size, see Section~\ref{app:sec:fisher_explosion_holds_in_lb}, which further shows that the observed effects cannot be explained by the difference in learning speed incurred by using a small learning rate.

To summarize, both the published evidence of \citet{Jastrzebski2020The,lewkowycz2020large,cohen2021gradient}, as well as our additional experiments, are inconsistent with the hypothesis that the results in this paper can be explained by differences in training speed between experiments using large and small learning rates.
 
\section{Catastrophic Fisher Explosion holds in training with large batch size}
\label{app:sec:fisher_explosion_holds_in_lb}

In this section, we show preliminary evidence that the conclusions transfer to large batch size training. Namely, we show that (1) catastrophic Fisher explosion also occurs in large batch size training, and (2) Fisher Penalty can improve generalization and close the generalization gap due to using a large batch size~\citep{keskar_large-batch_2017}.

\begin{figure}[H]
\centering
      \includegraphics[width=0.45\columnwidth]{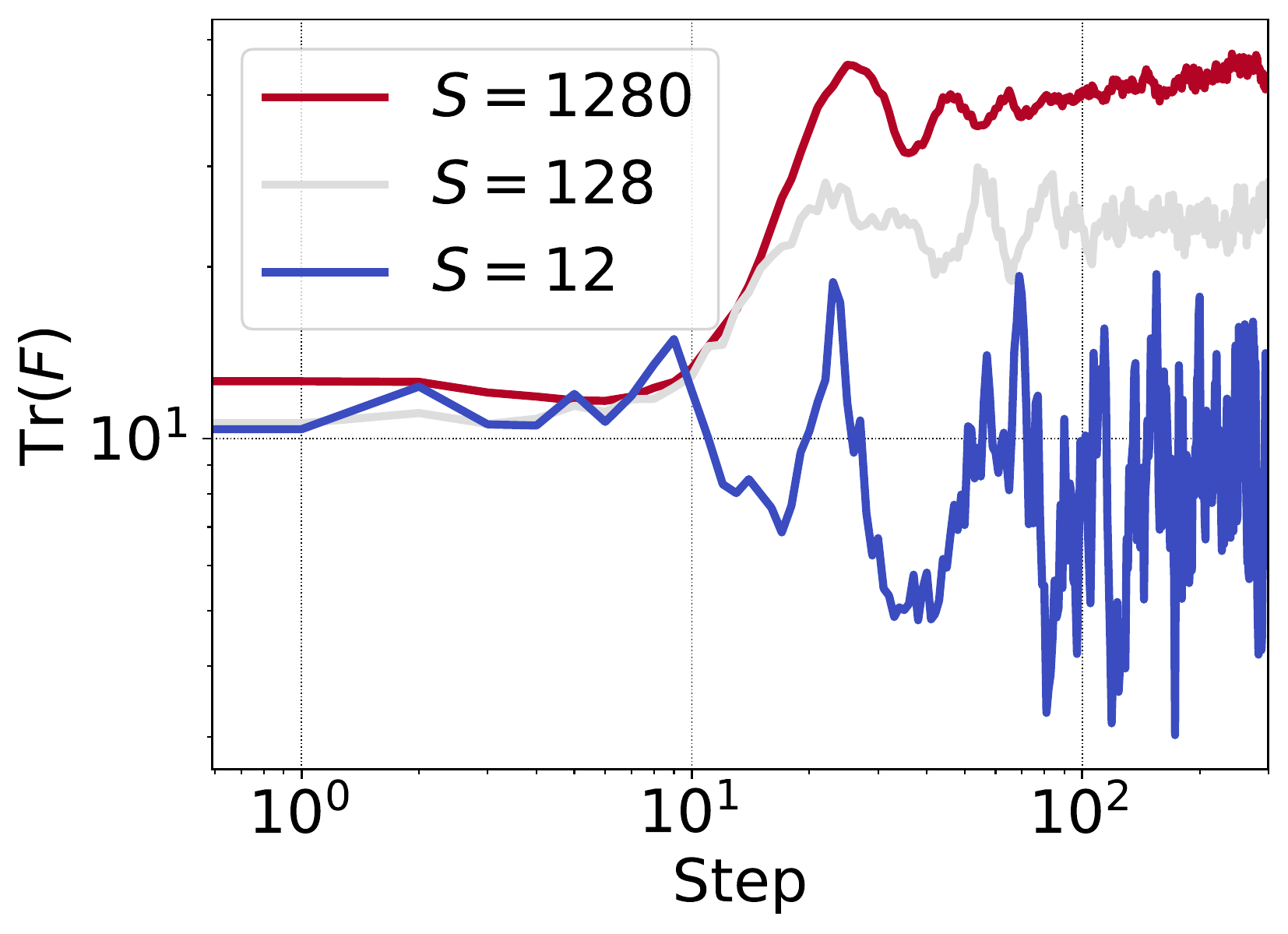}
      \includegraphics[width=0.45\columnwidth]{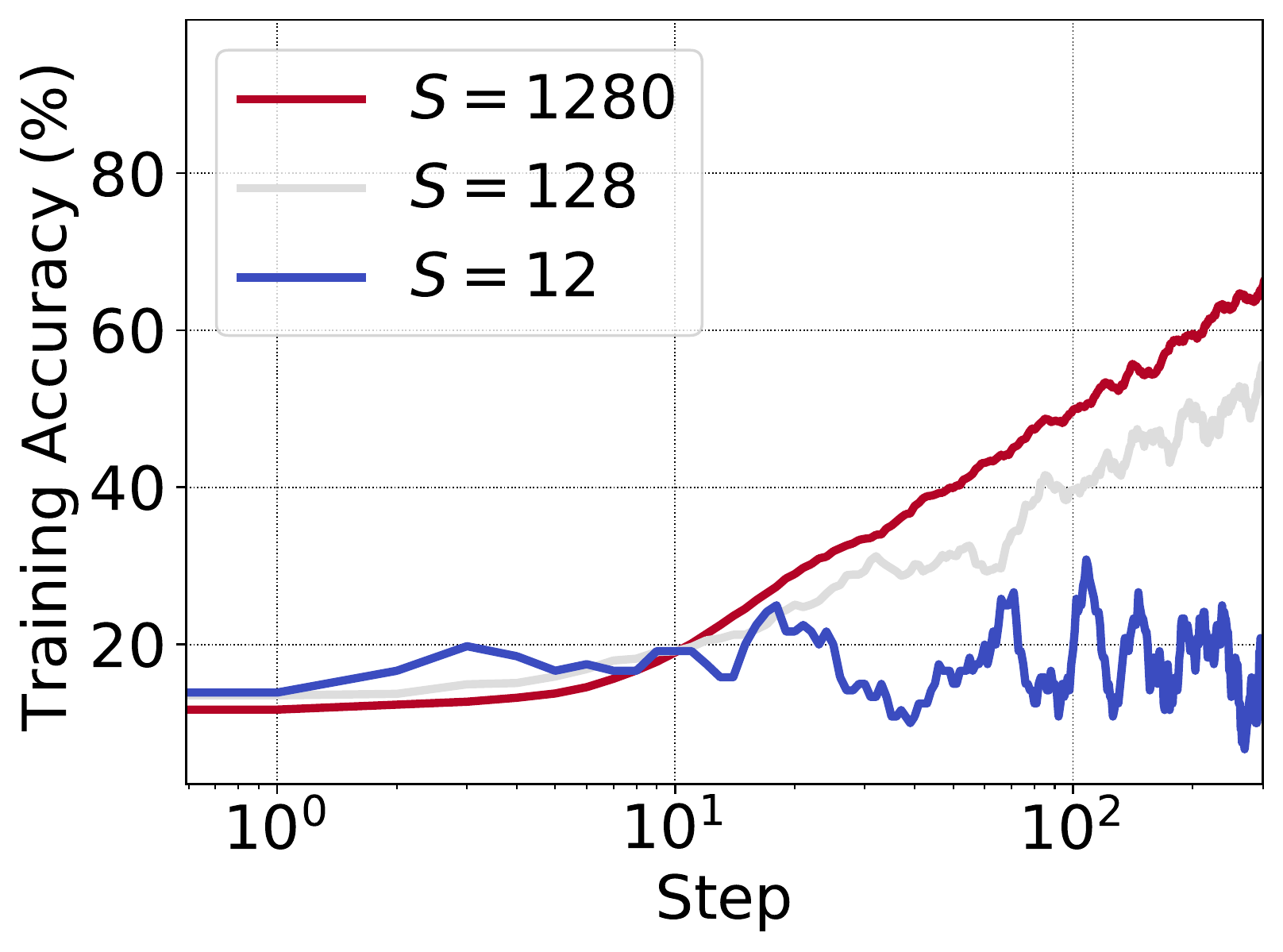}
\caption{Catastrophic Fisher explosion occurs also in large batch size training. Experiment run on the CIFAR-10 and dataset the SimpleCNN model. The left plot shows the value of \TrF computed at each iteration, and the right plot shows training accuracy computed on the current mini-batch (curve has been smoothed for clarity). }
\label{app:fig:closer_look_early_phase_S}
\end{figure}

We first train SimpleCNN on the CIFAR-10 dataset using three different batch sizes, while computing the value of \TrF for every mini-batch. We use 128 random samples in each iteration to estimate \TrF. 
Figure~\ref{app:fig:closer_look_early_phase_S} summarizes the experiment. We observe that training with a large batch size enters a region of the loss surface with a substantially larger value of \TrF than with the small batch size. 

Next, we run a variant of one of the experiments in Table~\ref{tab:fisher_penalty_setting1}. Instead of using a suboptimal (smaller) learning rate, we use a suboptimal (larger) batch size. Specifically, we train SimpleCNN on the CIFAR-10 dataset (without augmentation) with a 10x larger batch size while keeping the learning rate the same. Using a larger batch size results in $3.24\%$ lower test accuracy ($76.94$\% compared to $73.7\%$ test accuracy, c.f. with Table~\ref{tab:fisher_penalty_setting1}). 

\begin{figure}[H]
\centering
      \includegraphics[width=0.45\columnwidth]{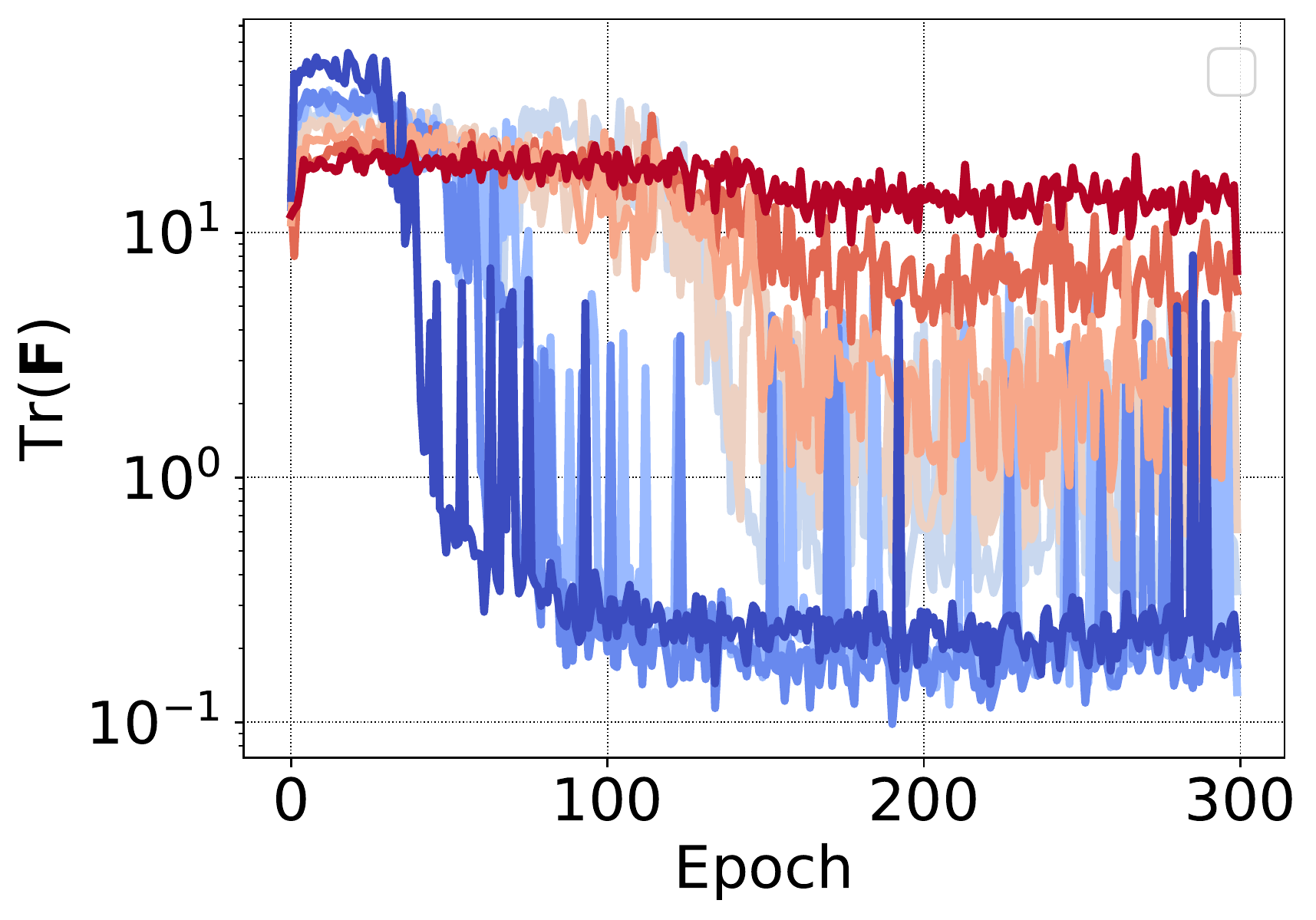}
      \includegraphics[width=0.45\columnwidth]{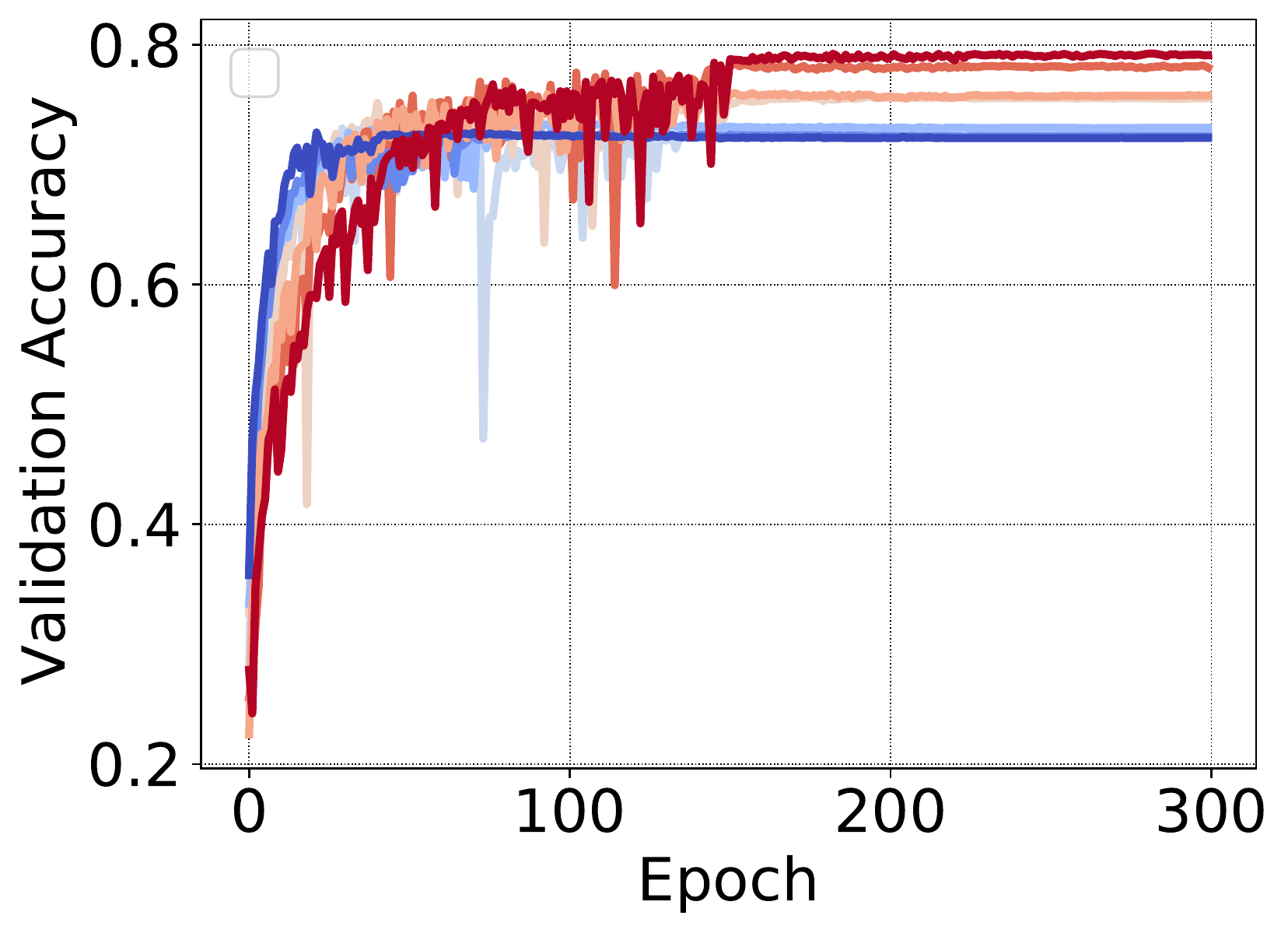}
\caption{Fisher Penalty improves generalization in large batch size training. Experiment run on the CIFAR-10 dataset (without augmentation) and the SimpleCNN model. Warmer color corresponds to larger coefficient used when applying Fisher Penalty.}
\label{app:fig:fisher_penalty_large_bs}
\end{figure}

We next experiment with Fisher Penalty. We apply the penalty in each optimization step and use the first $128$ examples when computing the penalty. We also use a 2x lower learning rate, which stabilizes training but does not improve generalization on its own (training with this learning rate reaches $73.59\%$ test accuracy). Figure~\ref{app:fig:fisher_penalty_large_bs} shows \TrF and validation accuracy during training for different values of the penalty. We observe that Fisher Penalty improves test accuracy from $73.59\%$ to $78.7\%$. Applying Fisher Penalty also effectively reduces the peak value of \TrF.

Taken together, the results suggest that Catastrophic Fisher explosion holds in large batch size training; using a small batch size improves generalization by a similar mechanism as using a large learning rate, which can be introduced explicitly in the form of Fisher Penalty.

\section{\TrH and \TrF correlate strongly}
\label{app:TrH_and_TrF_correlate}

\begin{figure}[H]
\centering
    \begin{subfigure}[t]{0.3\textwidth}
      \includegraphics[width=1.\columnwidth]{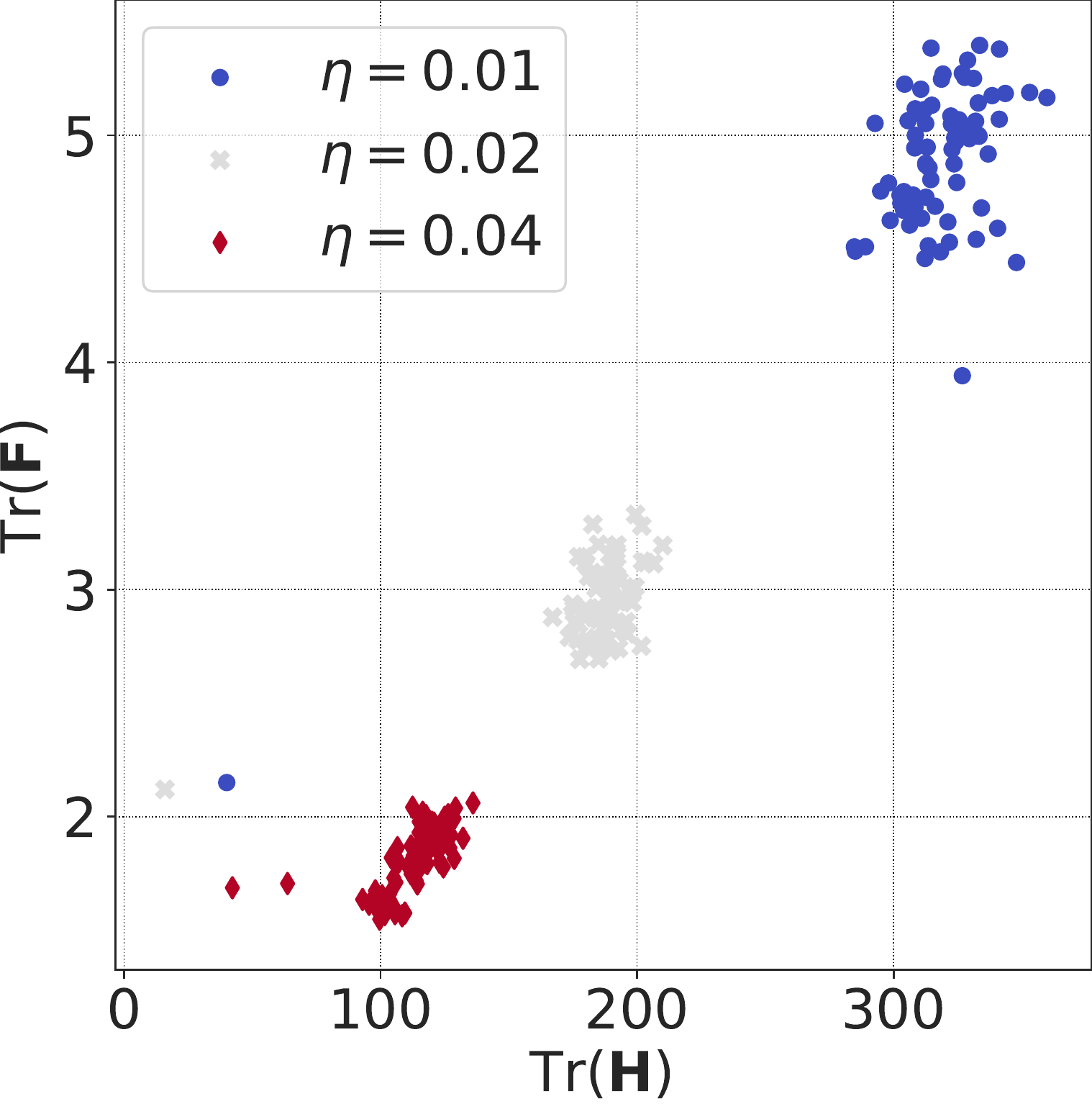}
\caption{DenseNet on CIFAR-100}
    \end{subfigure}
\begin{subfigure}[t]{0.3\textwidth}
      \includegraphics[width=1.\columnwidth]{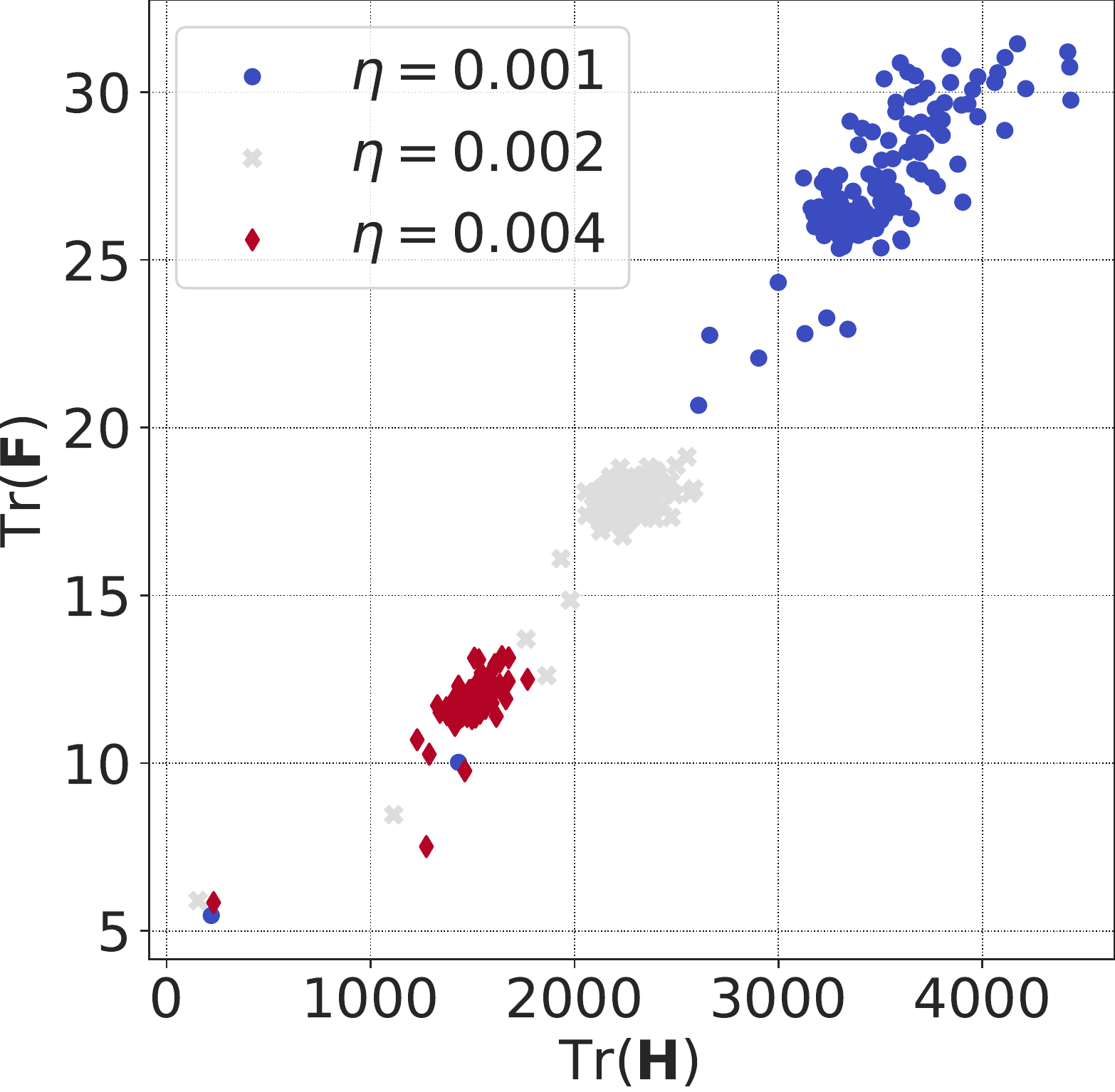}
\caption{SimpleCNN on CIFAR-10}
    \end{subfigure}
    \begin{subfigure}[t]{0.31\textwidth}
      \includegraphics[width=1.\columnwidth]{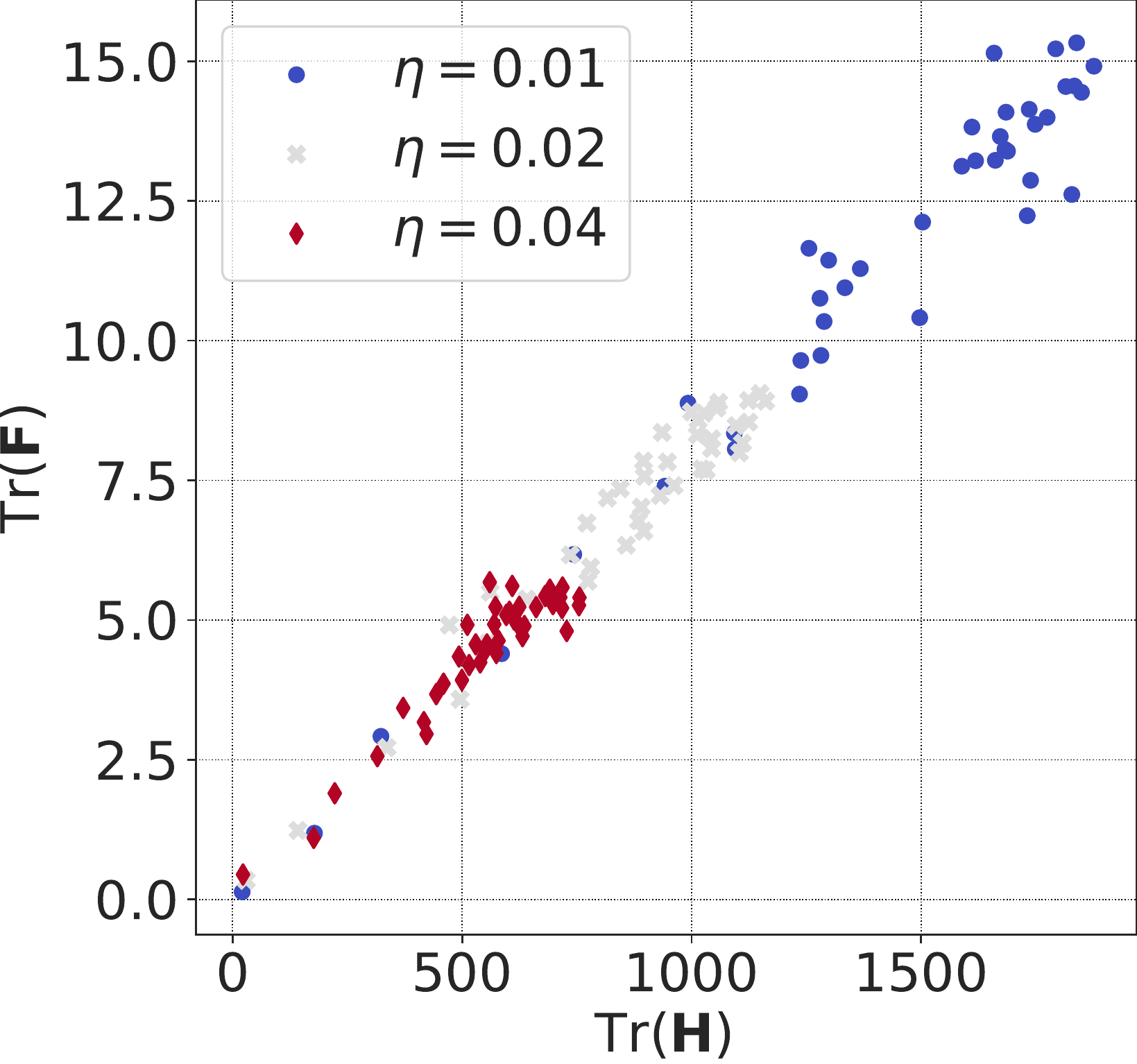}
\caption{ResNet-56 on CIFAR-100}
    \end{subfigure}
  \caption{Correlation between \TrF and \TrH.}
\label{fig:correlationTrHvTrF}
\end{figure}

We demonstrate a strong correlation between \TrH and \TrF for DenseNet, ResNet-56 and SimpleCNN in Figure~\ref{fig:correlationTrHvTrF}. We calculate \TrF using a mini-batch. We see that \TrF has a smaller magnitude (we use a mini-batch gradient) but correlates strongly with \TrH.

\section{Relationship between Fisher Penalty and gradient norm penalty}
\label{app:relationship_FP_and_gp}

We give here a short argument that \GP might act as a proxy for regularizing \TrF. Let $f(\mathbf{x})$ represents the logits of the network, and $\mathcal{L}$ represent the loss. Then $\|\mathbf{g}\|^2= \| \frac{\partial \mathcal{L}(\mathbf{x}, y)}{\partial f(\mathbf{x})}\frac{\partial f(\mathbf{x})}{\partial \mathbf{\theta}} \|^2$, and $\mathrm{Tr}(\mathbf{F}) = \| \frac{\partial \mathcal{L}(\mathbf{x},\hat y)}{\partial f(\mathbf{x})} \frac{\partial f(\mathbf{x})}{\partial \mathbf{\theta}} \|^2$. Hence, in particular, reducing the scale of the Jacobian of the logits with respect to weights will reduce both $\|\mathbf{g}\|^2$ and Tr($\mathbf{F}$). Empirically, penalizing gradient norm seems to be a less effective regularizer, which suggests that it acts as a proxy for regularizing \TrF. A similar argument can be also made for \GPx.

\section{Fisher Penalty Reduces Memorization}
\label{app:FP_and_memorization}

We present here a short argument that Fisher Penalty can be seen as reducing the training speed of examples that are both labeled randomly and for which the model makes a random prediction.

Let $\mathcal{D}_R = \{\bm{x}_i, y_i \}_i$ denote a set of examples where the label $y_i$ is sampled from the discrete uniform distribution $\mathcal{U}$. Let $g(\bm{x},y; \bm{\theta})=\frac{\partial}{\partial \mathbf{\theta}}\ell(\bm{x}, y; \bm{\theta})$ denote gradient of the loss function evaluated on an example ($\bm{x}, y$). 

Assume that the predictive distribution $p_{\bm{\theta}}(y|\bm{x})$ is uniform for all $\bm{x} \in \mathcal{D}_R$. Then the expected mean squared norm of gradient of examples in $\mathcal{D}_R$ over different sampling of labels is $\mathbb{E}[\|g(\bm{x},y)\|_2^2] = \frac{1}{|\mathcal{D}_R|} \sum_i \mathbb{E}_{y \sim \mathcal{U}}[[\| g(\bm{x_i},y) [\|^2] = \frac{1}{|\mathcal{D}_R|} \sum_i \mathbb{E}_{\hat y \sim p_{\bm{\theta}}(\hat y|\bm{x})}[\| g(\bm{x_i},\hat y)_2\|^2]$.

Fisher Penalty aims to penalize the trace of the Fisher Information Matrix. For examples in $\mathcal{D}_R$, \TrF evaluates to $\mathrm{Tr}(\mathbf{F}_R) = \frac{1}{|\mathcal{D}_R|} \mathbb{E}_{\hat y \sim p_{\theta}(y|\bm{x})} \left[ \Vert g(\bm{x_i},\hat y) \Vert_2^2 \right]$, which under our assumptions is equal to $\mathbb{E}[\|g(\bm{x},y)\|^2]$.

We are interested in understanding how Fisher Penalty affects the learning speed of noisy examples. The reduction in the training loss for a given example can be related to its gradient norm using Taylor expansion. Consider the difference in training loss $\Delta \ell = \ell \left( \bm{x}, y; \bm{\theta} - \eta g(\bm{x}, y) \right) - \ell \left( \bm{x}, y; \bm{\theta} \right)$ after performing a single step of gradient descent on this example. Using first-order Taylor expansion we arrive at $\Delta \ell \approx -\eta \|g(\bm{x}, y)\|_2^2$. 

Taken together, penalizing $\|g(\bm{x}, y)\|_2$, which is achieved by penalizing $\mathrm{Tr}(\mathbf{F})_R$, can be seen as slowing down learning on noisy examples.     

However, in practice, we apply Fisher Penalty to all examples in the training set because we do not know which ones are corrupted. Consider $\mathcal{D} = \mathcal{D}_R \cup \mathcal{D}_C$, where $\mathcal{D}$ is the whole training set and $\mathcal{D}_C$ denotes the subset with clean (not altered) labels. Then, $\mathrm{Tr}(\mathbf{F}) = \mathrm{Tr}(\mathbf{F}_R) + \mathrm{Tr}(\mathbf{F}_C)$, where $\mathrm{Tr}(\mathbf{F})$ denotes trace of the FIM evaluated on the whole dataset, and $\mathrm{Tr}(\mathbf{F}_C)$ ($\mathrm{Tr}(\mathbf{F}_R)$) denotes the trace of the FIM on the clean (noisy) subset of the dataset.

Hence, if $\mathrm{Tr}(\mathbf{F}_R) \gg \mathrm{Tr}(\mathbf{F}_C)$, we can expect Fisher Penalty to disproportionately slow down training of noisy examples. This assumption is likely to hold because the clean examples tend to be learned much earlier in training than noisy ones~\cite{arpit_closer_2017}. In experiments, we indeed observe that the gradient norm of examples with noisy labels is disproportionately affected by Fisher Penalty, and also that learning on noisy examples is slower.

\section{Additional Experimental Details}

\subsection{Early phase \TrF correlates with final generalization}
\label{sec_early_final_details}

Here, we describe additional details for experiments in Section~\ref{sec_trf_generalization}.

In the experiments with batch size, for CIFAR-10, we use batch sizes 100, 500 and 700, and $\epsilon = 1.2$. For CIFAR-100, we use batch sizes 100, 300 and 700, and $\epsilon = 3.5$. These thresholds are crossed between 2 and 7 epochs across different hyperparameter settings. The remaining details for CIFAR-100 and CIFAR-10 are the same as described in the main text. The optimization details for these datasets are as follows.

\textbf{ImageNet}: No data augmentation was used in order to allow training loss to converge to small values. We use a batch size of 256. Training is done using SGD with momentum set to 0.9, weight decay set to $1e-4$, and with base learning rates in $\{0.001, 0.01, 0.1\}$. Learning rate is dropped by a factor of 0.1 after 29 epochs and training is ended at around 50 epochs at which most runs converge to small loss values. No batch normalization is used and weight are initialized using Fixup \cite{zhang2019fixup}. For each hyperparameter setting, we run two experiments with different random seeds (due to the computational overhead). We compute \TrF using 2500 samples (similarly to \cite{Jastrzebski2020The}).

\textbf{CIFAR-10}: We used random flipping as data augmentation. In the experiments with variation in learning rates (used $\{0.007, 0.01, 0.05\}$), we use a batch size of 256. In the experiments with variation in batch size (used 100, 500 and 700), we use a learning rate of 0.02. Training is done using SGD with momentum set to 0.9, weight decay set to $1e-5$. Learning rate is dropped by a factor of 0.5 at epochs 60, 120, and 170, and training is ended at 200 epochs at which most runs converge to small loss values. No batch normalization is used and weight are initialized using \cite{arpit2019initialize}. For each hyperparameter setting, we run 32 experiments with different random seeds. We compute \TrF using 5000 samples.

\textbf{CIFAR-100}: No data augmentation was used for CIFAR-100 to allow training loss to converge to small values. We used random flipping as data augmentation for CIFAR-10. In the experiments with variation in learning rates (used $\{0.005, 0.001, 0.01\}$), we use a batch size of 100. In the experiments with variation in batch size (used 100, 300 and 700), we use a learning rates of 0.02. Training is done using SGD with momentum set to 0.9, weight decay set to $1e-5$. Learning rate is dropped by a factor of 0.5 at epochs 60, 120, and 170, and training is ended at 200 epochs at which most runs converge to small loss values. No batch normalization is used and the weights are initialized using \cite{arpit2019initialize}. For each hyperparameter setting, we run 32 experiments with different random seeds. We compute \TrF using 5000 samples.

\subsection{Fisher Penalty}
\label{app:fisher_penalty_experimental_details}

Here, we describe the remaining details for the experiments in Section~\ref{sec:fisher_penalty}. We first describe how we tune hyperparameters in these experiments. In the remainder of this section, we describe each setting used in detail.

\paragraph{Tuning hyperparameters} In all experiments, we refer to the optimal learning rate $\eta^*$ as the learning rate found using grid search. In most experiments, we check 5 different learning rate values uniformly spaced on a logarithmic scale, usually between $10^{-2}$ and $10^{0}$. In some experiments, we adapt the range to ensure that it includes the optimal learning rate. We tune the learning rate only once for each configuration (i.e. we do not repeat it for different random seeds).

In the first setting, for most experiments involving gradient norm regularizers, we use 10$\times$ smaller learning rate than $\eta^*$. For TinyImageNet, we use 30$\times$ smaller learning rate than $\eta^*$. To pick the regularization coefficient $\alpha$, we evaluate 10 different values uniformly spaced on a logarithmic scale between $10^{-1} \times v$ to $10^{1} \times v$ with $v \in \mathbb{R}_+$. We choose the best performing $\alpha$ according to validation accuracy. We pick the value of $v$ manually with the aim that the optimal $\alpha$ is included in this range. We generally found that $v=0.01$ works well for \GP, \GPr, and \FP. For \GPx we found in some experiments that it is necessary to pick larger values of $v$. 

\paragraph{Measuring \TrF} We measure \TrF using the number of examples equal to the batch size used in training. For experiments with Batch Normalization layers, we use Batch Normalization in evaluation mode due to the practical reason that computing \TrF uses a batch size of 1, and hence \TrF is not defined for a network with Batch Normalization layers in training mode. 

\paragraph{DenseNet on the CIFAR-100 dataset} We use the DenseNet (L=40, k=12) configuration from \cite{huang_densely_2016}. We largely follow the experimental setting in \cite{huang_densely_2016}. We use the standard data augmentation (where noted) and data normalization for CIFAR-100. We hold out random 5000 examples as the validation set. We train the model using SGD with a momentum of 0.9, a batch size of 128, and a weight decay of 0.0001. Following \cite{huang_densely_2016}, we train for 300 epochs and decay the learning rate by a factor of 0.1 after epochs 150 and 225. To reduce variance, in testing we update Batch Normalization statistics using 100 batches from the training set.

\paragraph{Wide ResNet on the CIFAR-100 dataset} We train Wide ResNet (depth 44 and width 3, without Batch Normalization layers). We largely follow experimental setting in \cite{he_deep_2015}.We use the standard data augmentation and data normalization for CIFAR-100. We hold out random 5000 examples as the validation set. We train the model using SGD with a momentum of 0.9, a batch size of 128, and a weight decay of 0.0010. Following \cite{he_deep_2015}, we train for 300 epochs and decay the learning rate by a factor of 0.1 after epochs 150 and 225. We remove Batch Normalization layers. To ensure stable training we use the SkipInit initialization~\citep{de2020batch}. 

\paragraph{VGG-11 on the CIFAR-100 dataset} We adapt the VGG-11 model~\citep{Simonyan15} to CIFAR-100. We do not use dropout nor Batch Normalization layers. We hold out random 5000 examples as the validation set. We use the standard data augmentation (where noted) and data normalization for CIFAR-100. We train the model using SGD with a momentum of 0.9, a batch size of 128, and a weight decay of 0.0001. We train the model for 300 epochs and decay the learning rate by a factor of 0.1 after every 40 epochs starting from epoch 80.

\paragraph{SimpleCNN on the CIFAR-10 dataset} We also run experiments on the CNN example architecture from the Keras example repository~\citep{chollet_keras_2015}\footnote{Accessible at \href{https://github.com/keras-team/keras/blob/master/examples/cifar10\_cnn.py}{https://github.com/keras-team/keras/blob/master/examples/cifar10\_cnn.py}.}, which we change slightly. Specifically, we remove dropout and reduce the size of the final fully-connected layer to 128.  We train it for 300 epochs and decay the learning rate by a factor of 0.1 after the epochs 150 and 225. We train the model using SGD with a momentum of 0.9, and a batch size of 128.

\paragraph{Wide ResNet on the TinyImageNet dataset} We train Wide ResNet (depth 44 and width 3, with Batch Normalization layers) on TinyImageNet~\citet{Le2015TinyIV}. TinyImageNet consists of a subset of 100,000 examples from ImageNet that we downsized to 32$\times$32 pixels. We train the model using SGD with a momentum of 0.9, a batch size of 128, and a weight decay of 0.0001. We train for 300 epochs and decay the learning rate by a factor of 0.1 after epochs 150 and 225. We do not use validation in TinyImageNet due to its larger size. To reduce variance, in testing we update Batch Normalization statistics using 100 batches from the training set.

\subsection{Fisher Penalty Reduces Memorization}
\label{app:fisher_penalty_prevents_memorization_experimental_details}

Here, we describe additional experimental details for Section~\ref{sec:fisher_penalty_prevents_memorization}. We use two configurations described in Section~\ref{app:fisher_penalty_experimental_details}: VGG-11 trained on CIFAR-100 dataset, and Wide ResNet trained on the CIFAR-100 dataset. We tune the regularization coefficient $\alpha$ in the range $\{ 0.01,0.1,0.3 1,10\}$, with the exception of \GPx for which we use the range $\{10, 30, 100, 300, 1000\}$. We tuned the mixup coefficient in the range $\{0.4, 0.8, 1.6, 3.2, 6.4\}$. We removed weight decay in these experiments. We use validation set for early stopping, as commonly done in the literature.

\subsection{Early \TrF influences final curvature}
\label{sec_trf_final_minima_details}

\textbf{CIFAR-10}: We used random flipping as data augmentation for CIFAR-10. We use a learning rate of 0.02 for all experiments. Training is done using SGD with momentum 0.9, weight decay $1e-5$, and batch size as shown in figures. The learning rate is dropped by a factor of 0.5 at 80, 150, and 200 epochs, and training is ended at 250 epochs. No batch normalization is used and the weights are initialized using \cite{arpit2019initialize}. For each batch size, we run 32 experiments with different random seeds. We compute \TrF using 5000 samples.

\textbf{CIFAR-100}: No data augmentation is used. We use a batch size of 100 for all experiments. Training is done using SGD with momentum 0.9, weight decay $1e-5$, and with base learning rate as shown in figures. The learning rate is dropped by a factor of 0.5 at 80, 150, and 200 epochs, and training is ended at 250 epochs. No batch normalization is used and the weights are initialized using \cite{arpit2019initialize}. For each learning rate, we run 32 experiments with different random seeds. We compute \TrF using 5000 samples.

\end{document}